\documentclass[letterpaper,9pt,twocolumn]{extarticle}

\usepackage{graphicx}
\usepackage{amsmath}
\usepackage{amsfonts}
\usepackage{mathtools}
\usepackage{subfigure}
\usepackage{xcolor}
\usepackage{here}
\usepackage{tcolorbox}
\usepackage{array}

\usepackage{url}
\usepackage{authblk}
\usepackage{listings}
\usepackage[ruled,linesnumbered,vlined]{algorithm2e}
\SetKwBlock{Function}{function}{end function}
\SetKwComment{CommentTmp}{\blue{$\triangleright$\:}}{}
\DontPrintSemicolon
\newcommand{\Comment}[1]{\CommentTmp*[r]{\small\blue{#1}\!\!\!\!\!\!}}
\newcommand{\CommentHere}[1]{\hfill\CommentTmp*[h]{\small\blue{#1}\!\!\!\!\!\!}}

\makeatletter
\newcommand\multiline[1]{\parbox[t]{\dimexpr\linewidth-\ALG@thistlm}{#1}}
\makeatother
\newcommand{\red}[1]{\textcolor{red}{#1}}
\newcommand{\blue}[1]{\textcolor{blue}{#1}}
\newcommand{\green}[1]{\textcolor{teal}{#1}}

\def\*#1{\boldsymbol{#1}}

\newcommand{\prm}[0]{\prime}
\newcommand{\tw}[0]{\textwidth}
\newcommand{\ig}[2]{\includegraphics[clip,width=#1\tw]{#2}}

\newcommand{\veps}[0]{\varepsilon}

\newcommand{\mr}[1]{\mathrm{#1}}

\newcommand{\argmax}{\mathop{\text{argmax}}}

\newcommand{\eq}[1]{(\ref{#1})}

\newcommand{\lw}[1]{\smash{\lower2.ex\hbox{#1}}}

\newcommand{\rbr}[1]{\left(#1\right)}
\newcommand{\sbr}[1]{\left[#1\right]}
\newcommand{\cbr}[1]{\left\{#1\right\}}

\newcommand{\RR}{\mathbb{R}}
\newcommand{\NN}{\mathbb{N}}

\newcommand{\EE}{\mathbb{E}}
\newcommand{\II}{\mathbb{I}}

\newcommand{\cA}{{\mathcal A}}

\newcommand{\cD}{{\mathcal D}}

\newcommand{\cG}{{\mathcal G}}

\newcommand{\cL}{{\mathcal L}}

\newcommand{\cN}{{\mathcal N}}
\newcommand{\cO}{{\mathcal O}}
\newcommand{\cP}{{\mathcal P}}

\newcommand{\cU}{{\mathcal U}}

\usepackage[margin=.5in]{geometry} 
\ifdefined\draft
\newcommand{\note}[1]{}
\newcommand{\remove}[1]{}
\else
\newcommand{\note}[1]{\red{(Memo: #1)}} % メモ
\newcommand{\remove}[1]{
\begingroup
\color{blue}
#1
\endgroup
}
\fi

\title{\Huge
Bayesian Optimization for Simultaneous Selection of \\
Machine Learning Algorithms and Hyperparameters on \\
Shared Latent Space
}

\author[1]{Kazuki~Ishikawa}
\author[1]{Ryota~Ozaki}
\author[1]{Yohei~Kanzaki}
\author[2,3]{Ichiro~Takeuchi}
\author[1]{Masayuki~Karasuyama\thanks{karasuyama@nitech.ac.jp}}
\affil[1]{Nagoya Institute of Technology}
\affil[2]{Nagoya University}
\affil[3]{RIKEN}

\date{}

\begin{document}

\maketitle

\begin{abstract}
% 機械学習アルゴリズムとそのハイパーパラメータの最適な組み合わせの選択は高性能な機械学習システム開発において重要である．
Selecting the optimal combination of a machine learning (ML) algorithm and its hyper-parameters is crucial for the development of high-performance ML systems.
%
% 一方で，機械学習アルゴリズムとハイパーパラメータの組み合わせは膨大であるため，網羅的な検証では多くの時間を必要とする．
However, since the combination of ML algorithms and hyper-parameters is enormous, the exhaustive validation requires a significant amount of time.
%
% 既存研究の主要なアプローチの一つとしてベイズ最適化に基づく方法が広く研究されている．
Many existing studies use Bayesian optimization (BO) for accelerating the search.
%
% しかし，機械学習アルゴリズム毎に異なるハイパーパラメータ空間が存在するため，効率的な探索手法を構築することが難しい．
On the other hand, a significant difficulty is that, in general, there exists a different hyper-parameter space for each one of candidate ML algorithms. 
BO-based approaches typically build a surrogate model independently for each hyper-parameter space, by which sufficient observations are required for all candidate ML algorithms. 
%
% 本稿では全ての機械学習アルゴリズムのハイパーパラメータ空間を共有の潜在空間上へと写像し，その潜在空間上でのベイズ最適化を提案する．
In this study, our proposed method embeds different hyper-parameter spaces into a shared latent space, in which a surrogate multi-task model for BO is estimated.
%
% この方法では，共有潜在空間上で各機械学習アルゴリズムの観測結果を共有することができるため，少ない観測をより効率的に活かした探索が期待できる．
This approach can share information of observations from different ML algorithms by which efficient optimization is expected with a smaller number of total observations.
%
% また，潜在空間推定のための敵対的正則化を用いた事前学習と，効果的な事前学習済みモデルの選択のためのランキング学習についても提案する．
We further propose the pre-training of the latent space embedding with an adversarial regularization, and a ranking model for selecting an effective pre-trained embedding for a given target dataset.
%
% 計算機実験ではOpenMLで利用可能なデータセットを用いて提案手法の有効性を示す．
Our empirical study demonstrates effectiveness of the proposed method through datasets from OpenML. 
\end{abstract}

% --------------------------------------------------
% \section{はじめに}\label{sec:introduction}
\section{Introduction}
\label{sec:introduction}

% 近年．機械学習はさまざまな分野で活用される．
% 例えば，金融分野ではクレジットカード利用の際，過去の取引内容や取引頻度，取引額などの特徴を基に，現在の取引が不正取引なのかを検証する．
% また，医療分野では生成可能な化合物の効果や副作用を予測することで，製品を販売するまでに必要な時間の短縮や費用の削減が期待できる．

% しかし，機械学習を活用するには機械学習アルゴリズムとハイパーパラメータの選択を行う必要がある．
% この機械学習アルゴリズムとハイパーパラメータの組み合わせは膨大であるため，適切な組み合わせの発見を行うためには多くの時間を必要としてしまう．
% また，タスク毎に適切な機械学習アルゴリズムとハイパーパラメータは異なる場合が多いことから，専門家であっても適切な組み合わせの発見は試行錯誤を必要とし，専門家にとって大きな負担となってしまう．

% 近年，機械学習は劇的な普及を遂げ，さまざまな分野で活用されるようになった．
In recent years, machine learning (ML) has gained widespread adoption and has been utilized in various fields.
%
% しかしながら，機械学習が高いパフォーマンスを発揮するには適切な機械学習アルゴリズムおよびハイパーパラメータ選択が必要になる．
However, to achieve high prediction performance, the precise selection of an appropriate ML algorithm and its hyper-parameters (HPs) is indispensable.
%
% 機械学習アルゴリズムとハイパーパラメータの組み合わせは膨大であるためこの選択は専門家であっても容易ではなく，多大なコストが必要となっている．
There exist a large number of possible combinations of ML algorithms and HPs, which makes their efficient selection is challenging.

% この問題に対して，Automated Machine Learning (AutoML)\cite{automl}と呼ばれる機械学習の自動化技術が有効である．
% AutoMLでは機械学習アルゴリズムとハイパーパラメータの組み合わせのように，機械学習アルゴリズムの構築に必要なチューニングを自動的に行うことで，試行錯誤に必要な労力や時間を削減しつつ，より精度の高いモデルを効率的に発見することを目指す．
% 適切な機械学習アルゴリズムとハイパーパラメータの組み合わせを自動的に発見する最も単純な解決手段は全ての組み合わせで実験を行うことである．
% しかし，組み合わせが膨大であり，一度の実験に多くの時間を必要としてしまう場合があるため，この方法は非効率的である．

% このように，膨大な組み合わせの中から最適な組み合わせを効率的に発見する方法としてベイズ最適化\cite{brochu2010tutorial}がよく知られている．
% ベイズ最適化はある入力値$\mathbf{x}$と出力値$y$の関係を確率モデルで表現し，$\mathbf{x}$に対してどの程度の$y$が期待されるかの予測を，
% 予測自身の不確かさも含めてモデル化する．これにより，現時点で確実に有望な$\mathbf{x}$のみならず，不確実ながらも改善があり得るような未知の領域も含めた探索が可能となる．

% この問題に対して，Automated Machine Learning (AutoML)\cite{automl}と呼ばれる機械学習の自動化技術が有効である．
The simultaneous selection of an ML algorithm and HPs is called Combined Algorithm Selection and Hyper-parameter optimization (CASH) problem \cite{thornton2013auto,NIPS2015_11d0e628}.
%
% AutoMLでは機械学習アルゴリズムとハイパーパラメータの組み合わせのように，機械学習モデルの構築に必要なチューニングを自動的に行うことで，より精度の高いモデルを効率的に発見することを目指す．
The CASH problem is formulated as the optimization of the prediction performance with respect to the pair of an ML algorithm and its HP setting.
%
% AutoMLでの探索に用いられる技術としてベイズ最適化\cite{brochu2010tutorial}がよく知られている．
To solve a CASH problem, Bayesian optimization (BO) \cite{brochu2010tutorial,snoek2012practical,bergstra2011algorithms} is often used.
%
% ベイズ最適化は探索空間上に構築した確率モデル（代理モデル）により最適化対象を表現し（AutoMLの場合は例えば検証データでの予測精度が最適化対象），不確実性を考慮しながら改善が期待される未知領域の探索を行うことができる．
BO constructs a probabilistic surrogate model (typically, the Gaussian process) to approximate the objective function in the search space (e.g., to predict the validation accuracy of an ML model in the HP space), by which a decision making considering the current uncertainty of the surrogate model becomes possible.

% しかし，機械学習アルゴリズム毎にハイパーパラメータ空間は異なる場合があるため，単純には機械学習アルゴリズム毎に独立な確率モデルを構築する必要がある\cite{nguyen2020bayesian}．
% この方法では観測値$y$の情報は機械学習アルゴリズム毎に独立となり，個々の機械学習アルゴリズム全てに十分な数の観測を得なければ$y$の予測は不確実性の高い予測となってしまう．
% この問題を避けるため，機械学習アルゴリズムのハイパーパラメータ空間を結合し，結合された単一の空間上でベイズ最適化を実行する方法が考えられる\cite{hutter2011sequential}．
% しかし，全ての機械学習アルゴリズムのハイパーパラメータ空間の結合は大規模な空間となることから，最適化の効率が低下してしまう．

% しかし，一般に機械学習アルゴリズム毎にハイパーパラメータ空間は異なるため，単純には機械学習アルゴリズム毎に独立な代理モデルを構築する必要がある\cite{nguyen2020bayesian}．
In general, since different ML algorithms have different HP spaces, different surrogate models are usually required to estimate for each one of ML algorithms \cite{nguyen2020bayesian}.
%
% この方法では観測値の情報は機械学習アルゴリズム毎に独立となり（観測情報の分断化），個々の機械学習アルゴリズム全てに十分な数の観測を得なければ探索を効率的に行えない可能性がある．
In this approach, each surrogate model is independently learned and information of observations is not shared across ML algorithms. 
Because of this observation separation, this approach requires sufficient observations for all candidate ML algorithms to build an effective surrogate model.
%
% 一方で，異なる機械学習アルゴリズムのハイパーパラメータ空間を結合し，結合された単一の空間上でベイズ最適化を実行する方法が考えられる\cite{hutter2011sequential}．
On the other hand, applying BO in the joint (concatenated) space of different HP spaces \cite{hutter2011sequential} has been also studied.
%
% しかし，全ての機械学習アルゴリズムのハイパーパラメータ空間を結合すると高次元空間となり，精度の高い代理モデルの構築が難しくなる（高次元化の問題）．
However, the joint space becomes high dimension and the default value setting is required for `non-active' ML algorithms whose justification is unclear.
Further details of related work is discussed in Section~\ref{sec:related_works}.
% 既存の研究については\ref{sec:related_works}節でより詳しく述べる．

% 本稿では，機械学習アルゴリズム毎に異なるハイパーパラメータ空間で観測したデータを共通の潜在空間に写像することで共有し，ベイズ最適化を実行する方法を提案する
In this study, we propose a BO based CASH solver in which a surrogate model is learned in a shared latent space.
The basic idea is to embed observations obtained by different ML algorithms into a common latent space.
%
% データを共有することで観測情報の分断化を防ぎ代理モデルの改善をはかり，また，単一の低次元潜在空間を使うことで高次元化の問題を回避する．
This approach enables the surrogate model 1) to be constructed by using information of all candidate ML algorithms, and 2) to avoid the difficulties in the joint space approach such as high dimensionality.
%
% 多くの機械学習アルゴリズムは異なるハイパーパラメータを持つが，それらの多くはいずれも予測モデルの複雑さを何らかの意味でコントロールするものであり，適切な変換を施すことで潜在的に共通する表現を持たせることが可能ではないかという仮定にこのアプローチは基づいている．
Since most of HPs control the model flexibility in different manners, we assume that a common latent representation can be created for HP spaces of many different ML algorithms through an appropriate transformation.  
%
% また，潜在空間に対して過去のデータを利用した事前学習を導入する．
%
% これにより興味のある現在のデータでの観測が少量であっても潜在空間の推定が安定的に行えるようにし，事前知識の活用により探索を効率化する．
We further introduce a pre-training framework for the latent space construction so that the search can be stably accelerated even at the early iterations of BO.

% データで潜在的に共通する表現の発見を可能とし，適切な機械学習アルゴリズムとハイパーパラメータの組み合わせ発見に必要な時間を短縮する．
%
% 計算機実験では，複数のクラス分類問題に対して提案手法と既存手法の比較実験を行い，提案手法の有効性を示す．

% --------------------------------------------------
% Fig: overview
% --------------------------------------------------
\begin{figure*}
 \centering
 % \ig{.7}{./figs-overview.pdf}
 \ig{.9}{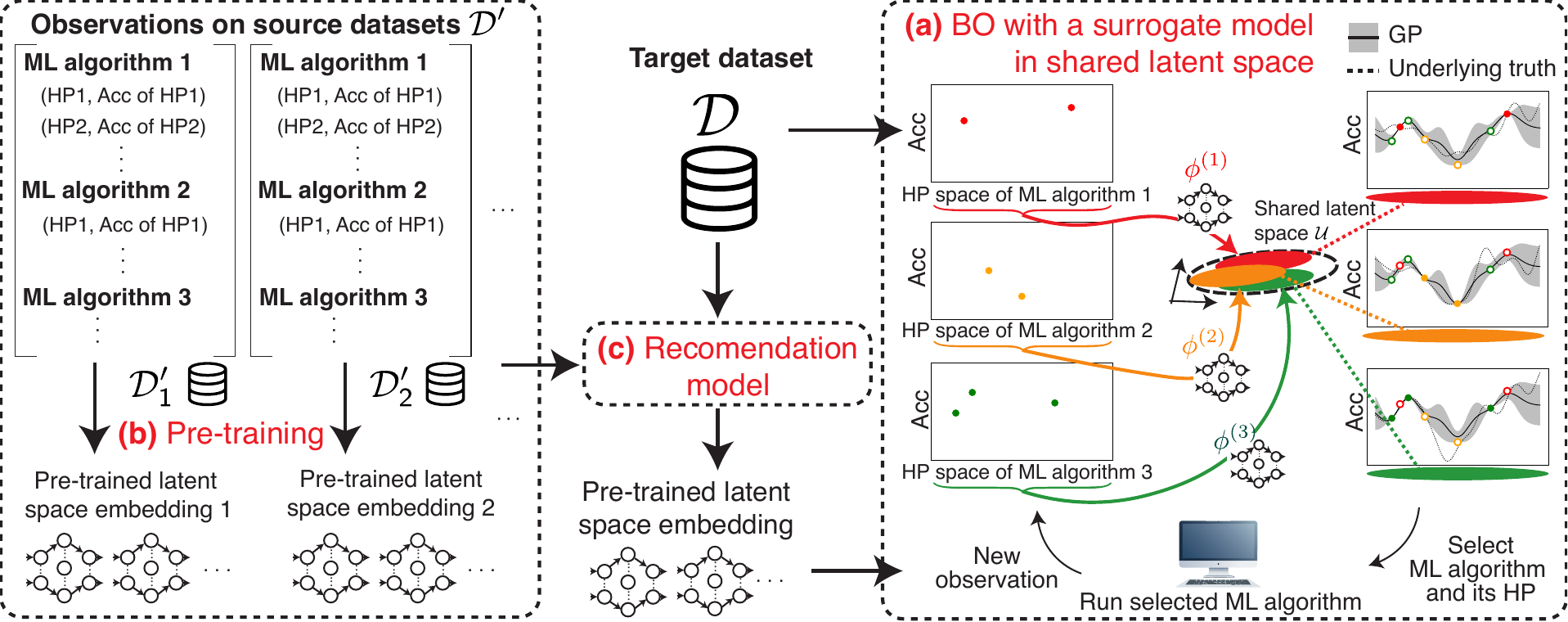}
 \caption{Overview of Proposed Framework.}
 \label{fig:overview}
\end{figure*}

Figure~\ref{fig:overview} illustrates the entire framework of the proposed method, which comprises three main components. 
Each of these components is a major contribution of this paper:
%
% 本研究の貢献を以下に要約する．
\begin{itemize}
 \setlength{\leftskip}{1em}
 \item % 機械学習アルゴリズムとハイパーパラメータの組み合わせの同時最適化に対して，共有潜在空間上の代理モデルに基づくベイズ最適化を提案する．
       The first component is the BO with a surrogate model in the shared latent space for the CASH optimization, illustrated in Fig.~\ref{fig:overview}(a).
       %
       % マルチタスクガウス過程とDeep kernel\cite{wilson16deep}を組み合わせることで，異なるハイパーパラメータ空間上の観測を同時に使って代理モデルが推定できることを示す．
       By combining the multi-task Gaussian process \cite{alvarez2012kernels} and deep kernel \cite{wilson16deep}, we build the surrogate model in such a way that observations in different HP spaces can be shared. 
 \item % 潜在空間推定のための事前学習を提案する．
       The second component is the pre-training of the latent space embedding, illustrated in Fig.~\ref{fig:overview}(b).
       %
       % この際，異なる機械学習アルゴリズムがハイパーパラメータ空間を積極的に共有するような正則化項をドメイン敵対的学習\cite{ganin2016domain}の考え方に基づいて設計する．
       To encourage information sharing among different ML algorithms, we introduce domain adversarial learning \cite{ganin2016domain} as a regularizer in the pre-training, which prevents observations from different ML algorithms from being isolated in the latent space.
 \item The pre-training can be performed for a variety of past datasets, called source datasets, beforehand.
       % 事前学習の有効性はどのようなデータセットで事前学習したかに依存する．
       In the third component, shown as (c) in Fig~\ref{fig:overview}, we construct a ranking model that recommends the best pre-trained embedding model for a given target dataset by using a technique in learning to rank \cite{liu2011learning}.
       %
       % 最適化したい所与のデータセットに対して，過去のどのデータセットによって事前学習した潜在空間を使うべきかランキングモデルで予測するアプローチを提案する．       
       %
       % 計算機実験では，複数のクラス分類問題に対して提案手法と既存手法の比較実験を行い，提案手法の有効性を示す．
\end{itemize}
%
% 提案法の有効性をOpenML\cite{OpenML2013}のデータセットにおけるクラス分類問題を使って実証した．
We empirically demonstrate the effectiveness of the proposed framework through datasets in OpenML \cite{OpenML2013}.
%
% 既存の手法に対する優位性や，事前学習の効果などを経験的に示した．
% Superiority compared with existing approaches and effect of pre-training are demonstrated.

% --------------------------------------------------
\section{Problem Setting} % Combined Algorithm Selection and Hyperparameter Optimization Problem
\label{sec:CASH}

We assume that there exist $M$ candidate machine learning (ML) algorithms 
$\cA = \{ A^{(1)}, \ldots, A^{(M)} \}$, 
% $\cA = \left\{ A^{(1)}, \ldots, A^{(L)} \right\}$, 
in which  the $m$-th ML algorithm 
$A^{(m)}$ has a hyper-parameter (HP) space 
$\*\Lambda^{(m)}$. 
%
% つまり，探索の候補空間はアルゴリズムとハイパーパラメータのペアであり，
% $\Xi = \{ (A^{(i)}, \*\lambda^{(i)}) \mid A^{(i)} \in \cA, \*\lambda^{(i)} \in \*\Lambda^{(i)}, i \in [L] \}$
% と表記する．
This means that our search space consists of a pair of an ML algorithm $A^{(m)}$ and an HP vector $\*\lambda^{(m)}$, denoted as
$\Xi = \{ (A^{(m)}, \*\lambda^{(m)}) \mid A^{(m)} \in \cA, \*\lambda^{(m)} \in \*\Lambda^{(m)}, m \in [M] \}$. 
%
% % また，あるデータセット$\mathcal{D}$が与えられた時，このデータセットを$\mathcal{D}_{\mathrm{train}}$, $\mathcal{D}_{\mathrm{valid}}$の2つに分割する．
A given dataset $\cD$ is partitioned into  
the training dataset $\cD_{\mr{train}}$
and
the validation dataset $\cD_{\mr{valid}}$, respectively.
%
% $A^{(i)}$があるハイパーパラメータ$\*{\lambda}^{(i)} \in \*{\Lambda}^{(i)}$を用いて$\mathcal{D}_{\mathrm{train}}$で学習を行い，$\mathcal{D}_{\mathrm{valid}}$で得られる評価値を$\mathrm{Acc}(A^{(i)}, \*{\lambda}^{(i)}, \mathcal{D}_{\mathrm{train}},\mathcal{D}_{\mathrm{valid}})$とする．
Suppose that 
$\mr{Acc}(A^{(m)}, \*\lambda^{(m)}, \cD_{\mr{train}},\cD_{\mr{valid}})$
is an evaluation score (such as the validation classification accuracy) on
$\cD_{\mr{valid}}$ 
of an ML algorithm $A^{(m)}$ trained by $\cD_{\mr{train}}$ with an HP
$\*\lambda^{(m)}$.
%
% この時，$\mathrm{Acc}(A^{(i)}, \*{\lambda}^{(i)}, \mathcal{D}_{\mathrm{train}},\mathcal{D}_{\mathrm{valid}})$が最大となる$(A^{*}, \*{\lambda}^*) \in \Xi$を発見する問題はCombined Algorithm Selection and Hyperparameter optimization (CASH) problem\cite{thornton2013auto,NIPS2015_11d0e628}と呼ばれ，以下のように書くことができる．
The optimization problem for identifying the pair $(A^{*}, \*{\lambda}^*) \in \Xi$ that maximizes 
$\mr{Acc}(A^{(m)}, \*\lambda^{(m)}, \cD_{\mr{train}},\cD_{\mr{valid}})$
is called Combined Algorithm Selection and Hyperparameter optimization (CASH) problem \cite{thornton2013auto,NIPS2015_11d0e628}:
\begin{align}
 (A^{*}, \*{\lambda}^*) = 
 \argmax_{ (A^{(m)}, \*\lambda^{(m)}) \in \Xi } 
 \mr{Acc}(A^{(m)}, \*{\lambda}^{(m)}, \cD_{\mr{train}},\cD_{\mr{valid}}). 
 \label{eq:CASH}
 % (A^{*}, \*{\lambda}^*) = \argmax_{A^{(m)} \in \mathcal{A}, \*{\lambda} \in \*{\Lambda}^{(m)}} \mathrm{Acc}(A^{(m)}, \*{\lambda}^{(m)}, \mathcal{D}_{\mathrm{train}},\mathcal{D}_{\mathrm{valid}}). \label{eq:CASH}
\end{align}
% 一般に$\mathrm{Acc}$と$(A^{(m)}, \*{\lambda}^{(m)})$の関係を陽に書き下すことは難しく，この問題はブラックボックス最適化問題として扱うことが多い．
In general, it is often difficult to analytically represent a relation between 
$\mr{Acc}$
and
$(A^{(m)}, \*{\lambda}^{(m)})$, 
and therefore, this problem is often regarded as a black-box optimization problem.

% --------------------------------------------------
% \section{提案手法}\label{sec:Proposed}
\section{Proposed Method}
\label{sec:Proposed}

% この章では我々が提案した(\ref{eq:CASH})を解くためのフレームワークを紹介する．
Here, we introduce our proposed framework for efficiently solving \eq{eq:CASH}.
%
% Section~\ref{ssec:shared-space-BO}では，Fig.~\ref{fig:overview}(a)に示した潜在空間上のsurrogateによるBOについて述べる．
In Section~\ref{ssec:shared-space-BO}, we describe our BO with a latent space surrogate model (Fig.~\ref{fig:overview}(a)). 
%
% 次に，Section~\ref{ssec:pre-train}では，Fig.~\ref{fig:overview}(b)に示した事前学習について述べる．
The pre-training (Fig.~\ref{fig:overview}(b)) of the latent space embedding is described in Section~\ref{ssec:pre-train}.
%
% 最後に，Fig.~\ref{fig:overview}(c)で示したranking modelについてSection~\ref{ssec:ranking-model}で述べる．
The ranking model (Fig.~\ref{fig:overview}(c)) for the selection of a pre-trained model is shown in Section~\ref{ssec:ranking-model}. 
The algorithm of the proposed method is shown in Algorithm~\ref{alg:proposed-method}, whose details are described throughout this section.

% 最適化問題(\ref{eq:CASH})は異なる入力変数$\*\lambda^{(m)}$を持つ$L$個のブラックボックス関数から構成されている．
%
% 提案するベイズ最適化では，個々の$\*\lambda^{(m)}$を別々に扱うのではなく共通の潜在空間に移すことで効率的に(\ref{eq:CASH})を解く．

% --------------------------------------------------
% Algorithm: Proposed method
% --------------------------------------------------
\begin{algorithm}[t]
 \caption{Proposed Method}
 \label{alg:proposed-method}

% \SetKwInOut{Input}{Input}
\SetKwInOut{Require}{Require}

\Require{
    Initial observations $\cO$, 
    Max iteration $T$,   
    Meta-features $\*x_s^{\mr{meta}}$ and pre-trained MLPs for source datasets $\{ \cD^\prm_s \}_{s \in [S]}$,     
    % Meta-features $\*x_s^{\mr{meta}}$ for source datasets $\cD^\prm_s$,     
    % Pre-trained MLPs $\{ \phi^{(m)} \}_{m \in [M]}$, 
    Ranking model $f_{\mr{rank}}$
}

% (Beforehand) Perform pre-training of MLPs for latent space embedding $\{ \phi^{(m)} \}_{m \in [M]}$ \Comment{Pre-training (Fig.~\ref{fig:overview}(b))}

Compute meta-feature $\*x^{\mr{meta}}$ for target dataset $\cD$

Select the best source dataset $\cD^\prm_{s_{\rm{best}}}$ by ranking model $f_{\mr{rank}}$:  
\[
    s_{\mr{best}} \leftarrow \argmax_{s \in [S]} f_{\mr{rank}}(\*x^{\mr{meta}}, \*x_s^{\mr{meta}}) 
\] \Comment{Recommendation of pre-trained model (Fig.~\ref{fig:overview}(c))}

Set $\{ \phi^{(m)} \}_{m \in [M]}$ as MLPs trained beforehand by $\cD^\prm_{s_{\mr{best}}}$  

\For(\CommentHere{Main loop of BO (Fig.~\ref{fig:overview}(a))}){$t < T$}{ 
    Optimize MTGP with deep kernel via regularized marginal likelihood \eq{eq:loss_func}

    Select next observation by EI:  
    \[
        (A_{\mathrm{next}}, \*\lambda_{\mr{next}}) \leftarrow \max_{(A^{(m)}, \*\lambda^{(m)}) \in \Xi} a(A^{(m)}, \*\lambda^{(m)}) 
    \]

    Compute validation accuracy:
    \[
        y_{\mr{next}} \leftarrow \mr{Acc}(A_{\mr{next}}, \*\lambda_{\mathrm{next}}, D_{\mr{train}}, D_{\mr{valid}})
    \]

    \If{ $y_{\mr{next}}$ \rm{is current best} }{
        $(A_*,{\lambda}_*) \leftarrow (A_{\mr{next}},{\lambda}_{\mr{next}})$\;
    }
    Update observations:  
    $O \leftarrow O \cup \{(A_{\rm{next}}, {\lambda}_{\rm{next}}, y_{\rm{next}})\}$

}

\Return{$(A_*,{\lambda}_*)$}
\end{algorithm}
\subsection{Bayesian Optimization on Shared Latent Space}
\label{ssec:shared-space-BO}

% --------------------------------------------------
% \subsubsection{機械学習アルゴリズムのHPの共有}
% \subsubsection{共有潜在空間でのマルチタスクガウス過程}
\subsubsection{Multi-task Gaussian Process on Latent Space}
\label{sssec:MTGP}

% 機械学習アルゴリズム$A^{(m)}$とハイパーパラメータ$\*{\lambda}_j^{(m)}$に対する観測値を$y_j^{(m)} = \mathrm{Acc}(A^{(m)}, \*{\lambda}_j^{(m)}, \mathcal{D}_{\mathrm{train}},\mathcal{D}_{\mathrm{valid}})$とする．
For $n \in \NN$, suppose that 
$y_n^{(m)} = \mr{Acc}(A^{(m)}, \*{\lambda}_n^{(m)}, \cD_{\mr{train}},\cD_{\mr{valid}})$
is an observation for a pair $(A^{(m)}, \*{\lambda}_n^{(m)})$.
%
% このとき，$A^{(m)}$の観測済みデータ数を$n^{(m)}$とすると，全体の観測データ集合$\mathcal{O}$は次のように定義される．
The entire set of observations is written as 
\begin{align}
 \cO &= \bigcup_{m \in [M]} O^{(m)}, 
 \label{eq:observed_data}
 % \\
 % \mathcal{O} &= \left\{ S^{(m)}\right\}_{i=1}^{L}, \label{eq:observed_data}\\
 % \mathcal{S}^{(m)} &= \left\{(A^{(m)}, \*{\lambda}_{j}^{(m)}, y_j^{(m)}) \right\}_{j=1}^{n^{(m)}}. \label{eq:each_observed_data}
\end{align}
where 
$\cO^{(m)} = \{(A^{(m)}, \*\lambda_n^{(m)}, y_n^{(m)}) \}_{n \in [N_m]}$
is a set of observations for the $m$-th ML algorithm $A^{(m)}$, and $N_m$ is the number of observations from $A^{(m)}$.
% ただし，
% $\cS^{(m)} = \left\{(A^{(m)}, \*\lambda_j^{(m)}, y_j^{(m)}) \right\}_{j \in [n^{(m)}]}$
% は$A^{(m)}$に対する観測データ集合である．

% ここで，$A^{(m)}$のハイパーパラメータ$\*{\lambda}_j^{(m)} \in \*{\Lambda}^{(m)}$を以下のように潜在変数$\*{u}_j^{(m)} \in \mathcal{U}$へと変換することを考える．
We consider embedding an HP vector
$\*{\lambda}_n^{(m)} \in \*{\Lambda}^{(m)}$
of
$A^{(m)}$
into a shared latent space $\cU$:
\begin{align}
 \*u_n^{(m)} = \phi^{(m)}(\*\lambda_n^{(m)}), 
\end{align}
where 
$\*u_n^{(m)} \in \cU$
is a latent variable 
and
$\phi^{(m)}: \*\Lambda^{(m)} \rightarrow \cU$
is an embedding function defined by a Multi-Layer Perceptron (MLP).
%ただし，
%$\phi^{(m)}: \*\Lambda^{(m)} \rightarrow \cU$
%は潜在空間への埋め込み関数であり，多層パーセプトロン（Multi-Layer Perceptron: MLP）によって構成されているとする．
%%
% 潜在空間の次元数は任意の次元数を設定することができるが，今回の問題において我々は潜在空間の次元数を2次元とした．
% 潜在空間$\cU$は全ての機械学習アルゴリズム$A^{(m)}$に対して共通であるとし，この共有された潜在空間上で確率モデルを構築し，ベイズ最適化による$(A^*,\*{\lambda}^*)$の探索を行う．
The latent space $\cU$ is shared by all ML algorithms $A^{(m)}$.

\begin{figure}[t]
 \centering
 \includegraphics[width=0.9\linewidth]{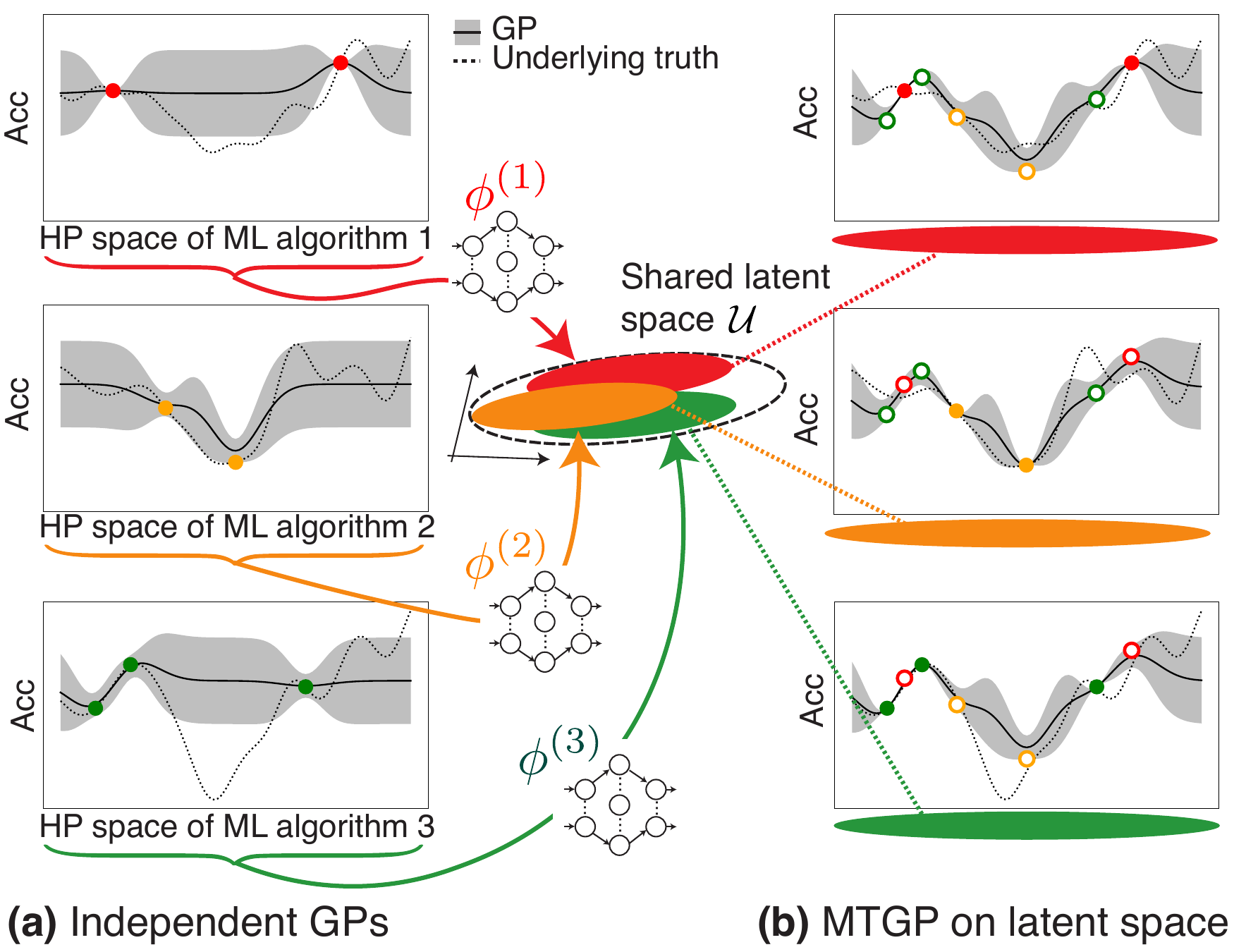}
 \caption{
 % 潜在空間上のマルチタスクガウス過程回帰の概念図．
 Schematic illustration of MTGP on latent space. 
 (a) Independent GPs are fitted to each ML algorithm separately. 
 (b) The MTGP is fitted in the latent space, by which information from different ML algorithms are shared.
 % の青領域は各入力に対する不確実性の高さを表している．
 % 提案手法の概要図．図中の青領域は各入力に対する不確実性の高さを表している．
 % 候補機械学習アルゴリズムとして決定木，ニューラルネットワーク，SVMを用意した．
 % 図の左では各機械学習アルゴリズムは自身の観測データのみで予測を実行しているため，不確実性の高い予測となっている．
 % 各機械学習アルゴリズムのハイパーパラメータ空間を共通の潜在空間$\cU$上へと写像し，$\cU$上でマルチタスクガウス過程回帰を推定している．
 }
  \label{fig:proposed}
\end{figure}

% ベイズ最適化では確率モデルにガウス過程回帰\cite{rasmussen:williams:2006}がよく利用される．
In Bayesian optimization (BO), the Gaussian process (GP) \cite{rasmussen:williams:2006} is typically used as a surrogate model of the objective function. 
% しかし，機械学習アルゴリズム毎に異なる多層パーセプトロン(MLP)の使用しているため，$y_j^{(m)} = y_{j'}^{(i')}$となる場合，必ずしも$f_{\*{A^{(m)}}}(\*{\lambda}_j^{(m)}) = f_{\*{A^{(i')}}}(\*{\lambda}_{j'}^{(i')})$とはならず，僅かなズレが生じると考えられる．
% したがって，我々はそのズレを補完するために，確率モデルにマルチタスクガウス過程回帰\cite{NIPS2007_66368270}を使用する．
%
% ここでは，$i = 1, \ldots, L$のアルゴリズム$A^{(m)}$が作る$\mathrm{Acc}$の$L$個の曲面を別々に扱うのではなく，潜在空間上のマルチタスクガウス過程回帰\cite{NIPS2007_66368270, alvarez2012kernels}として統合的に表現する．
Instead of separately modeling $M$ different $\mr{Acc}$ created by $A^{(m)}$ for 
$m = 1, \ldots, M$, 
we employ a multi-task GP (MTGP) \cite{NIPS2007_66368270, alvarez2012kernels} in which $M$ ML algorithms are seen as correlated $M$ tasks. 
%
% ある
% $y_j^{(m)}$
% が潜在空間上の未知関数$f_{A^{(m)}}$により
We assume that 
$y_n^{(m)}$
can be represented by a function $f_{A^{(m)}}$ on latent space:
\begin{align*}
 y_n^{(m)} 
 & = f_{A^{(m)}}(\phi^{(m)}(\*\lambda_n^{(m)})) + \veps
 = f_{A^{(m)}}(\*u_n^{(m)}) + \veps,
\end{align*}
% として定まっているとする．
%
% ただし，
% $\veps \sim \cN(0, \sigma_{\rm noise}^2)$
% は独立なノイズ項とする．
where
$\veps \sim \cN(0, \sigma_{\rm noise}^2)$
is an independent noise term.
%
% 関数
% $f_{A^{(m)}}(\*u_j^{(m)})$
% を以下のマルチタスクガウス過程回帰によってモデル化する（図\ref{fig:proposed}）．
The MTGP, defined on the latent space, represents
$f_{A^{(m)}}(\*u_n^{(m)})$
as follows:
\begin{align}
 f_{A}(\*u) &\sim
 \cG\cP \rbr{\mu_0(\*u), k((\*u,A), (\*u^\prime, A^\prime))},
 % f_{A^{(m)}}(\*{\lambda}_j^{(m)}) &\sim \mathcal{N}\left(m(\*{u}_j^{(m)}), k((\*{u}_j^{(m)},i), (\*{u}_j^{(m)},i))\right). 
% \\  k((\*{u}_j^{(m)},i), (\*{u}_j^{(m)},i)) &= k_{\mathrm{input}}(\*{u}_j^{(m)}, \*{u}_j^{(m)}) * k_{\mathrm{task}}(i,i),
 \label{eq:MTGP}
\end{align}
% ただし，
% $\*u \in \cU$であり, 
% $m: \cU \rightarrow \RR$
% は潜在空間上の事前平均関数，
% $k((\*u,A), (\*u^\prime, A^\prime))$
% は
% $\cU \times \cA$
% のペアに対するカーネル関数である．
where 
$\*u \in \cU$ is an input in the latent space,
$\mu_0: \cU \rightarrow \RR$ 
is a prior mean function, and 
$k((\*u,A), (\*u^\prime, A^\prime))$
is a kernel function for a pair
$\cU \times \cA$.
Figure~\ref{fig:proposed} shows an illustration of this MTGP.
In the literature of MTGPs, a variety of approaches to building
$k((\*u, A), (\*u^\prime, A^\prime))$
have been discussed, by which relations among tasks can be controlled.
% マルチタスクガウス過程回帰では様々な方法で入力とタスクの結合カーネル
% $k((\*u, A), (\*u^\prime, A^\prime))$
% を定義する方法が考えられてきたが，最もシンプルなアプローチは
% $k((\*u, A), (\*u^\prime, A^\prime)) = k(\*u, \*u^\prime) \times k(A, A^\prime)$
% のように分解できる形を用いることである．
%
% 今回はLinear Model of Coregionalization (LMC) \cite{alvarez2012kernels}と呼ばれるマルチタスクカーネルを用いる．
We employ the well-known linear model of coregionalization (LMC) \cite{alvarez2012kernels} kernel, which captures task dependencies using a low-rank plus diagonal covariance matrix across tasks.

% 事後分布の計算は標準的なガウス過程回帰と同様である．
The posterior distribution can be calculated by the same procedure as the standard GP.
%
% いま，アルゴリズム$A^{(m)}$のあるハイパーパラメータ$\*\lambda^{(m)}$に対して
% $\*u^{(m)} = \phi^{(m)}(\*\lambda^{(m)})$
% とし，この点の予測分布を考えるとする．
Consider a predictive distribution for an HP
$\*\lambda^{(m)}$ 
of an ML algorithm $A^{(m)}$, for which the latent variable is
$\*u^{(m)} = \phi^{(m)}(\*\lambda^{(m)})$.
%
% 訓練データの，ハイパーパラメータの潜在空間上での点とアルゴリズムのペアの集合を
% $U = \bigcup_{i \in [L]} \{ (\*u_j^{(m)}, A^{(m)}) \}_{j \in [n^{(m)}]}$
% とし，
Let
$U = \bigcup_{m \in [M]} \{ (\*u_n^{(m)}, A^{(m)}) \}_{n \in [N_m]}$
be a set of pairs of an embedded latent variable and an ML algorithm in the observed set $\cO$, and
\begin{align*}
 % \*k &= k((\*\lambda^{(m)}, A^{(m)}), U)
 \*k &= k(U, (\*u^{(m)}, A^{(m)}))
 \in \RR^{|\cO|}, 
 \\
 \*K &= 
 k(U, U)
 \in \RR^{|\cO| \times |\cO|}, 
\end{align*}
% はそれぞれ$U$の要素をカーネル関数に代入して得られる値を並べたベクトルと行列とする．
be a vector and a matrix created by substituting each element of $U$ into the kernel function $k$.
%
% このとき観測データ(\ref{eq:observed_data})を得た後の事後分布は以下となる．
Then, the posterior given observations (\ref{eq:observed_data}) is 
\begin{align*}
 f_{A^{(m)}}(\*u^{(m)}) \mid \cO &\sim 
 \cN \rbr{ \mu(\*u^{(m)}), \sigma^2(\*u^{(m)}) },  
\end{align*}
where 
% ただし，
\begin{align*}
 \mu(\*u^{(m)}) &= 
 \*k^\top 
 (\*{K} + \sigma_{\rm noise}^2\*I)^{-1} (\*y - \*\mu_0) + \mu_0(\*u^{(m)}), 
 \\
 \sigma^2(\*u^{(m)}) &= 
 k((\*u^{(m)}, A^{(m)}), (\*u^{(m)}, A^{(m)})) - \*{k}^{\top}(\*{K}+ \sigma_{\rm noise}^2 \*I)^{-1} \*k,
\end{align*}
% であり$\*m$と$\*y$はそれぞれ$U$に対応する$m(\*u_j^{(m)})$と$y_j^{(m)}$を並べたベクトルである．
and $\*\mu_0$ and $\*y$ are vectors concatenating 
$\mu_0(\*u_n^{(m)})$ 
and 
$y_n^{(m)}$, 
respectively. 
We will discuss how to construct the prior $\mu_0$ in Section~\ref{sssec:pre-train-formulation}.

% --------------------------------------------------
% Figure: MTGP
% --------------------------------------------------

% 図\ref{fig:proposed}は各機械学習アルゴリズムで独立に予測を行う場合と潜在空間上のマルチタスクガウス過程回帰を用いる場合を概念図で比較している．
Figure~\ref{fig:proposed} compares independent GPs for each ML algorithm (left) and the MTGP through the latent space embedding (right).
%
% 図中左側では各機械学習アルゴリズムが自身の観測データのみでガウス過程を推論するため，少量の観測データしか個々のモデル構築に用いることができない．
In the left three plots, each GP can only use their own observations $\cO^{(m)}$.
%
% これに対して，潜在空間上での予測を行うことで機械学習アルゴリズム$A^{(m)}$の$\*\lambda_j^{(m)}$に対する予測を行うとき，$A^{(m)}$以外の機械学習アルゴリズムが観測したデータも用いて予測が可能となる．
On the other hand, the MTGP can use all observations $\cO$ to give a prediction for any $\*\lambda_n^{(m)}$ of any $A^{(m)}$.

% --------------------------------------------------
% \subsubsection{パラメータ最適化}
\subsubsection{Parameter Optimization}
\label{sssec:parameter-optimization}

% 提案手法のマルチタスクガウス過程回帰モデルが持つハイパーパラメータ（例えばカーネル関数のlength scaleなど）と$\*\Lambda^{(m)}$を変換するMLP $\phi^{(m)}$の持つパラメータをまとめて$\*\Theta$とすると，$\*\Theta$は以下の損失関数を最小化することで最適化が行われる．
Let 
$\*\Theta$
be all the parameters including the parameters of the MTGP (such as the kernel length scale) and the weight parameters of the MLP 
$\phi^{(m)}$.
% 提案手法のマルチタスクガウス過程回帰モデルが持つパラメータと$\*{\Lambda}^{(m)}$を変換するMLPの持つパラメータをまとめて$\*{\Theta}$とすると，$\*{\Theta}$は以下の損失関数を最小化することで最適化が行われる．
%
We optimize $\*\Theta$ by minimizing 
\begin{align}
 \cL(\*\Theta) 
 &=  
 -\left(- \frac{1}{2}
 (\*y - \*\mu_0)^\top \*C^{-1} (\*y - \*\mu_0)
 % (\*y - \*\mu_0)^\top (\*K + \sigma_{\rm noise}^2 \*I)^{-1} (\*y - \*\mu_0)
 - \frac{1}{2}\log \left| \*C \right|\right) 
 % - \frac{1}{2}\log \left| \*K + \sigma_{\rm noise}^2 \*I \right|\right) 
 \notag \\
 & \qquad
 + \alpha \sum_{m \in [M]} R(\*\theta^{(m)}), 
 \label{eq:loss_func}
\end{align}
% ただし，
% $\*K_\Theta = \*K + \sigma_{\rm noise}^2 \*I$
% とし，
% $\*\theta^{(m)}$
% はMLP $\phi^{(m)}$の持つパラメータベクトル，
% $R(\*\theta^{(m)})$
% は$\*\theta^{(m)}$の正則化項，
% $\alpha$は正則化係数とする．
where 
% $\*K_\Theta = \*K + \sigma_{\rm noise}^2 \*I$,
$\*C = \*K + \sigma_{\rm noise}^2 \*I$,
$\*\theta^{(m)}$ is the weight parameters of
$\phi^{(m)}$, 
$R(\*\theta^{(m)})$ 
is a regularization term for
$\*\theta^{(m)}$, 
and
$\alpha$ 
is a regularization parameter.
%
% Note that $\*C$ and $\*\mu_0$ depend on $\*\Theta$. 
%
% 式(\ref{eq:loss_func})の第1項はガウス過程回帰モデルの対数周辺尤度\cite{rasmussen:williams:2006}を表している．
The first term in (\ref{eq:loss_func}) is the (negative) marginal likelihood of the MTGP. 
%
% このように$\*\Theta$を最適化することで，全ての$A^{(m)}$に対して（周辺尤度の意味で）当てはまりの良いガウス過程回帰及び潜在空間を構築することができる．
By minimizing (\ref{eq:loss_func}), we can estimate the MTGP hyper-parameters and latent space embedding, simultaneously.
%
% カーネル関数中にMLPが含まれているため，この問題はdeep kernel learning\cite{wilson16deep}の一種であると解釈できる．
In this formulation, the kernel function including MLPs can be seen as a deep kernel \cite{wilson16deep}.

% ベイズ最適化は繰り返し初期においては小サンプルであるため，MLPを直接安定的に学習することは難しい可能性がある．
% Because of possible small observations in the early iterations of BO, learning the MLP can be unstable.
Due to limited observations in the early iterations of BO, training the MLPs may be unstable.
%
% そこで，\ref{ssec:pre-train}節でMLPの事前学習を導入する．
To mitigate this issue, we will introduce a pre-training of the MLPs in Section~\ref{ssec:pre-train}.
%
% これにより，異なる機械学習アルゴリズムのハイパーパラメータ空間に対して，最適化が容易で，かつ互いの情報が共有できるような潜在空間を構成する$\phi^{(m)}$を予め推定しておく．
% This enables us to construct 
% In this pre-training phase, we estimate
% $\phi^{(m)}$ 
% that enables us to perform easier optimization and to effectively share information across tasks. 
%
% 式(\ref{eq:loss_func})の正則化項$R(\*\theta^{(m)})$は事前学習済みのMLPからの差分に対して罰則を導入することで，小サンプル化での過適合の発生を防止する．
The regularization term $R(\*\theta^{(m)})$ in (\ref{eq:loss_func}) penalizes the deviation from the pre-trained parameters, by which the over-fitting under small observations is inhibited: 
\begin{align}
 R(\*\theta^{(m)}) = \frac{1}{K_m} \left\|\*\theta^{(m)} - \*\theta_{\rm pre-trained}^{(m)} \right\|_2^2,
 \label{eq:regularization}
\end{align}
% ただし，$N_i$はMLP $\phi^{(m)}$の持つパラメータ$\*\theta^{(m)}$の次元数，
% $\*\theta_{\rm pre-trained}^{(m)}$
% は事前学習されたパラメータベクトルを表している．
where 
$\*\theta^{(m)}$ is the weight parameter vector of the MLP $\phi^{(m)}$ whose dimension is $K_m$, and 
$\*\theta_{\rm pre-trained}^{(m)}$
is the corresponding pre-trained vector.
%
% つまり，(\ref{eq:regularization})の正則化項は事前学習によって学習された$\phi^{(m)}$を維持するよう学習が行われる\cite{kirkpatrick2017overcoming}．
Therefore, the regularization
(\ref{eq:regularization})
tries to maintain the pre-trained $\phi^{(m)}$
\cite{kirkpatrick2017overcoming}.

% --------------------------------------------------
% \subsubsection{次の観測データの決定方法}
\subsubsection{Selecting Next Observation}
\label{sssec:acq}

% 潜在空間およびガウス過程回帰の学習が終了後，次に実験を行う機械学習アルゴリズムとハイパーパラメータ
% $(A_{\mathrm{next}}, \*{\lambda}_{\mathrm{next}}) \in \Xi$
% の決定を行
Based on the estimated MTGP, we determine a pair
$(A_{\mathrm{next}}, \*{\lambda}_{\mathrm{next}}) \in \Xi$
that we evaluate $\mr{Acc}$ next.
%
% この組み合わせの決定方法として，通常のベイズ最適化と同様に獲得関数最大化を行う．
In BO, a next observation is determined by the acquisition function maximization: 
\begin{align}
 (A_{\mathrm{next}}, \*{\lambda}_{\mathrm{next}}) = 
 \argmax_{( A^{(m)}, \*\lambda^{(m)} ) \in \Xi} \
 a(A^{(m)}, \*\lambda^{(m)}). 
 \label{eq:acquisition_function}
\end{align}
% ただし，$a$はベイズ最適化の獲得関数であり，全ての候補$\rbr{ A^{(m)}, \*\lambda^{(m)} }$の中から$a$の値が最大となるペアを選択する．
where $a$ is an acquisition function. 
We here employ the standard Expected Improvement (EI) \cite{brochu2010tutorial}, though any acquisition function studied in BO is applicable:
% 今回は$a$に，標準的な期待改善量（Expected Improvement: EI）\cite{brochu2010tutorial}を用いる．
\begin{align}
 a(A^{(m)}, \*\lambda^{(m)})
 = 
 \EE\sbr{
 \max \cbr{
 f_{A^{(m)}}( \phi^{(m)}( \*\lambda^{(m)} ) ) - y_{\rm best}, 0
 }
 }, 
 \label{eq:EI}
\end{align}
% ただし，$y_{\rm best}$はこれまでの全ての観測データの中で最も大きな$y_j^{(m)}$とする．
where 
$y_{\rm best}$
is the maximum observed value until the current iteration.

% 獲得関数最大化(\ref{eq:acquisition_function})はそれぞれの$A^{(m)}$ごとに$\*\lambda^{(m)}$を最大化して，最後に最も大きな値を持つ
% $\rbr{ A^{(m)}, \*\lambda^{(m)} }$
% のペアを選べばよい．
To solve the acquisition function maximization 
(\ref{eq:acquisition_function}), 
we simply maximize $\*\lambda^{(m)}$ for each $A^{(m)}$, and select 
$( A^{(m)}, \*\lambda^{(m)} )$
that maximizes $a$.
%
% そのため，既存のベイズ最適化で用いられる獲得関数最大化の手法をそのまま適用できる．
For each maximization, any optimization algorithm (such as DIRECT \cite{Jones1993-Lipschitzian} that is often used in BO) is applicable.
%
% ハイパーパラメータ空間の入力は連続変数，離散変数の両方が混在していることが多いため，ここでは\cite{hutter2011sequential}が提案しているヒューリスティックな最適化をベースにした方法を採用する．
We employ a heuristic local optimization algorithm \cite{hutter2011sequential} based approach that is applicable to an HP space containing both continuous and discrete variables (see Appendix~\ref{app:acquisition_function} for detail).

%
% この最適化手法の詳細はAppendix~\ref{app:acquisition_function}に記載する．

% 獲得関数最大化の方法として勾配法の利用が考えられるが，ハイパーパラメータ空間の入力は連続変数，離散変数の両方が混在していることが多いため，勾配法の活用では離散変数の最適化が難しい．
% 遺伝的アルゴリズムは離散変数，連続変数の両方を扱うことは可能だが，1度の最適化に多くの時間が必要である．
% また，提案手法では機械学習アルゴリズム毎に獲得関数計算を行うため，機械学習アルゴリズムの数だけ遺伝的アルゴリズムによる獲得関数最大化の実行は非効率的である．
% そこで我々は\cite{hutter2011sequential}を参考に獲得関数最大化を行なった．
% この最適化は遺伝的アルゴリズムと比べても少量の時間で最適化を行うことが可能であり，離散変数，連続変数の両方が混在している探索空間に対しても最適化を実行することが可能である．
% この最適化手法の詳細はAppendixに記載する．

% --------------------------------------------------
% \subsection{共有潜在空間の事前学習}
\subsection{Pre-training for Shared Latent Space}
\label{ssec:pre-train}

% \ref{ssec:shared-space-BO}節で説明したように，提案法は候補となる機械学習アルゴリズムの観測データを共有する潜在空間上でCASH problem最適化を行う．
As shown in Section~\ref{ssec:shared-space-BO}, our proposed method considers the CASH problem through the latent space shared by all candidate ML algorithms.
%
% この方法は潜在空間の構築が最も重要だが，逐次的に観測データが増加するベイズ最適化では繰り返し初期においては少量の観測データで潜在空間の学習を行わなければならない場合がある．
In this approach, constructing an appropriate latent space is obviously important.
%
% この問題を解消するために，潜在空間構築に対して事前学習を導入する．
We introduce a pre-training approach so that effective latent space can be used even with small amount of observations $\cO$.

% --------------------------------------------------
% \subsubsection{アルゴリズム間の対応関係の事前学習}
% \subsubsection{事前学習の定式化}
\subsubsection{Loss Function for Pre-training}
\label{sssec:pre-train-formulation}

% 3章で説明したように，我々は候補となる機械学習アルゴリズムの観測データを共有することができる潜在空間上でCASH problem最適化を行う．
% この方法は潜在空間の構築が最も重要だが，逐次的に観測データが増加するこの問題では少量の観測データで潜在空間の学習を行わなければならない．
% しかし，少量の観測データでは異なる機械学習アルゴリズム間の関係を捉えることが難しく，誤った共有が行われる潜在空間
% (例えば，$(A^{(m)},\*{\lambda}_j^{(m)})$に対する予測精度$y_j^{(m)}=0.45$，$(A^{(i')},\*{\lambda}_{j'}^{(i')})$に対する予測精度$y_{j'}^{(i')}=0.65$と予測精度が大きく異なるデータが共有されるような潜在空間，)
% が構築されてしまうかもしれない．
% この問題を解消するために，我々は潜在空間に対して事前学習を導入する．

% 事前学習では適切な機械学習アルゴリズムとハイパーパラメータの組み合わせを発見したいターゲットデータセット$\cD$とは異なるデータセット$\cD^\prm$を用意する．
Suppose that there is an additional `source' dataset $\cD^\prm$ that the user would like to use for the knowledge transfer to the the target dataset $\cD$.
% Suppose that we have an additional `source' dataset $\cD^\prm$ different from the target dataset $\cD$, for which we assume that their $\mr{Acc}$ surfaces are expected to have some similarity.
%
% この$\cD^\prm$に対して，候補となる機械学習アルゴリズムとハイパーパラメータの組み合わせに対する予測精度$\mathrm{Acc}$を事前に大量に観測済みであるとする．
Further, for $\cD^\prm$, the validation evaluation $\mr{Acc}$ for many HPs of each candidate ML algorithm is assumed to be already observed.
It should be noted that these observations for $\cD^\prm$ and the entire process of pre-training are performed beforehand (before we obtain $\cD$).
%
%
% $\mathcal{D}'$に対する観測データ集合$\mathcal{O'}$を次のように定義する．
A set of the observations for $\cD^\prm$ is written as
\begin{align}
 \cO^\prm &= \bigcup_{m \in [M]} \cO^{\prm (m)}, 
 \label{eq:observed_predata}
 % \mathcal{O}' &= \left\{\mathcal{S}^{'(m)}\right\}_{i=1}^L, 
 % \label{eq:observed_predata}\\
 % \mathcal{S}^{'(m)} &= \left\{(A^{(m)}, \*{\lambda}_{j}^{(m)}, y_j^{(m)}) \right\}_{j=1}^{m^{(m)}}.
\end{align}
% ただし，
% $\cS^{\prm (m)} = \left\{(A^{(m)}, \*\lambda_j^{\prm (m)}, y_j^{\prm (m)}) \right\}_{j \in [m^{(m)}]}$
% であり，$m^{(m)}$は$A^{(m)}$の観測データ数を表している．
where 
$\cO^{\prm (m)} = \{(A^{(m)}, \*\lambda_n^{\prm (m)}, y_n^{\prm (m)}) \}_{n \in [N^\prm_m]}$
and $N^\prm_m$ are observations for the $m$-th ML algorithm and its size, respectively.
% is a set of pre-training observations, and $N^\prm_m$ is the number of observations from $A^{(m)}$.
%
% また，$y_{\rm best}^\prm$を$\cO^\prm$における$y_j^{\prm (m)}$の最大値とする．

% 後の最適化が容易になるように以下のような性質を持つ潜在空間の推定を考える．
% To construct the latent space in which the objective function becomes simple, we try to approximate the (standardized) $y_n^{\prm (m)}$ by a quadratic function:
To construct the latent space in which the objective function becomes simple, we try to approximate $y_n^{\prm (m)}$ by a quadratic function:
\begin{align}
 % \mathrm{Standardize}\rbr{ y_{\rm best}^\prm - y_n^{\prm (m)} } \approx 
 y_n^{\prm (m)} \approx - \| \*u_n^{\prm (m)} \|_2^2 + y_{\rm best}^\prm,
 % \| \*u_n^{\prm (m)} - \*c \|_2^2,
 \label{eq:def_quad_approx}
\end{align}
% ただし，
% $\mathrm{Standardize}$
% は$y_j^{\prm (m)}$を$[0,1]$に標準化する関数，
% $\*u_j^{\prm (m)} = \phi^{(m)}( \*\lambda_j^{\prm (m)} )$
% であり，$\*c$は潜在空間上に設定した定数の中心ベクトルとする．
where 
$y_{\rm best}^\prm$
is the maximum of 
$y_n^{\prm (m)}$
in 
$\cO^\prm$.
% , $\mr{Standardize}$ is the standardization function that transforms the range of $y_{\rm best}^\prm - y_n^{\prm (m)}$ into $[0,1]$.
% , and $\*c$ is a constant vector.
%
% 左辺は最大値$y_{\rm best}^\prm$からの差分を標準化した値であり，右辺は潜在空間上に設定した中心$\*c$からの距離の$2$乗である．
% The right hand side is the squared distance from the center parameter $\*c$ in the latent space.
%
% このようにすることで，目的関数が潜在空間上で最適値を$\*c$に持つような単純な$2$次曲面になるべく近づくようにできる．
This approximation means that the latent space is estimated in such a way that the objective function can be represented as a simple quadratic surface, which attains the maximum ($y_{\rm best}^\prm$) at the origin.
Although other more complicated functions (with trainable parameters) can also be used, we employ the quadratic function for simplicity.
%
% 標準化した値を
% $\tilde{y}_n^{(m)} = \mathrm{Standardize}\rbr{ y^* - y_n^{\prm (m)} }$
% とすると，$i$番目のMLP $\phi^{(m)}$の事前学習の目的関数を以下として定式化できる．
By defining 
% $\tilde{y}_n^{(m)} = \mr{Standardize}( y_{\rm best}^\prm - y_n^{\prm (m)} )$,
$\tilde{y}_n^{(m)} = y_{\rm best}^\prm - y_n^{\prm (m)}$,
the objective function of the pre-training of the $m$-th MLP $\phi^{(m)}$ is written as
\begin{align}
 & \cL_m^{\rm (Pre-train)}(\*\theta^{(m)}) 
 % \notag \\
 % & 
 = 
 \sum_{n \in [N^\prm_m]} 
 \frac{1}{N^\prm_m}
 % \rbr{ \tilde{y}_n^{(m)} - \| \phi^{(m)}( \*\lambda_n^{\prm (m)} ) - \*c \|_2^2 }^2. 
 \rbr{ \tilde{y}_n^{(m)} - \| \phi^{(m)}( \*\lambda_n^{\prm (m)} ) \|_2^2 }^2. 
 \label{eq:pre_train_loss1}
\end{align}
% この観測データ集合を用いて，以下の関数の最小化を行う．
% \begin{align}
%   \mathcal{L}_m &= \sum_{i=1}^{L}\sum_{j=1}^{m^{(m)}} \frac{1}{m^{(m)}}\left( d|y_{\max} - y_{j}^{(m)}| - \| f_{A^{(m)}}(\*{\lambda}_{j}^{(A^{(m)})}) -\*{c}\|_2^2\right), \label{eq:pre_train_loss1}
% \end{align}
% ただし，(\ref{eq:pre_train_loss1})における$d$は定数パラメータを表しており，$y_{\max}$は(\ref{eq:observed_predata})の$y_j^{(m)}$の最大値を，$\*{c} \in \RR^D$は定数ベクトルを，$D$は潜在空間の次元数を表している．
% \textcolor{red}{
% (\ref{eq:pre_train_loss1})はハイパーパラメータ$\*{\lambda}_j^{(m)}$を対応する$y_j^{(m)}$に応じて図{\ref{fig:contour}}の二次曲面上へと写像を促す関数である．
% 例えば，$(A^{(m)},\*{\lambda}_j^{(m)},0.8),(A^{(i')},\*{\lambda}_{j'}^{(i')},0.8),(i\neq i')$という観測データが存在する場合，これらのハイパーパラメータ$\*{\lambda}_j^{((m))},\*{\lambda}_{j'}^{((i'))}$は3.1節で説明したMLP$f_{A^{(m)}},f_{A^{(i')}}$を用いて潜在空間上へと写像される．
% この時，$y_{j}^{(m)}=y_{j'}^{(i')}=0.8$であるため，$\*{\lambda}_j^{(A^{(m)})},\*{\lambda}_{j'}^{(A^{(i')})}$はともに図\ref{fig:contour}のAcc$=0.8$の2次局面上へと写像される．
% つまり，(\ref{eq:pre_train_loss1})は$y$の値に応じて各ハイパーパラメータの潜在空間上への写像先を制限するよう学習が行われるため，予測精度が異なるハイパーパラメータ同士の共有を妨げる役割を果たしている．
% }
% また，(\ref{eq:pre_train_loss1})の定数パラメータ$d$,$\*{c}$は次の値に設定した．
% \begin{align*}
%   \*{c} &= \begin{bmatrix}
%     0.5\\
%     0.5
%   \end{bmatrix},\\
%   d &= \frac{0.5}{y_{\max} - y_{\min}},
% \end{align*}
% ただし，$y_{\min}$は(\ref{eq:observed_predata})の$y_{j}^{(m)}$の最小値を表している．
% この値に設定することで，(\ref{eq:observed_predata})の$y_j^{(m)}$の値域によらず，任意の$\*{\lambda}_j^{(m)}$は中心$[0.5 , 0.5]^{\top}$, 定義域$\*{u}_j^{(m)} \in [0,1]^D$上へと写像されるよう学習が行われる．
%
This objective function is a quadratic error of the approximation \eq{eq:def_quad_approx}.
%
% 図\ref{fig:contour}に推定された二次曲面の例を示す．
% Figure~\ref{fig:contour}\red{(a)} shows an example of estimated quadratic surfaces.
%
% このようにして$\phi^{(m)}$を構成することで，ターゲットデータセットにおいても二次曲面に近しい曲面が$\cU$上にできることが期待される．
Because of this pre-training, a similar quadratic surface in the latent space is (approximately) expected for the target dataset $\cD$.
%
% そのため，事前学習を行った際にはマルチタスクガウス過程回帰(\ref{eq:MTGP})の事前平均を\eq{eq:def_quad_approx}の逆変換から以下のように設定する．
Therefore, we set the prior mean of the MTGP (\ref{eq:MTGP}) based on %the inverse transformation of \eq{eq:def_quad_approx}: 
\begin{align*}
 \mu_0(\*u) = 
 %- \mathrm{Standardize}^{-1}\rbr{
 % \| \*u - \*c \|_2^2
 - \| \*u \|_2^2 + y_{\rm best}^\prm.
 %}
 % + y_{\rm best}^\prm, 
\end{align*}
Note that appropriate standardization can be applied to $\mu_0(\*u)$ if the scale adjustment is required in the MTGP on the target dataset.
% where 
% $\mr{Standardize}^{-1}$
% is the inverse transformation of 
% $\mr{Standardize}$.
% ただし，
% $\mathrm{Standardize}^{-1}$
% は標準化の逆変換とする．

% --------------------------------------------------
% Figure: Example of Quadratic Function on Latent Space
% --------------------------------------------------
\begin{figure}[t]
 \centering
 \includegraphics[width=.4\tw]{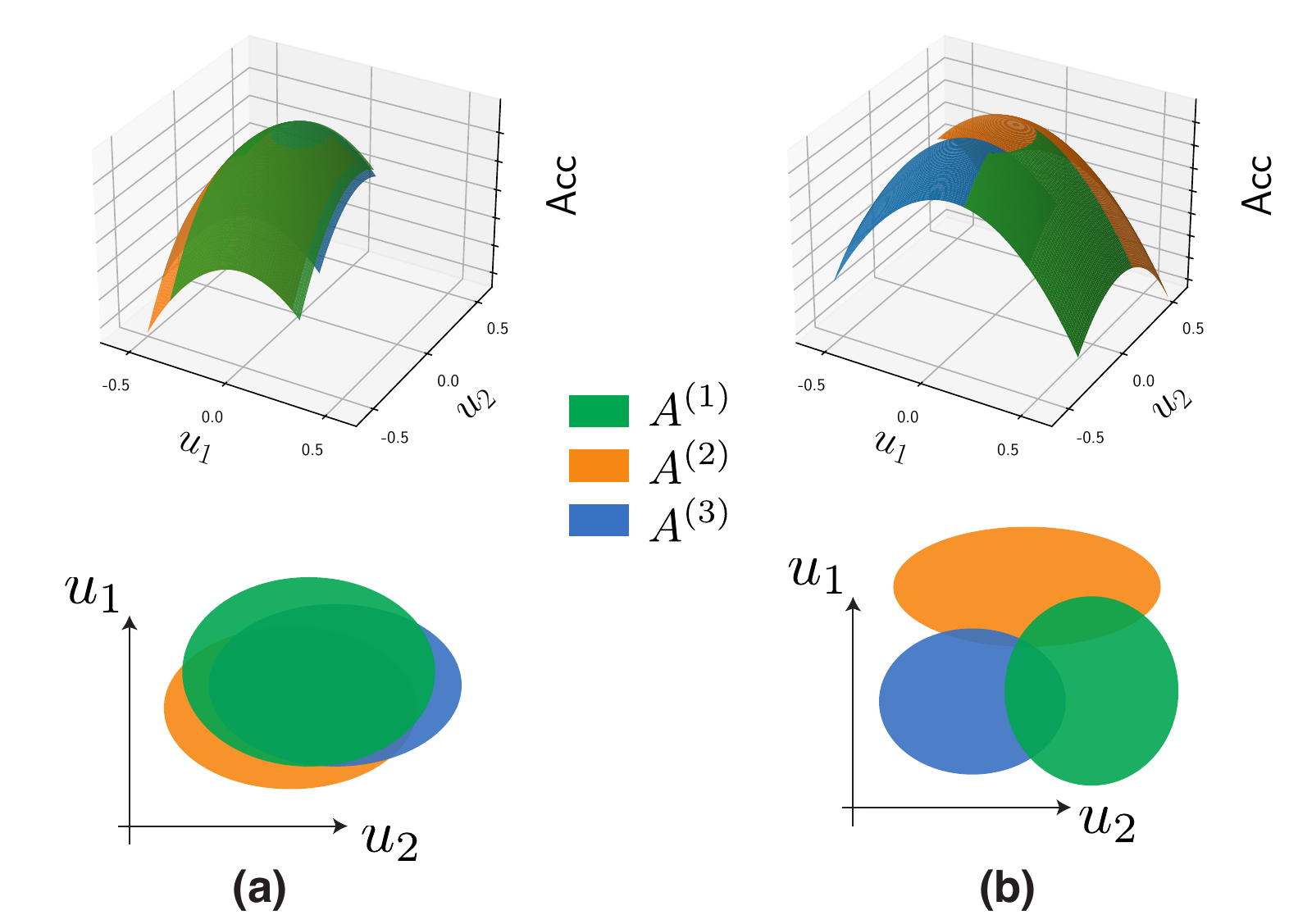}
 \caption{
 Schematic illustrations of quadratic surface prior obtained by pre-training (latent dimension is two).
 Each color corresponds to $\*\Lambda^{(m)}$ embedded in $\cU$.
 % Example of quadratic surface obtained by pre-training (latent dimension is two). 
 % (two dimensional latent space and $\*c = [0.5 \ 0.5]^\top$).
 % 事前学習によって獲得された$2$次曲面の例（潜在空間の次元数$2$，中心は$\*c = [0.5 \ 0.5]^\top$）．
 }
  \label{fig:contour}
\end{figure}

% --------------------------------------------------
% \subsubsection{敵対的正則化}
\subsubsection{Adversarial Regularization}

To construct the latent space in which the information of observations from different ML algorithms $A^{(m)}$ are fully shared, each $\*\Lambda^{(m)}$ should be `overlapped' in $\cU$.
In the illustration of Fig.~\ref{fig:contour}(a), the embedded $\*\Lambda^{(m)}$ is largely overlapped.
On the other hand, in Fig.~\ref{fig:contour}(b), there are almost no shared region by which information sharing cannot be expected.
%
% 事前学習を(\ref{eq:pre_train_loss1})で行うことで，共有潜在空間上での$2$次曲面を構築できる．
%
% しかし，各$\*\theta^{(m)}$は別々に学習されるため，異なる機械学習アルゴリズムのドメイン$\*\Lambda^{(m)}$を$\cU$上の全く異なる領域に写像してしまう可能性もあり，この場合，観測データの情報が上手く共有されない．
%
% そこで，我々は敵対的正則化\cite{ganin2016domain,zhao2018adversarial}の考え方に基づいて，異なる機械学習アルゴリズムのハイパーパラメータ空間が$\cU$内の領域をなるべく共有するような正則化を導入する．
Therefore, we introduce an idea of adversarial regularization \cite{ganin2016domain,zhao2018adversarial}, which encourages each $\*\Lambda^{(m)}$ to be projected onto an overlapped space.
% (as shown in Fig.~\ref{fig:pre-train_learning}).
% (\ref{eq:pre_train_loss1})によって異なる予測精度間での共有を妨げることが可能となったが，この関数は2次曲面上のどの位置に写像されるかを指定することができない．
% したがって，この関数では異なる機械学習アルゴリズムの観測データの共有を促すためには不十分である．
% そこで，我々は敵対的正則化\cite{ganin2016domain,zhao2018adversarial}を参考に異なる機械学習アルゴリズムの観測データの共有を促す関数を導入した．

% いま$N$個の機械学習アルゴリズムのハイパーパラメータ空間$\Lambda^{(m)}$が$\cU$に写像されている．
We embed $M$ domains of the HP space $\*\Lambda^{(m)}$ into $\cU$.
%
% これらが潜在空間$\cU$において互いに分けられないように重なって分布している場合，$\cU$上に構築する予測モデルは異なる機械学習アルゴリズムの観測データ情報をより強く共有すると考えられる．
When those $M$ domains cannot be separated in $\cU$, the surrogate model on $\cU$ should give higher similarity across different ML algorithms (i.e., sharing knowledge of different tasks more strongly).
%
% そこで敵対的学習として，ある与えられた潜在空間上の点$\*u_j^{(m)}$に対して，それがどの機械学習アルゴリズムのハイパーパラメータから作られた点なのかを予測する分類問題を考える（つまり，この場合正解はクラス「$i$」）．
Thus, we consider an adversarial classifier that predicts ``Which ML algorithm is the given $\*u_n^{(m)}$ created from?'', for which the true label is ``$m$'' (the $m$-th ML algorithm $A^{(m)}$).
%
% 例えば，図\ref{fig:contour}の状態では潜在空間上で各アルゴリズムが大きく重なって分布しており分類は困難である．
For example, in Fig.~\ref{fig:contour}(a), ML algorithms are largely overlapped by which the classification is difficult.
%
% このような状態を目指して分類誤差が大きくなるような正則化項を潜在空間の推定に付与する．
Our adversarial regularizer encourages a large classification error during pre-training of the latent space embedding.

% ここでは任意の機械学習アルゴリズムの${}_N C_2$個のペアに対して空間共有を促すためにone-versus-one設定で考える．
To enforce overlaps for any ${}_M C_2$ pairs of ML algorithms, we employ the one-versus-one formulation.
%
% 任意の機械学習アルゴリズムのペア$A^{(m)}, A^{(i^\prm)}$に対して，ある$\*u$が$A^{(m)}$由来である確率を
For a given pair $A^{(m)}$ and $A^{(m^\prm)}$, the probability that $\*u$ is an embedding from $A^{(m)}$ is represented by
\begin{align*}
 p(A = A^{(m)} \mid A \in \{ A^{(m)}, A^{(m^\prm)} \} ) = g_{m,m^\prm}(\*u ; \*\psi_{m,m^\prm}),
\end{align*}
%と表現する．
%
% ただし，
% $g_{i,i^\prm}: \cU \rightarrow [0,1]$
% はパラメータ
% $\*\psi_{i,i^\prm}$
% を持つ$2$値分類ニューラルネットワークとする．
where 
$g_{m,m^\prm}: \cU \rightarrow [0,1]$
is a binary classifier network having a parameter vector
$\*\psi_{m,m^\prm}$.
%
% この$2$値分類の訓練データを事前学習用データセット$\cO^\prm$から
% $\cD^{\rm (Adv)}_{i,i^\prm} = 
% \{ (\*\lambda_j^{\prm (m)}, A^{(m)}) \}_{j \in [m^{(m)}]} 
% \bigcup 
% \{ (\*\lambda_j^{\prm (i^\prm)}, A^{(i^\prm)}) \}_{j \in [m^{(i^\prm)}]}$
% のようにして作ることができる．
This classifier can be learned by using 
$\cD^{\rm (Adv)}_{m,m^\prm} = 
\{ (\*\lambda_n^{\prm (m)}, A^{(m)}) \}_{n \in [N^\prm_m]} 
\bigcup 
\{ (\*\lambda_n^{\prm (m^\prm)}, A^{(m^\prm)}) \}_{n \in [N^\prm_{m^\prm}]}$, 
which is created from $\cO^\prm$.
%
% 各事例$(\*\lambda, A) \in \cD^{\rm (Adv)}_{i,i^\prm}$に対して，$\*\lambda$から$A$を予測するクロスエントロピー損失を考えると以下となる．
For an instance of the classification
$(\*\lambda, A) \in \cD^{\rm (Adv)}_{m,m^\prm}$, 
the cross-entropy loss can be defined, by which we have an objective function for the classifier:
\begin{align}
 & \cL_{m,m^\prm}^{\rm (CE)}(\*\psi_{m,m^\prm} ; \*\theta^{(m)}, \*\theta^{(m^\prm)}) = \notag
 \\
 & \frac{1}{| \cD^{\rm (Adv)}_{m,m^\prm} |} 
 \sum_{ (\*\lambda, A) \in \cD^{\rm (Adv)}_{m,m^\prm} } \bigl\{
 - \II(A = A^{(m)}) \log g_{m,m^\prm}(\phi^{(m)}(\*\lambda) ; \*\psi_{m,m^\prm})  \notag
 \\
 & \hspace{6em} - \II(A = A^{(m^\prm)}) \log (1- g_{m,m^\prm}(\phi^{(m^\prm)}(\*\lambda) ; \*\psi_{m,m^\prm}))
 \bigr\}, 
 \label{eq:DANN_loss}
\end{align}
where $\II$ is the indicator function.
% ただし，$\II$は指示関数である．
%
% ある与えられた$\*\theta^{(m)}, \*\theta^{(i^\prm)}$のもとで分類器を最適化する問題は
For given 
$\*\theta^{(m)}$
and
$\*\theta^{(m^\prm)}$, 
the optimization problem for the classifier: 
\begin{align*}
 \min_{\*\psi_{m,m^\prm}} \ 
 \cL_{m,m^\prm}^{\rm (CE)}(\*\psi_{m,m^\prm} ; \*\theta^{(m)}, \*\theta^{(m^\prm)}).
\end{align*}

\begin{figure}[t]
  \centering
  \includegraphics[width=0.95\linewidth]{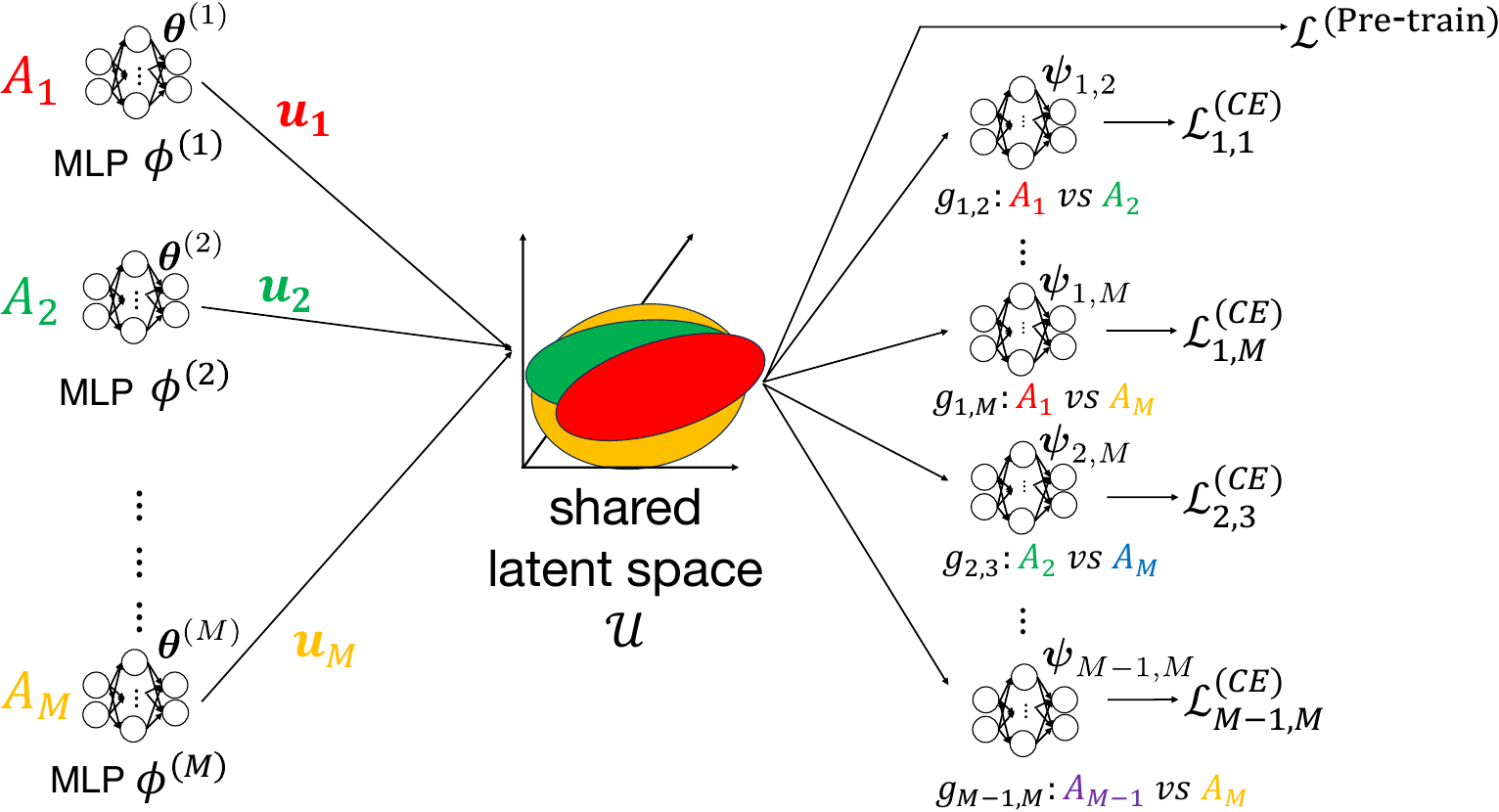}
 \caption{
 Illustration of objective function in pre-training.
 $\mr{Acc}$ is fitted by the quadratic function through 
 $\cL^{(\mr{Pre-train})}$, 
 while $\cL_{(\mr{CE})}$ encourages sharing the latent space among different ML algorithms.
 %  提案手法の事前学習モデル．
 %  %
 %  各機械学習アルゴリズムのハイパーパラメータ空間をMLPによって潜在空間上に写像する．
 %  %
 % 潜在空間上で$\cL^{(\mathrm{Pre-train})}$によって各機械学習アルゴリズムのハイパーパラメータ空間に二次曲面を構築し，$\mathcal{L}_{(\mathrm{CE})}$によって異なる機械学習アルゴリズム間の空間共有を促す．
 }
 % 潜在空間上で$\mathcal{L}^{(\mathrm{Pre-train})}$によって各機械学習アルゴリズムのHPを二次曲面上へと写像し，$\mathcal{L}_{(\mathrm{CE})}$によって異なる機械学習アルゴリズム間の観測データの共有を促す．
  \label{fig:pre-train_learning}
\end{figure}

% 式(\ref{eq:pre_train_loss1})で定義した
% $\*\theta^{(m)}$
% の学習に対して，クロスエントロピー損失が「大きく」なるような正則化を考えると以下の最適化問題を得る．
For the pre-training (\ref{eq:pre_train_loss1}), we add the negative value of the cross entropy loss as follows:
\begin{align}
 \min_{ \*\Theta^{\rm MLP} } & \
 \sum_{m \in [M]} \cL_m^{\rm (Pre-train)}(\*\theta^{(m)}) \notag\\
 & - \beta 
 \sum_{m = 1}^M \sum_{m^\prm = m+1}^M 
 \min_{ \*\psi_{m,m^\prm} } 
 \cL_{m,m^\prm}^{\rm (CE)}(\*\psi_{m,m^\prm} ; \*\theta^{(m)}, \*\theta^{(m^\prm)}). 
 \label{eq:pre-train_loss}
\end{align}
% ただし，$\beta$は正則化係数，
% $\*\Theta_{\rm MLP} = \{ \*\theta^{(m)} \}_{i \in [L]}$
% である．
where $\beta$ is a regularization coefficient and 
$\*\Theta_{\rm MLP} = \{ \*\theta^{(m)} \}_{m \in [M]}$ 
is the set of parameters of $\phi^{(m)}$.
%
% この最適化問題は，外側の
% $\min_{ \*\Theta^{\rm MLP} }$
% は
% $\cL_i^{\rm (Pre-train)}$
% の最小化を分類の損失
% $\cL_{i,i^\prm}^{\rm (CE)}$
% 最大化で正則化した問題として見ることができ，内側の
% $\min_{ \*\psi_{i,i^\prm} }$
% は敵対的に潜在空間内の分類器を最適化する構造になっている．
In this optimization, 
the outer 
$\min_{ \*\Theta^{\rm MLP} }$
is the minimization of
$\cL_m^{\rm (Pre-train)}$ 
regularized by the maximization of
$\cL_{m,m^\prm}^{\rm (CE)}$, 
and the inner 
$\min_{ \*\psi_{m,m^\prm} }$
optimizes the adversarial classifier.
%
% 図\ref{fig:contour}の二次曲面は実際に敵対的正則化を導入して推定した結果であり，各機械学習アルゴリズムが潜在空間上の領域を共有してる様子が確認できる．
% The surfaces in Fig.~\ref{fig:contour} are constructed with the adversarial regularization, and the different ML algorithms are largely overlapped.
Figure~\ref{fig:pre-train_learning} shows an illustration of \eq{eq:pre-train_loss}.

% --------------------------------------------------
% Figure: 事前学習
% --------------------------------------------------

% --------------------------------------------------
% \subsubsection{事前学習データセットの選択方法}
\subsection{Selection of Pre-trained Embedding Model by Learning to Rank}
\label{ssec:ranking-model}

The pre-training can be performed in advance.
% , which means that it can be completed before obtaining the target dataset.
% %
% Therefore, the pre-training does not cause an increase of cost for the target dataset optimization.
% %
% Further, 
Therefore, we can even perform the pre-training for multiple different source datasets, without increasing the cost of the target dataset optimization. 
Suppose that the pre-training is already finished for each one of $S$ source datasets $\cD_1^\prm, \ldots, \cD_S^\prm$.
This means that we have $S$ different pre-trained embdding models, which we call PTEMs (a PTEM is a set of MLPs $\{ \phi^{(m)} \}_{m \in [M]}$ trained by one of source datasets $\cD^\prm_s$). 
By selecting an appropriate PTEM for a given target dataset $\cD$, a greater performance improvement of BO can be expected.
% For a given target dataset
% $\cD$,
% the selection from the $S$ PTEMs can affect the performance of BO.
% the selection from the $S$ pre-trained embdding models can affect the performance of BO.
% ターゲットタスクである
% $\cD$
% に対して，どのような
% $\cD^\prm$
% で事前学習した$\*\Theta^{\rm MLP}$を使うかは当然最終的なベイズ最適化の効率に影響する．
% For the target dataset 
% $\cD$,
% the selection of the dataset used for the pre-training of
% $\*\Theta^{\rm MLP}$ 
% can affect the performance of BO.
% %
% % 事前学習では任意のデータセットを使うことができるため，ここでは
% % $\cD_1^\prm, \ldots, \cD_N^\prm$
% % の$N$種類のデータセットについて事前学習が実行済みであると仮定する（既知データセットと呼ぶ）．
% We can use any dataset for the pre-training, and we assume that, for $S$ datasets 
% $\cD_1^\prm, \ldots, \cD_S^\prm$, 
% the pre-training is already finished, called source datasets.
% %
% このとき，与えられた$\cD$に対してベイズ最適化の効率の意味で有益な$\cD_i^\prm$を既知データセット$\cD_1^\prm, \ldots, \cD_N^\prm$から推薦するランキングモデルの学習を考える．
To this end, we consider building a ranking model that recommends an effective PTEM.
% source dataset from
% $\cD_1^\prm, \ldots, \cD_S^\prm$. 

% ランキングモデルの学習では既知データセット
% $\cD_1^\prm, \ldots, \cD_N^\prm$
% をCross-validationのように使って学習に使う教師ランキングを作る．
The supervision for training of the ranking model is created by partitioning source datasets
$\{ \cD_s^\prm \}_{s \in [S]}$, 
% $\cD_1^\prm, \ldots, \cD_S^\prm$, 
similarly to cross-validation.
%
% つまり，
% $\cD_1^\prm, \ldots, \cD_N^\prm$
% の中のある１つのデータセット$\cD_k^\prm$を擬似的なターゲットタスクとし，残りの$N-1$個のデータセット$\cD_\ell^\prm (\ell \neq k)$を既知データセットとして使うことで，$\cD_k^\prm$をターゲットとした場合の真のランキングを学習データ用に作ることが可能となる．
We first select 
$\cD_\tau^\prm$ 
% $\cD_{s_{\rm target}}^\prm$ 
as a pseudo target dataset from
$\{ \cD_s^\prm \}_{s \in [S]}$,
% $\cD_1^\prm, \ldots, \cD_S^\prm$,
and use the remaining $S - 1$ datasets as source datasets.
Then, we can create the ground-truth ranking of 
$\{ \cD_s^\prm \}_{s \neq \tau}$
% $\cD_1^\prm, \ldots, \cD_S^\prm$
for the target dataset
$\cD_\tau^\prm$, 
so that it can be used for training of the ranking model.
Note that this ranking model optimization can also be performed before we obtain $\cD$.
% このランキングデータによる学習は真のターゲットデータ$\cD$を得る前に予め実施できることに注意されたい（つまり，学習に時間を掛けてもターゲットデータ最適化の遅延を招くものではない）．

% ランキングの順位は擬似ターゲットタスク$\cD_k^\prm$に対して，実際に$\cD_\ell^\prm (\ell \neq k)$のそれぞれで事前学習したベイズ最適化（獲得関数(\ref{eq:acquisition_function})を使う）のパフォーマンスによって決定する．
The ranking for a pseudo target dataset 
$\cD_\tau^\prm$
% $\cD_\tau^\prm$
is determined by the actual performance of BO with a PTEM trained by 
$\cD_s^\prm (s \neq \tau)$.
%
% ある$\cD_\ell^\prm$を事前学習に使用した際に，$\cD_k^\prm$に対するベイズ最適化が$t$回目の繰り返しまでに発見した最大の$\mathrm{Acc}$を
% $y^{\rm best}_{k,t}(\cD_\ell^\prm)$
% とする．
Suppose that 
% $y^{\rm best}_{\tau,t}(\cD_s^\prm)$
$y^{\rm best}_{t}(\cD_\tau^\prm ; \cD_s^\prm)$
% $y^{\rm best}_{s_{\rm target},t}(\cD_\ell^\prm)$
is the maximum 
$\mr{Acc}$
identified by $t$-iterations of BO on the pseudo target 
$\cD_\tau^\prm$
with the pre-training dataset
$\cD_s^\prm$.
%% $y^{\rm best}_{t}(\*\Theta_i^{\rm MLP})$
%
% ベイズ最適化を$T$ iteration実行すると$\forall t \in [T]$に対して
% $y^{\rm best}_{k,t}(\cD_\ell^\prm)$
% が得られるため，この値の平均値をランキング用の評価値に用いる．
After $T$ iterations of BO, 
% $y^{\rm best}_{k,t}(\cD_\ell^\prm)$ 
$y^{\rm best}_t(\cD_\tau^\prm ; \cD_s^\prm)$ 
for 
$\forall t \in [T]$
are obtained, and we use the average of them as a performance measure to define the ranking:
%
% ある$\cD_k^\prm$に対して$\cD_\ell^\prm$を事前学習に使った場合の評価値を
% We define 
\begin{align*}
% \mathrm{Score}_{\tau,s} = 
 \mr{Score}_\tau(s) = 
 % \mr{Average}_{\rm Initialize}(\mr{Average}_{t \in [T]}
 \mr{Average}
 (
 y^{\rm best}_t(\cD_\tau^\prm ; \cD_s^\prm)
 )),
 % \mathrm{Score}_{k\ell} = \mathrm{Average}_{\rm Initialize}(\mathrm{Average}_{t \in [T]}( y^{\rm best}_{k,t}(\cD_\ell^\prm) )), 
 % \mathrm{Score}_{s_{\rm target},\ell} = \mathrm{Average}_{\rm Initialize}(\mathrm{Average}_{t \in [T]}( y^{\rm best}_{s_{\rm target},t}(\cD_\ell^\prm) )), 
 % \mathrm{Average}(y^{\rm best}_{i,t} ; T, n_{\rm trial})
\end{align*}
where
$\mr{Average}$ 
is the average over the different initializations and $T$ iterations of BO (in our experiments, the average over $10$ different initializations and every $20$ iterations out of total $T = 200$ iterations were taken).
For the target dataset 
$\cD_\tau^\prm$,
the score 
$\mr{Score}_\tau(s)$ 
can be seen as a performance measure of a PTEM obtained by a source dataset 
$\cD_s^\prm$.
% と表現する．
%
% ただし，
% $\mathrm{Average}_{\rm Initialize}$
% はベイズ最適化の初期探索点を変えた平均，
% $\mathrm{Average}_{t \in [T]}$
% は$T$回の繰り返しの中での平均である(実験ではそれぞれ，$10$回平均，$T = 200$までの$20$iterationおきの平均とした)．
% Here, 
% $\mathrm{Average}_{\rm Initialize}$
% is the average over the different initialization of BO, and 
% $\mathrm{Average}_{t \in [T]}$
% is the average over $T$ iterations of BO (in our experiments, the average over $10$ different initialization and every $20$ iterations out of total $T = 200$ iterations were taken).
%
% $\mathrm{Score}_{k\ell}$を$\ell$について降順にソートしたものが擬似ターゲットタスク$k$に対する正解ランキングとなる．
As a result, the descending sort of 
% $\mr{Score}_{\tau,s}$ 
$\mr{Score}_\tau(s)$ 
with respect to $s$ is the ground-truth ranking for the target dataset $\cD^\prm_\tau$.

% ランキングモデルはメタ特徴量\cite{NIPS2015_11d0e628,feurer2015initializing}と呼ばれるデータセットの統計量や情報量，軽量な機械学習アルゴリズムでの予測精度など，そのデータセットを表現する特徴量を活用する．
Our ranking model uses so-called meta-features \cite{NIPS2015_11d0e628,feurer2015initializing}, such as various statistics of the dataset, for defining the input features.
%
% 擬似ターゲットタスク$\cD_k^\prm$のメタ特徴量ベクトルを
% $\*x_k^{\rm meta}$，
% 候補となる既知データセット
% $\cD_\ell^\prm$
% のメタ特徴量を
% $\*x_\ell^{\rm meta}$
% とする．
Let 
$\*x_\tau^{\rm meta}$
and
$\*x_s^{\rm meta}$
be the meta-feature vectors of the pseudo target  
$\cD_\tau^\prm$
and 
a source dataset 
$\cD_s^\prm$, 
respectively.
%
% ，\textcolor{red}{$\*{m},\*{m}^{(m)}$間の差$\*{m} - \*{m}^{(m)}$を計算する．}
%
% この操作を
%$\cD_1^\prm, \ldots, \cD_N^\prm$
%の全てのデータセットに対して行い，
% \textcolor{red}{$\*{m} - \*{m}^{(m)}$}
% を
%
% ターゲットタスク$\cD_k^\prm$に対する$\cD_\ell^\prm$のランキングを予測する場合は
% $| \*x_k^{\rm meta} - \*x_\ell^{\rm meta} |$
% を入力特徴量とする．
To predict the ranking of $\cD_s^\prm$ for the target 
$\cD_\tau^\prm$, 
the ranking model uses 
$\*x_\tau^{\rm meta}$
and
$\*x_s^{\rm meta}$
as the input features.
% the input feature of the ranking model is defined as
% $| \*x_k^{\rm meta} - \*x_\ell^{\rm meta} |$.
%
% ランキングモデルを$f_{\rm rank}$とすると，
% $s_{k\ell} = f_{\rm rank}(|\*x_k^{\rm meta} - \*x_\ell^{\rm meta}|)$
% のようにターゲット$k$に対する$\ell$のスコア値を出力する．
The ranking model, denoted as $f_{\rm rank}$, outputs the score value of $\cD_s^\prm$ for the target $\cD_\tau^\prm$ as 
$f_{\rm rank}(\*x_\tau^{\rm meta}, \*x_s^{\rm meta})$.
% $s_{k\ell} = f_{\rm rank}(|\*x_k^{\rm meta} - \*x_\ell^{\rm meta}|)$.
%
% ターゲットタスク$\cD_k^\prm$に対する既知データセット
% $\cD_\ell^\prm (\ell \neq k)$
% の予測ランキングは$s_{k\ell}$を$\ell$について降順にソートしたものとなる．
The prediction of the ranking for the target $\cD_\tau^\prm$ is defined by the descending sort of 
$f_{\rm rank}(\*x_\tau^{\rm meta}, \*x_s^{\rm meta})$ 
with respect to $s$ for the fixed $\tau$.

% ランキングモデル
% $f_{\rm rank}$
% の最適化にはLightGBMライブラリ\cite{LightGBM}に含まれる{\tt LGBMRanker}を用いる．
To optimize the ranking model $f_{\rm rank}$, we employ {\tt LGBMRanker} in the LightGBM library\footnote{\url{https://lightgbm.readthedocs.io/en/stable/}}. 
{\tt LGBMRanker} is based on a ranking model called LabmdaMART \cite{burges2010ranknet}, in which the base boosting tree model is replaced from the original MART with LightGBM \cite{ke2017lightgbm}.
% これはMARTと呼ばれるブースティング木に基づくランキングモデルであるLabmdaMART\cite{burges2010ranknet}をLightGBM\cite{ke2017lightgbm}ベースで実装したものである．
%
% LabmdaMARTでは$\mathrm{Score}_{k\ell}$の具体的な値を$s_{k\ell}$で直接近似するのではなく，相対的な大小関係のみに着目することで正確なランキング予測を目指す．
LabmdaMART does not directly approximate $\mr{Score}_\tau(s)$.
Instead, only relative order of the score is used to define the loss function, by which different target datasets can be incorporated into the model optimization.
%
% 特にNormalized Documented Cumulative Gain (NDCG)\cite{liu2011learning}と呼ばれる指標を用いて，$s_{k\ell}$が作るランキングについて上位の精度を重視する目的関数を定義している．
The objective function is defined through Normalized Documented Cumulative Gain (NDCG) \cite{liu2011learning}, in which errors occurred in better positions of the ranking have larger penalties.
%
% NDCGはDCGと呼ばれる正解データのランキングをどれだけ再現できているかを示す指標を$[0,1]$に正規化した指標である．
%
% ランキングモデルが順位$r$と予測した事前データ$\cD_\ell^\prm$の真の順位を$\pi(r)$とする（つまり，$r$は$s_{k\ell}$によって決まる順位であり，$\pi(r)$は$\mathrm{Score}_{k\ell}$によって決まる順位）．
Suppose that $\pi(r)$ indicates the true ranking of a source dataset $\cD_s^\prm$ for which the ranking is predicted as $r$. 
%
% このときNDCGは次のように定義される．
Top $k$ NDCG, which evaluates the accuracy of the top $k$ items in the ranking list, is defined as
\begin{align*}
 \mr{NDCG@}k &= \frac{\mr{DCG@}k}{\max \mathrm{DCG@}k}, 
\end{align*}
% ただし，
% $\mathrm{DCG} = \sum_{r \in [k]} \frac{\mathrm{rel}_{\pi(r)}}{\log_2(r+1)}$
% であり, 
% $\mathrm{rel}_r = (k-r+1)^5$
% を$r$番目の順位に対する適合度，
% $\max \mathrm{DCG}$は$\mathrm{DCG}$の取り得る最大値であり正規化の定数とする． 
where
$\mr{DCG@}k = \sum_{r \in [k]} \frac{\mathrm{rel}_{\pi(r)}}{\log_2(r+1)}$,
$\mr{rel}_r = (k-r+1)^2$
is the relevance score of the $r$-th position, and
$\max \mr{DCG@}k$ is the maximum value of $\mr{DCG@}k$ that makes 
$\mr{NDCG@}k \in [0,1]$. 
%
% NDCGは上位のランキングが正しいほど評価が高くなるように設計されている．
%
% LabmdaMARTは$s_{k\ell}$の相対的な大小関係に対するペアワイス損失をNDCGにより重み付けた目的関数を考え，ブースティングで最適化することで高いNDCGを達成する$f_{\rm rank}$を推定する（詳細は\cite{burges2010ranknet}を参照）．
To optimize $f_{\rm rank}$, LabmdaMART defines a pairwise loss function weighted by NDCG, to which a booting based algorithm is applied (See \cite{burges2010ranknet} for detail).
%
% ランキングモデルのその他の詳細についてはAppendix~\ref{app:create_rankingdata}で説明する．

% \section{関連研究}
\section{Related Work}
\label{sec:related_works}

% CASH problemを解くための最も単純な手法として，ランダムサーチやグリッドサーチ\cite{bergstra2012random}がある．
% しかし，CASH problemの探索空間は通常のハイパーパラメータ最適化と比べて大規模であるため，ランダムサーチやグリッドサーチでは予測精度が高くなるハイパーパラメータと機械学習アルゴリズムの組み合わせの発見には通常のハイパーパラメータ最適化よりも膨大な時間を必要とする．

% CASH problemを解くために，ハイパーパラメータ最適化にも利用されるベイズ最適化\cite{thornton2013auto,snoek2012practical}や遺伝的アルゴリズム\cite{whitley1994genetic,OlsonGECCO2016}を利用した手法が開発されてきた．
For the CASH problem, BO based approaches \cite{thornton2013auto,snoek2012practical} and genetic algorithm based approaches \cite{whitley1994genetic,OlsonGECCO2016} have been studied.
%
% ベイズ最適化や遺伝的アルゴリズムを用いてCASH problemを解く方法の一つとして，
% 異なる機械学習アルゴリズムのハイパーパラメータ空間を以下のように一つの探索空間$\*{\Lambda}$として定式化する方法が活用されている．\cite{thornton2013auto,levesque2017bayesian,feurer2015initializing,NIPS2015_11d0e628}．
A well-known approach to dealing with different HP spaces is to create the concatenated space 
\cite{thornton2013auto,levesque2017bayesian,feurer2015initializing,NIPS2015_11d0e628}: 
\begin{align*}
 \*{\Lambda} = \*{\Lambda}^{(1)} \cup \*{\Lambda}^{(2)} \cdots \cup \*{\Lambda}^{(M)} \cup \left\{\lambda_r\right\}, 
 % \label{eq:SMAC_Search_Space}
\end{align*}
%ただし，$\lambda_r$は候補となる機械学習アルゴリズム$A^{(1)}, \ldots, A^{(L)}$を識別するパラメータである．
where 
$\lambda_r$
is a parameter indicating which candidate ML algorithm $A^{(1)}, \ldots, A^{(M)}$ is now selected (active).
%
% この方法により構成された探索空間上で表現されるブラックボックス関数を1つの確率モデルで予測を行うことで，適切な機械学習アルゴリズムとハイパーパラメータの組み合わせの発見を行う．
% このハイパーパラメータ空間を結合する方法でベイズ最適化を活用する場合，conditional hyperparameter\cite{thornton2013auto,levesque2017bayesian}を扱うことができるSMAC\cite{hutter2011sequential}を活用した手法がよく用いられる．
To search this concatenated space, SMAC \cite{hutter2011sequential} based approaches have been often used, which can handle conditional HPs \cite{thornton2013auto,levesque2017bayesian}.
%
% SMACでは構成された探索空間において，機械学習アルゴリズム$A^{(m)}$が持たないハイパーパラメータにはdefault valueという固定値を与える．
% \red{SMAC gives a default value for an HP that an ML algorithm $A^{(m)}$ does not have.}
The surrogate model in SMAC needs some default value for an HP that an ML algorithm $A^{(m)}$ does not have \cite{levesque2017bayesian}.
%
% 例えば，候補となる機械学習アルゴリズムにサポートベクターマシンとランダムフォレストが存在する場合，ランダムフォレストはサポートベクターマシンのハイパーパラメータ$C$を持っていないため，default valueとして$0.1$を与える．
For example, in the case of support vector machine (SVM) and random forest (RF), RF does not have the SVM HP `$C$' (regularization coefficient). 
%
% このdefault valueは任意の値を与えることができる．
Then, RF has some default value for the dimension corresponding to the SVM $C$ in the concatenated vector.
%
% しかし，これらの手法はハイパーパラメータ空間の結合による探索空間の大規模化やdefault valueの値による結果の変化により，最適化が困難であると考えられる．
However, this approach makes the dimension of the search space (i.e., dimension of the concatenated vector) large, and the theoretical justification of the default value is unclear.
On the other hand, \cite{swersky2013raiders} proposes a kernel that partially uses a shared space for `relevant' parameters, but the relevance should be defined manually and most of hyper-parameters are typically seen as irrelevant each other (e.g., SVM and RF does not have relevant parameters).

% CASH problemを解く別の方法として，機械学習アルゴリズム選択とハイパーパラメータ最適化を別々に解く手法も良く考えられる\cite{nguyen2020bayesian,liu2020admm,li2020efficient,li2023volcanoml}．
Another approach to the CASH problem is to separate the ML algorithm selection and the HP selection \cite{nguyen2020bayesian,liu2020admm,li2020efficient,li2023volcanoml}.
%
% \cite{nguyen2020bayesian}では機械学習アルゴリズム毎に独立にベイズ最適化を行うことで，複数の小規模な探索空間に対して最適な機械学習アルゴリズムとハイパーパラメータの組み合わせの発見に取り組まれている．
\cite{nguyen2020bayesian} applied independent BO to each ML algorithm, by which each one of the HP spaces can be small.
However, sufficient observations should be required for all candidate ML algorithms because of the independence. 
% しかし，機械学習アルゴリズム毎に独立な予測モデルの構築が必要となり，各機械学習アルゴリズムの観測結果を共有することができなくなる．
% したがって，精度の高い機械学習アルゴリズムとハイパーパラメータの組み合わせを発見するためには多くの実験を必要とするため非効率である．
% \cite{li2020efficient,li2023volcanoml}では\cite{nguyen2020bayesian}のように機械学習アルゴリズム毎に独立にベイズ最適化を行い，最適化中に最も高い予測精度の観測が期待できる機械学習アルゴリズムの選択を行う．
\cite{li2020efficient,li2023volcanoml} also use independent BO, and select only a promising ML algorithm as the final search candidate. 
This approach can reduce the search space, but to select an ML algorithm appropriately, sufficient observations are again required for all ML algorithms.
Further, in practice, there is a risk of discording the true best ML algorithm.
% そして，選択された機械学習アルゴリズムに対してのみ探索を行う．
% これにより，最終的に単一の機械学習アルゴリズムに対してのみのハイパーパラメータ最適化が行われるため，小規模な探索空間での最適化が可能となる．
% しかし，この手法は\cite{nguyen2020bayesian}と同様に異なる機械学習アルゴリズムの観測結果の共有を行うことができないため，適切な機械学習アルゴリズムの選択が行われるまでは探索が非効率となってしまう．
% また，選択された機械学習アルゴリズムが最も高い予測精度を記録できないアルゴリズムの場合，最終的に得られる予測精度が低下するため，アルゴリズムの選択条件は慎重に設定する必要がある．

% これに対して，我々の手法は潜在空間上での探索を行うことから，(\ref{eq:SMAC_Search_Space})のように大規模な探索空間上での探索を必要とせず，効率的な探索が可能となる．
% また，共有潜在空間上で異なる機械学習アルゴリズムの観測データの共有を可能とし，\cite{li2020efficient,li2023volcanoml}のように単一の機械学習アルゴリズムのハイパーパラメータ最適化を行うわけではないため，不適切な機械学習アルゴリズムの選択による性能低下を回避することができる．

% 我々は事前学習された潜在空間を使用してベイズ最適化によるCASH problemを解いているが，我々の手法と同様に事前知識を用いてCASH problemを解くためにメタ学習\cite{lemke2015metalearning}を活用した手法が提案されている．
While we introduce a pre-training for the latent space learning, meta-learning \cite{lemke2015metalearning} approaches to the CASH problem have also been studied.
%
% \cite{mu2022auto,wang2020auto}では機械学習アルゴリズム選択をメタ学習で行い，選択された機械学習アルゴリズムに対してハイパーパラメータ最適化を行う．
% これらの手法は上述した\cite{li2020efficient,li2023volcanoml}とは異なり，探索開始前に機械学習アルゴリズム選択を行うため，\cite{li2020efficient,li2023volcanoml}よりも早い段階で小規模な探索空間上での最適化を行うことが可能である．
\cite{mu2022auto,wang2020auto} learn the ML algorithm selection through meta-learning and the HP optimization is performed only for the selected ML algorithm.
%
% しかし，これらの手法は単一の機械学習アルゴリズムに対してのみ最適化が行われるため，不適切な機械学習アルゴリズムが選択されると得られる予測精度の低下につながると考えられ，メタ学習の精度が得られる結果に大きく依存する．
Since these approaches select a single ML algorithm before the optimization starts, the risk of the miss-selection of the ML algorithm can be high.
%
% \cite{dagan2024automated,cohen2019autogrd,laadan2019rankml}では主に機械学習アルゴリズムとハイパーパラメータの組み合わせ選択にメタ学習を用いている．
\cite{dagan2024automated,cohen2019autogrd,laadan2019rankml} considers meta-learning for the simultaneous selection of an ML algorithm and an HP.
%
% これらの手法は事前に決められた機械学習アルゴリズムとハイパーパラメータの組み合わせに対して，与えられたタスクが高い予測精度を記録できる組み合わせをメタ学習により予測し，予測精度が高いと予測された組み合わせから順番に評価を行なっていく．
The meta-learner predicts a pair consisting of an ML algorithm and an HP that achieves high accuracy, and generates a fixed-size ranking list used to evaluate performance sequentially.
%
% これらの手法は高い予測精度が期待される機械学習アルゴリズムとハイパーパラメータの組み合わせから順番に評価が行われるため，少ない実験時間，実験回数で高い予測精度を得ることが期待できる．
% しかし，この手法は事前に決められた機械学習アルゴリズムとハイパーパラメータの有限個の組み合わせに対してのみ探索されない．
% 従って，与えられたタスクに対して最も高い予測精度を記録する機械学習アルゴリズムとハイパーパラメータの組み合わせが候補に含まれず，最適解を必ず得ることができない可能性がある．
However, this strategy is not adaptive to the observations of the given target dataset unlike our approach, by which the error in the meta-learner cannot be corrected. % (meaning that even the optimal solution can be removed at the beginning).}

% これに対して，我々の手法は事前学習済みの潜在空間選択に対して事前知識を活用している．
% 我々の手法は連続したハイパーパラメータ空間を潜在空間上へと写像するため，探索空間上に必ず最適な組み合わせが含まれており，全ての機械学習アルゴリズムが探索候補となるため，既存のメタ学習を用いた手法と比べても
% 探索精度の低下のリスクは軽減される．

BO algorithms using the latent space surrogate have been studied (e.g., \cite{gomez2018automatic}). 
Typically, to avoid the acquisition function maximization in the structured input (such as sequences and graphs) or high dimensional space, the acquisition function is maximized in the latent space from which the next search point is `decoded'. 
On the other hand, we do not employ this decoding approach and the acquisition function maximization is performed in the original space as described in \eq{eq:acquisition_function}.
This is to avoid the cycle consistency problem \cite{jha2018disentangling}, i.e., the selected latent variable $\*u$ may not be consistent with the encoded value of the decoded $\*u$, by which the GPs cannot obtain the observation at the selected point.
Combining recent techniques mitigating this problem \cite{boyar2024latent} with our proposed method is a possible future direction.
Recently, transfomrer based latent space approaches have been studied \cite{lyu2023efficient,li2024an}.
%
% They regard an HP as `token' by which a different number of HPs can be handled, while we employ the simple MLP because the number of HPs is fixed beforehand in our setting.
They regard an HP as a `token' by which a variable size of HPs can be handled, while we employ the simple MLP because the size of HPs is fixed beforehand in our setting.
Further, \cite{li2024an} does not discuss pre-training and \cite{lyu2023efficient} does not consider the CASH problem.

% --------------------------------------------------
% \section{実験}
\section{Experiments}
\label{sec:experiments}

% The algorithm of the proposed method is shown in Algorithm~\ref{alg:proposed-method}. 
%
% 提案手法の性能を評価するため，CASH problemに利用されるAutoML手法と比較実験を行った．
We compare the performance of the proposed method with seven existing AutoML methods applicable to the CASH problem.

% --------------------------------------------------
% \subsection{実験設定}
\subsection{Settings}
\label{ssec:experimental_setup}

% 比較に利用する手法として，以下の手法を使用する．
We used the following methods as baselines:
\begin{itemize}
 \item % Random Search\cite{bergstra2012random}: 観測する機械学習アルゴリズムとハイパーパラメータの組み合わせをランダムに決定する．
	Random Search\cite{bergstra2012random}: Both the ML algorithm and its HPs are randomly determined.

 \item % SMAC\cite{hutter2011sequential}: ランダムフォレストを予測モデルに使用したベイズ最適化手法．主にハイパーパラメータ最適化に利用されるが，CASH problemの比較実験によく使用される．
       SMAC \cite{hutter2011sequential}: SMAC is a well-known HP optimization method with a random forest surrogate, and has been often employed as a baseline of the CASH problem.

  \item % Bandit BO\cite{nguyen2020bayesian}: 機械学習アルゴリズム毎に予測モデルを構築する方法．
	Bandit BO \cite{nguyen2020bayesian}: This approach is based on BO with independent surrogate models for candidate ML algorithms.
	%
	% 各機械学習アルゴリズムに対してガウス過程回帰で予測および獲得関数計算を行い，獲得関数値が最も高い機械学習アルゴリズム，ハイパーパラメータの組み合わせを次の観測データとする．
	GPs are estimated for all ML algorithms and their acquisition functions are maximized, respectively.
	A pair of an ML algorithm and HPs with the highest acquisition function is selected as a next candidate. 

 \item % Rising Bandit by SMAC(RB-SMAC)\cite{li2020efficient}: 機械学習アルゴリズム毎に予測モデルを構築する手法．
       Rising Bandit by SMAC (RB-SMAC) \cite{li2020efficient}: This approach also based on independent surrogate models for candidate ML algorithms.
       %
       % 探索中にテストデータセットに適した機械学習アルゴリズムの選択を行い，選択された機械学習アルゴリズムのみ探索を行う．最適化のフレームワークにSMACを使用．
       During the optimization, RB-SMAC gradually discords candidate ML algorithms that are estimated unlikely to be the optimal selection.
       SMAC is used for the based optimizer. 
 
  \item % Pre-train BO\cite{wang2024pre}: 事前学習したガウス過程回帰モデルを使用してベイズ最適化を実行．ベイズ最適化フレームワークにBandit BOを使用．
	Pre-train BO \cite{wang2024pre}: BO with pre-trained GPs. 
	The optimization procedure is same as Bandit BO.

  \item % Algorithm Selection by Meta-Learning(AS-ML): メタ学習により機械学習アルゴリズムを選択し，選択されたアルゴリズムに対してハイパーパラメータ最適化を実行する．\cite{mu2022auto}を参考に実装した比較手法．
	% Algorithm Selection by Meta-Learning (AS-ML): An ML algorithm is selected by a prediction model obtained from a meta-learning algorithm, which is based on \cite{mu2022auto}. 
	Algorithm Selection by Meta-Learning (AS-ML): An ML algorithm is selected by a ranking model using meta-features. 
	An HP optimization is performed for the selected ML algorithm by BO.
	This strategy can be seen as a simplified version of \cite{mu2022auto} (see Appendix~\ref{ssapp:AS-ML} for detail). 

  \item % Algorithm and Hyperparameter Selection by Meta-Learning(AHS-ML): メタ学習により機械学習アルゴリズムとハイパーパラメータの組み合わせを選択する．\cite{dagan2024automated,cohen2019autogrd,laadan2019rankml}を参考に実装した比較手法．
	% Algorithm and Hyperparameter Selection by Meta-Learning (AHS-ML): A pair of an ML algorithm and an HP setting is predicted by a meta-learning algorithm, which is based on \cite{dagan2024automated,cohen2019autogrd,laadan2019rankml}.
	Algorithm and Hyper-parameter Selection by Meta-Learning (AHS-ML): As an additional meta-learning based baseline, we extended AS-ML so that hyper-parameters can be simultaneously selected.
	A similar meta-feature based ranking model to AS-ML is constructed, by which the ranking list for a pair of ML algorithm and its HPs is generated (see Appendix~\ref{ssapp:AHS-ML} for detail). 
	An observation is obtained from the top of the ranking list. 

\end{itemize}

% また，実験ではscikit-learn\cite{scikit-learn}で利用可能な$L = 12$種類の機械学習アルゴリズムを使用した．
We used $M = 12$ ML algorithms available at scikit-learn \cite{scikit-learn}, and 
%
% それぞれハイパーパラメータ空間は$1$から$5$の次元を持っている．
each ML algorithm has $1$ to $5$ dimensional HPs (see Appendix~\ref{app:MLAlg_and_HP} for the complete list).
%
% 実験に利用するデータセットには機械学習研究者が共有して使うことができるツールであるOpenML\cite{OpenML2013}のデータセットを使用した．
% 実験に利用するデータセットにはOpenML\cite{OpenML2013}のデータセットを使用した．
The datasets is from OpenML \cite{OpenML2013}.
%
% 今回はクラス分類問題のみを対象に実験を行う．
We used classification problem datasets.
%
% このデータセットの中から提案手法およびpre-train BOに用いる事前学習データセットとして120種類のデータセットを用意し，ターゲットデータセットとして32種類のデータセットを用意した．
For the proposed method and pre-train BO, $161$ source datasets were used for pre-training, and $40$ datasets were used for target datasets (see Appendix~\ref{app:pre-train_daataset} for the complete list).
%
% 各機械学習アルゴリズムのハイパーパラメータ空間は$[0,1]$となるよう正規化処理を行なった．

% 提案手法において，ハイパーパラメータ空間を潜在空間へと写像するMLPは中間層が2層のMLPを使用する．
In the proposed method, the latent space embedding $\phi^{(m)}$ was implemented by an MLP with two hidden layers, and the Gaussian kernel was used for the covariance of $\*u$ in LMC (the LMC rank parameter is set as 1 which is default of {\tt gpytorch}). 
%
% また，テストデータセットに対して最適な機械学習アルゴリズムとハイパーパラメータの最適化を行う際，事前学習されたMLPに対して，我々は2つ目の中間層から出力層へ変換する重み・バイアスパラメータのみを最適化する．
% また，ターゲットデータセットに対して最適化を行う際，事前学習されたMLPに対して最終層のみ重み・バイアスパラメータを最適化する（Fine tuning）．
In the parameter optimization \eq{eq:loss_func} during the target dataset optimization, for the pre-trained MLPs, we only optimize the last layer (fine tuning).
%
% 提案手法の潜在空間の次元数は2とし，事前学習の目的関数(\ref{eq:pre_train_loss1})における定数$\*{c}$は
% $\*{c} = \begin{bmatrix}
%   0.5 & 0.5
% \end{bmatrix}^{\top}$
% とした．
% The dimension of the latent space is $2$, and the constant $\*c$ in the pre-training objective function \eq{eq:pre_train_loss1} is $\*c = (0.5, 0.5)^\top$.
% The constant $\*c$ in the pre-training objective function \eq{eq:pre_train_loss1} is $\*c = (0.5, 0.5)^\top$.
%
% また，(\ref{eq:pre_train_loss1})の$\tilde{y}_j^{(m)}$を計算するための$\textrm{Standardize}$は
% $\mathrm{Standardize}(x) = \frac{x}{y_{\rm best}^\prm - y_{\rm worst}^\prm}$
% とした．
% ただし，$ y_{\rm worst}^\prm$は(\ref{eq:observed_predata})における$y_j^{\prm(m)}$の最小値とする．
% The standardization in (\ref{eq:pre_train_loss1}) for 
% $\tilde{y}_n^{(m)}$ 
% is defined by
% $\mathrm{Standardize}(x) = \frac{x}{y_{\rm best}^\prm - y_{\rm worst}^\prm}$, 
% where 
% $ y_{\rm worst}^\prm$
% is the minimum of $y_n^{\prm(m)}$ in the pre-training dataset (\ref{eq:observed_predata}).
%
% また，(\ref{eq:loss_func})，(\ref{eq:pre-train_loss})における正則化係数$\alpha$，$\beta$はともに$10^{-3}$とした．
The regularization coefficients 
$\alpha$ 
in \eq{eq:loss_func} and 
$\beta$ 
in \eq{eq:pre-train_loss} were set as $10^{-3}$ and $10^{-4}$, respectively.
The dimension of the latent space was $3$.
%
% これは3.2節で紹介した事前学習用のデータセットを用いて事前最適化を行い，決定した．
% これらは事前学習用のデータセットを用いて事前最適化を行なって決定した．
The regularization coefficients and the latent dimension were determined by optimizing for pre-training datasets (described in Appendix~\ref{app:decide_alpha_and_beta}).
%
% 詳細はAppendix~\ref{app:decide_alpha_and_beta}で紹介する．

% --------------------------------------------------
% Fig: Main result
% --------------------------------------------------
\begin{figure}[t]
 \centering
 \ig{.45}{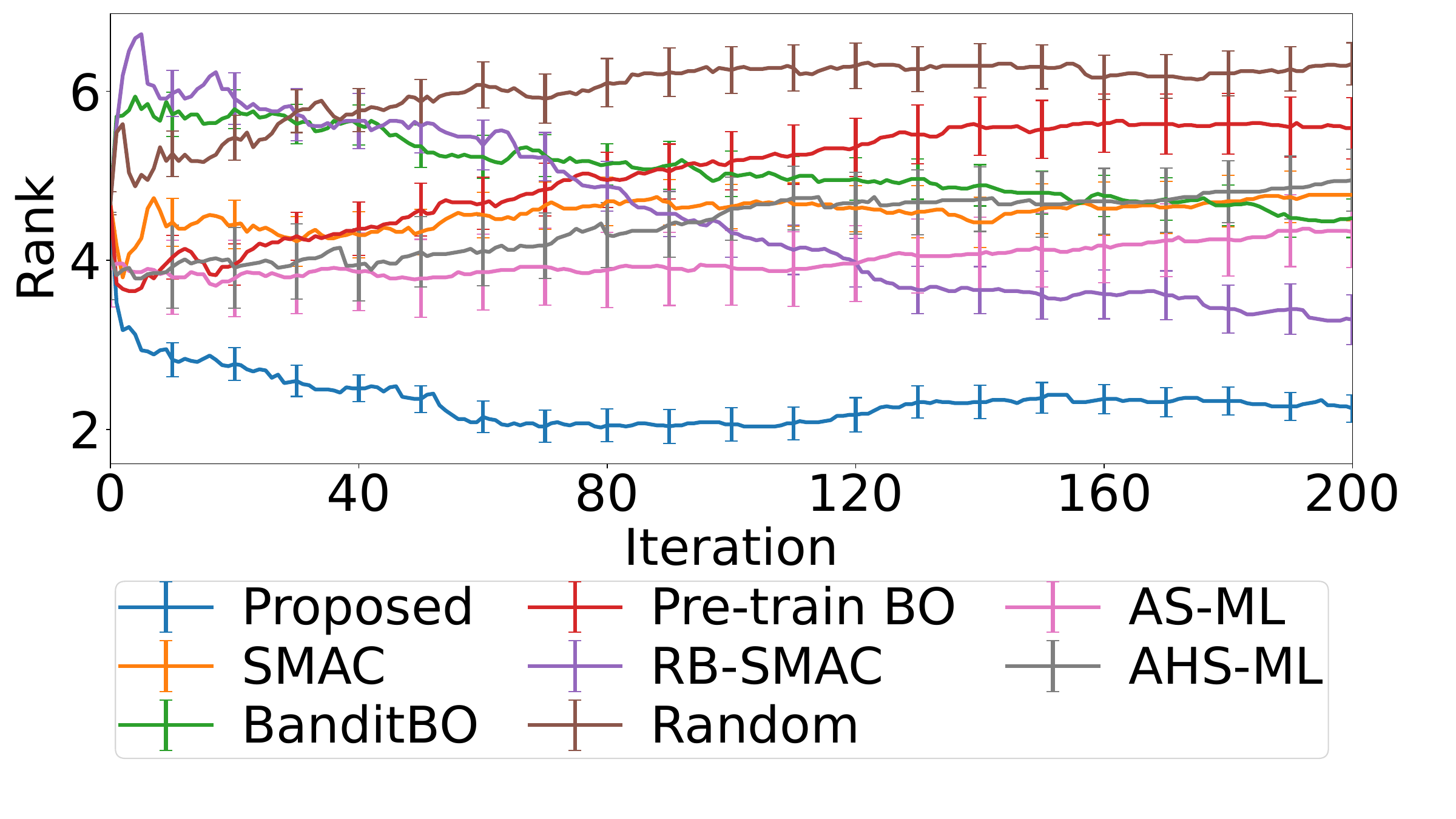}
 % \caption{提案手法と比較手法の比較結果．横軸：探索回数，縦軸：データセット毎の手法間の平均ランキング}
 \caption{
 Ranking-based performance comparison ($10$ runs average ranking over $40$ target datasets). 
 }
 \label{fig:iteration}
\end{figure}

% --------------------------------------------------
% \subsection{実験結果}
\subsection{Main Result}
\label{ssec:experimental_result}

% 実験結果は図\ref{fig:iteration}のとおりである．
The results are shown in Fig.~\ref{fig:iteration}, in which the performance of each method is compared by the ranking among eight compared methods (lower is better).
%
% この結果はテストデータセット$\mathcal{D}^{(m)}$に対して，各手法が何番目に良い手法であるかをランク付けを行い，全てのテストデータセットの平均ランクで評価を行う．
For each one of $40$ target datasets, each method run $10$ times by changing initial points (randomly selected two HP settings for each ML algorithm $A^{(m)}$).
The search was performed $200$ iterations. 
At each iteration, the average of the objective function \eq{eq:CASH} over $10$ runs are used to create the ranking at that iteration.
%
% Figure~\ref{fig:iteration} shows the average ranking and its standard deviation over $32$ target datasets.
Figure~\ref{fig:iteration} shows the average ranking and its standard error over $40$ target datasets.

% 実験結果を見ると，比較手法よりも明らかに低いランクを記録していることがわかる．
Throughout the iterations, the proposed method achieved better rankings compared with all the other methods.
%
% SMACは実践的に安定的に良いRankingではあるが，提案法はoutperformしてる
% SMAC and RB-SMAC showed high performance particularly in later iterations, but the proposed method outperform them.
RB-SMAC showed high performance particularly in later iterations, but the proposed method outperformed it.
%
% Pre-train BOはBandit BOをほとんどのiterationで改善してはいるが，iteration後半ではBandit BOと似たRankとなった
% Although pre-train BO improved Bandit BO for the large part of iterations, the ranking results in similar values to Bandit BO at the end of iterations.
Although pre-train BO improved BanditBO in early iterations, BanditBO outperformed pre-train BO at the end of iterations.
%
% 事前学習の効果は見られるものの，観測が孤立しているためtarget datasetの観測収集には長いiterationがかかる
% This suggests the benefit of pre-training, but the observation isolation may delay the optimization on the target dataset.
This suggests that the benefit of pre-training can be seen in the beginning, but after observations in the target dataset were accumulated, BanditBO without pre-training showed better performance.  
%
% AHS-MLはiteration初期で良い性能だがRankが少しずつ悪くなった
AHS-ML showed high performance in early iterations, but the ranking became worse gradually. 
%
% AHS-MLはメタ学習でML alg.とHPの組み合わせの推薦リストを作り，上位から順番に観測するので，iterationを進めてもより良いものを発見しにくい
AHS-ML creates a recommended list of combinations of an ML algorithm and an HP setting through an meta-learning model, and sequentially observes from the top of that list.
Therefore, possibility of discovering better solutions during the iterations becomes lower unlike other optimization-based methods.
Comparison based on the objective function value (validation accuracy) is also shown in Appendix~\ref{app:result_details}. 
% 提案法は初期段階からAHS-MLと似たperformanceであり(事前学習のおかけで)，その後，探索を進めることで予測精度の高い機械学習アルゴリズムとハイパーパラメータの組み合わせの発見ができたと考えられる．
% Our proposed method was comparable performance with AHS-ML at the beginning of the iterations, and subsequently, 

% 実験結果を見ると，比較手法よりも明らかに低いランクを記録していることがわかる．
% また，200iteration時に2番目に良いRB-SMACは初期iterationでは他の手法と比べても性能が劣っていることが図\ref{fig:iteration}からわかる．
% これは，RB-SMACは初期iterationで探索するアルゴリズムの決定を行うためである．
% アルゴリズムの決定を行うため，様々な機械学習アルゴリズムの観測を行うが，この手法は機械学習アルゴリズム毎に独立な予測モデルの構築を行うため，ある機械学習アルゴリズム$A^{(m)}$の観測結果は他の機械学習アルゴリズム$A^{(j)}$の予測に影響を及ぼさない．
% 従って，最適な機械学習アルゴリズム$A^{(*)}$が決定するまでに観測した$A^{(*)}$以外の観測が無駄となるため，初期iterationにおいてRB-SMACのランクは高い結果となってしまう．
% これに対して，提案手法はある機械学習アルゴリズム$A^{(m)}$の観測は共有潜在空間上で他の機械学習アルゴリズムの予測に影響を及ぼすことが可能であるため，初期iterationの時点で他の手法と比べて良い結果を得ている．
% また，AHS-MLは初期iterationでは提案手法と同等の性能を示しているが，200iteration時点では他の手法と比べても性能が低下していることがわかる．
% これはAHS-MLがメタ学習により初期iterationで高い予測精度を記録できる機械学習アルゴリズムとハイパーパラメータの組み合わせを発見できる一方，それ以上の性能を発揮する組み合わせの発見が難しいためである．
% 一方，提案手法は初期iterationからAHS-MLと同等以上の結果を得ていることから，提案手法は初期iterationでメタ学習を用いた手法と同等性能を発揮する機械学習アルゴリズムとハイパーパラメータの組み合わせの発見が可能であり，
% 探索を進めることでより予測精度の高い機械学習アルゴリズムとハイパーパラメータの組み合わせの発見ができたと考えられる．

% --------------------------------------------------
\subsection{Ablation Study}
\label{ssec:ablation_study}

% 次に，我々の手法における各構成要素の重要性を観察するため，ablation studyを行う．
We performed ablation study to evaluate the effect of 1) pre-training, and 2) selection of a pre-training dataset.
%
% 実施するablation studyは，1. 事前学習の有無，2. ランキング学習による事前学習データセットの選択，である．

\begin{figure}[t]
 \centering
 \subfigure[
 Proposed method vs Proposed w/o pre-train.
 ]{\ig{.4}{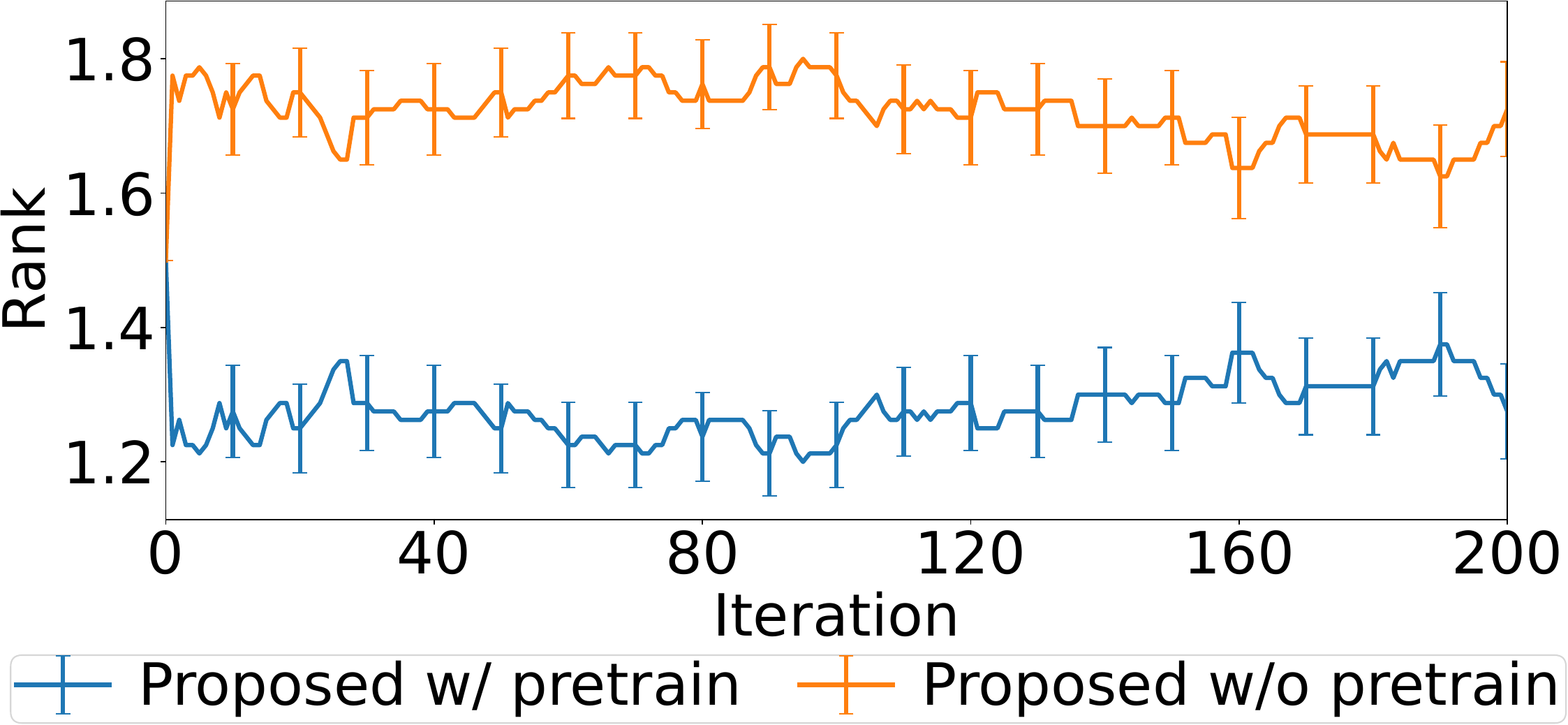}}

 \subfigure[
 Proposed w/o pre-train compared with other methods.
 % Comparison of Proposed w/o pre-train with other methods
 ]{\ig{.45}{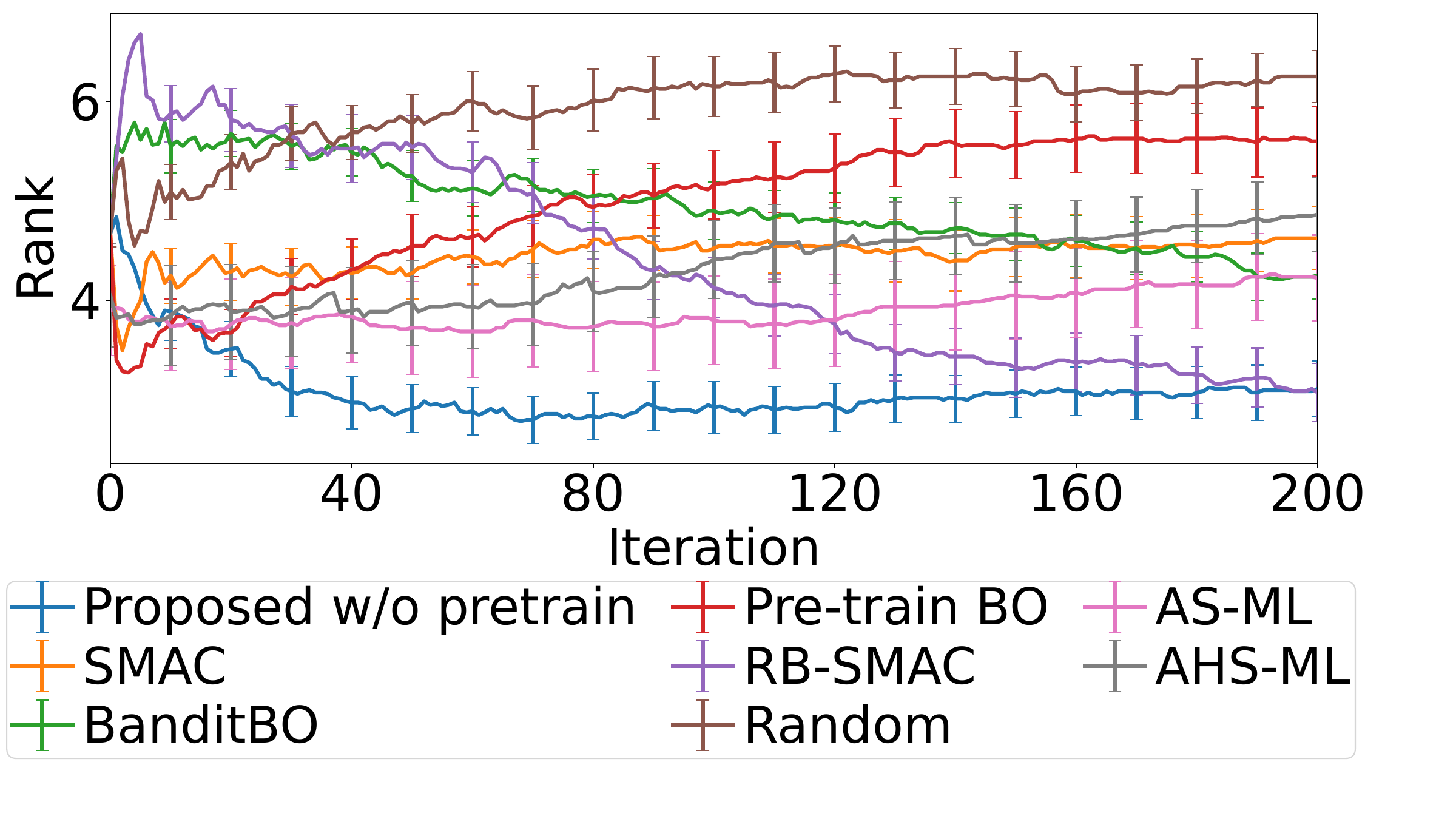}}
  % \begin{minipage}{0.49\linewidth}
  %   \centering
  %   \includegraphics[width=0.9\linewidth]{./figs-result-ablation_study-pretrain.pdf}
  % \end{minipage}
  % \begin{minipage}{0.49\linewidth}
  %   \centering
  %   \includegraphics[width=0.9\linewidth]{./figs-result-ablation_study-pretrain-all.pdf}
  % \end{minipage}
 \caption{
 Ablation study for pre-training. 
 % 事前学習の有無の結果．左：事前学習の有無の違い，右：事前学習がない場合と用意した比較手法の違い
 }
 \label{fig:ablation_study}
\end{figure}

% ablation studyの実施にあたり，各要素を個別に無効にした場合の提案手法の性能を評価した．
% 事前学習を取り除いた場合，\ref{ssec:pre-train}節で説明した事前学習は行わないものとする．
When we remove pre-training, denoted as `Proposed w/o pre-train', the latent space embedding $\phi^{(m)}$ is learned only from the target datasets using \eq{eq:loss_func} without the pre-training described in \ref{ssec:pre-train}.
%
% 実験において，提案手法に用いるMLPは最終層のパラメータのみを学習していたが，事前学習がない場合は全ての層のパラメータを学習する．
In `Proposed w/o pre-train', all the parameters in MLP is learned unlike the original proposed method in which only the last layer is optimized for the target datasets.
%
% また，(\ref{eq:loss_func})の正則化項は事前学習で学習された潜在空間を維持する役割を持っているため，この項も同時に無効化される．
Note that since the regularization term in \eq{eq:loss_func} is the penalty for deviating from the pre-trained models, this term is vanished when we remove the pre-training.
%
% 実験結果を図\ref{fig:ablation_study}に示す．
The results are shown in Fig.~\ref{fig:ablation_study}. 
Fig.~\ref{fig:ablation_study}(a) shows direct comparison between the proposed methods with and without pre-training.
The vertical axis is the ranking same as Section~\ref{ssec:experimental_result} (i.e., here, the rank is 1 or 2).
The results clearly show that the proposed method with pre-training outperforms the proposed method without pre-training.
%
% この結果から，特に事前学習は提案手法の性能向上に大きく貢献していることがわかる．
Fig.~\ref{fig:ablation_study}(b) is performance evaluation of Proposed w/o pre-train compared with other existing methods. 
In this figure, Proposed w/o pre-train still shows mostly comparable performance with the best methods
We can also see the effect of the pre-training by comparing Fig.~\ref{fig:ablation_study}(b) with Fig.~\ref{fig:iteration}, in which clearer differences from existing methods can be seen.

\begin{figure}[t]
 \subfigure[
 Proposed method with ranking-based and random selection of a PTEM.
 ]{
 \ig{.4}{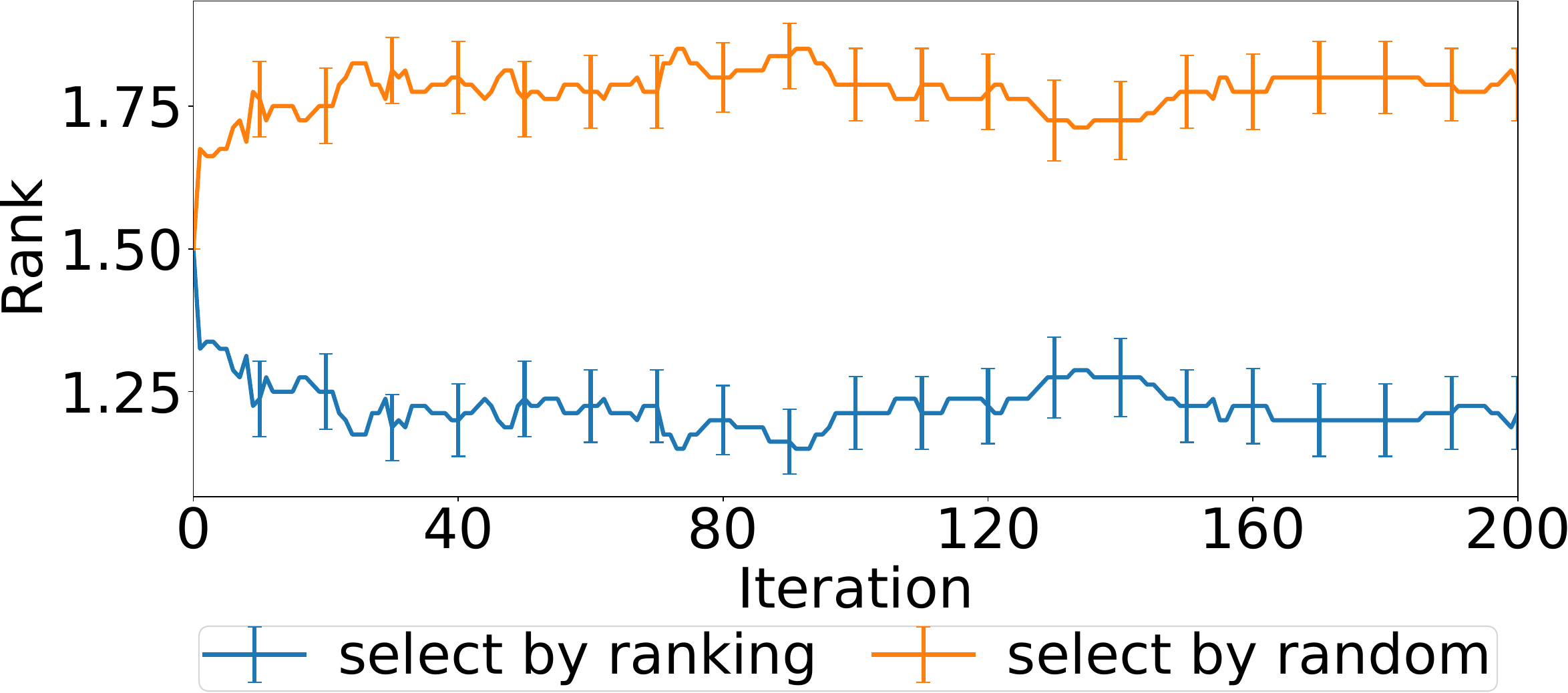}}
 % \ig{.4}{./figs-result-ablation_study-select_predata-L1.pdf}}

 \subfigure[The proposed method with random selection of a PTEM compared with other methods.]{
 \ig{.45}{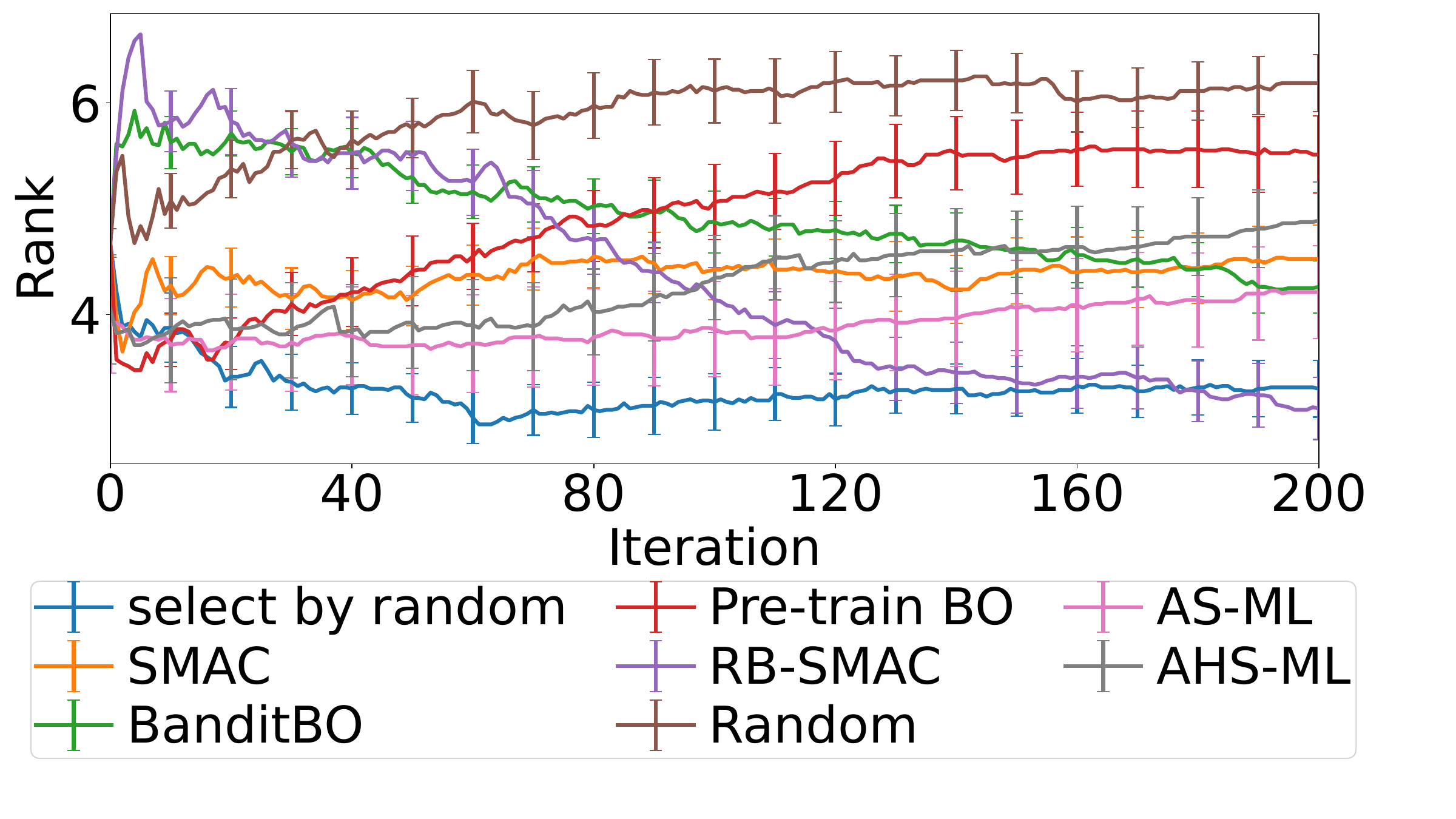}}
 % \ig{.45}{./figs/result/ablation_study/select_predata-all.pdf}}
  % \begin{minipage}{0.49\linewidth}
  %   \centering
  %  \includegraphics[width=0.9\linewidth]{./figs/result/ablation_study/select_predata.pdf}
  % \end{minipage}
  % \begin{minipage}{0.49\linewidth}
  %   \centering
  %   \includegraphics[width=0.9\linewidth]{./figs/result/ablation_study/select_predata-all.pdf}
  % \end{minipage}
 \caption{
 Ablation study for the selection of a PTEM.
 % 事前学習データセット選択の違いの結果．左：事前学習データセット選択の違い，右：L1距離で事前学習データセットを選択した場合と比較手法の違い
 }
  \label{fig:ablation_study2}
\end{figure}

% 一方，ランキング学習による事前学習データセットの選択を取り除いた場合，我々は事前学習データセットをメタ特徴量のL1距離で決定する．
To verify the effect of the ranking model for the selection of a PTEM, we compared the ranking model with the random selection.  
The results are shown in Fig.~\ref{fig:ablation_study2}.
In Fig.~\ref{fig:ablation_study2}(a), we can clearly see that the ranking based selection achieved better performance compared with the random selection throughout iterations. 
Fig.~\ref{fig:ablation_study2}(b) indicates that the proposed method still shows comparable performance to the best methods among the compared methods.  

% つまり，テストデータセットのメタ特徴量を
% $\*x^{\rm meta}$，
% 事前学習データセット
% $\cD_i^\prm$
% のメタ特徴量を
% $\*x_i^{\rm meta}$
% % $\*{m}^{(m)} = (m_1^{(m)}, \ldots,m_f^{(m)})$
% とすると，
% % つまり，テストデータセットのメタ特徴量を$\*{m} = (m_1, \ldots,m_F)$，事前学習データセット$\mathcal{D}_i'$のメタ特徴量を$\*{m}^{(m)} = (m_1^{(m)}, \ldots,m_F^{(m)})$とすると，
% % 
% % $|\*{m} - \*{m}^{(m)}|$
% $|\*x^{\rm meta} - \*x_i^{\rm meta}|$
% が最小となるデータセットを選択した．
% %
% スケール差の影響を緩和するために，L1距離を計算する場合はメタ特徴量は$[0,1]$に正規化する．
% % ただし，メタ特徴量の種類によってrangeが大きく異なり，そのまま使用すると選択される事前学習データセットがrangeが大きいメタ特徴量の影響をうけてしまう．
% % 従って，L1距離による選択を行う場合，メタ特徴量を0-1に正規化した．
% 実験結果は
% \ref{fig:ablation_study2}に示す．
% % 実験結果は図\ref{fig:ablation_study},\ref{fig:ablation_study2}，表\ref{table:ablation_study}のとおりである．
% %
% 同様に，ランキング学習による事前学習データセット選択はわずかであるが，提案手法の性能向上に貢献していることがわかる．

% --------------------------------------------------
% Fig: Ablation 1 without pre-train
% --------------------------------------------------
% --------------------------------------------------
% Fig: Ablation 2 without ranking model
% --------------------------------------------------
% \begin{table}[t]
%   \centering
%   \begin{tabular}{|c|c|}\hline
%     取り除いた構成要素 & 200iteration実行時の平均ランク\\ \hline
%     取り除きなし & $2.234$\\ \hline
%     ランキング学習 & $2.531$\\ \hline
%     事前学習 & $3.406$\\ \hline
%   \end{tabular}
%   \caption{ablation studyにおける，提案手法の200iteration実行時の平均ランク}
%   \label{table:ablation_study}
% \end{table}

\subsection{Wall-Clock-Time Evaluation}
\label{ssec:wall-clock-time}

% --------------------------------------------------
% Fig: Wall-clock
% --------------------------------------------------
\begin{figure}[t]
  \centering
 \includegraphics[width=.8\linewidth]{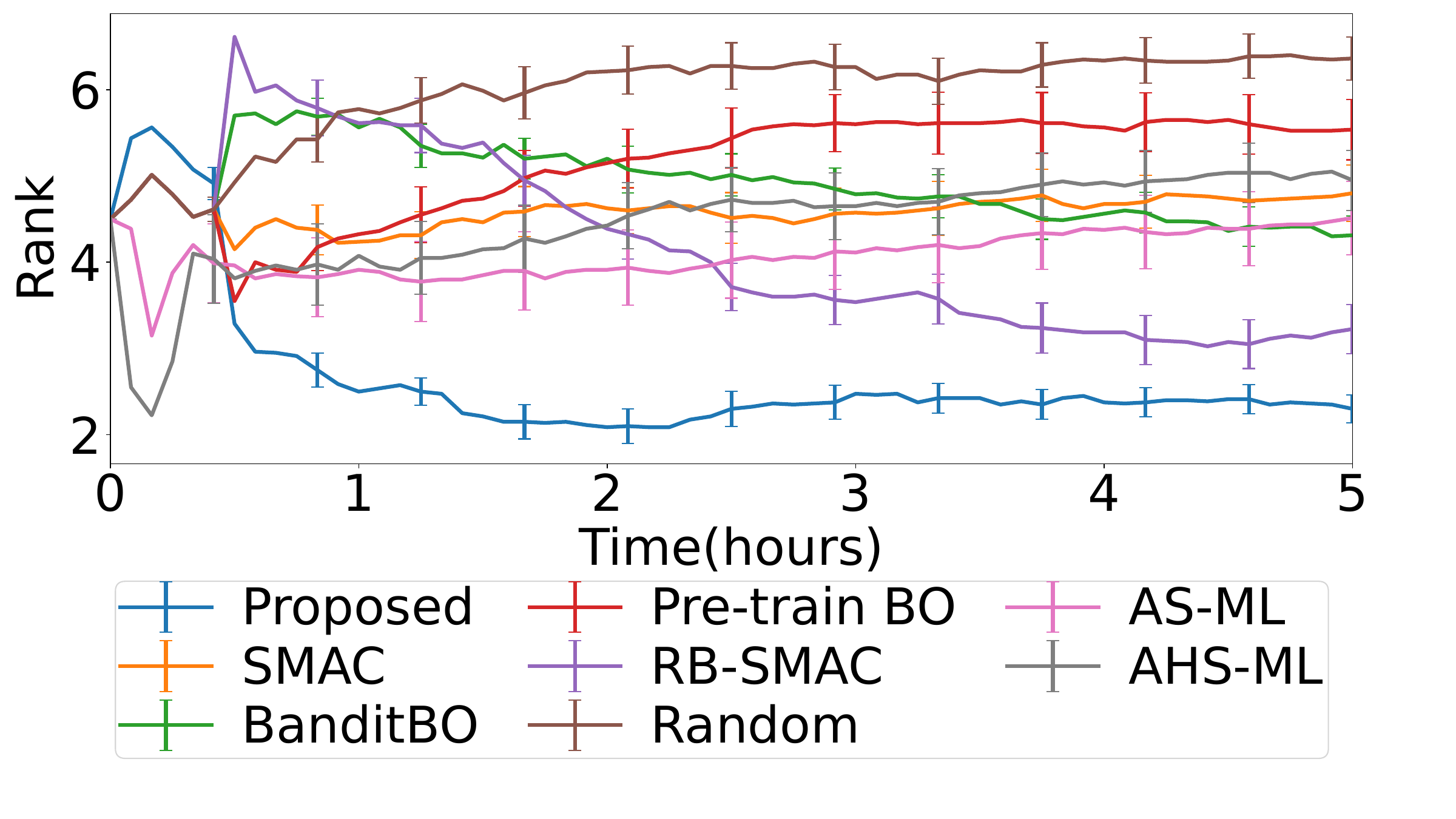}
 \caption{
 Comparison based on wall-clock time.
 %
 % wall-clock timeでの提案手法と比較手法の比較結果．横軸：実験時間(分)，縦軸：データセット毎の手法間の平均ランキング．一度の観測に必要な時間を60秒とした．
 The horizontal axis is computational time (min) and the vertical axis is the average ranking.
 The observation time is set as 1 min.
 }
  \label{fig:wall-clock_time}
\end{figure}

% \begin{table}[t]
%   \centering
%   \begin{tabular}{|c|c|}\hline
%     手法名 & 平均観測データ数\\\hline
%     Proposed & 274.94\\
%     SMAC & 357.55\\
%     BanditBO & 302.84\\
%     BanditBO-pretrain & 300.41\\
%     RB-SMAC & 354.88\\
%     Random & 359.0\\
%     AS-ML & 330.20\\
%     AHS-ML & 357.34\\\hline
%   \end{tabular}
%   \caption{wall-clock-timeでの提案手法と比較手法の観測データ数．提案手法は他の手法と比べて観測できるデータ数が少ないことがわかる．}
%   \label{table:wall-clock_time_num-of-observed}
% \end{table}

% 次に，我々の手法の学習や予測に必要な時間が最適化時にどれだけ影響するかを測るため，wall-clock timeを指標として再度実験を行った．
% 通常，機械学習アルゴリズムの種類や，その機械学習アルゴリズムのハイパーパラメータによって一度の観測に必要な時間は異なることが多いため，
% ある比較手法が予測に大きな時間を必要とする機械学習アルゴリズムとハイパーパラメータが何度も選択されると，他の手法よりも観測可能なデータ数が少なくなり，手法の学習や予測に必要な時間の評価を適切に行うのができない．
%そこで，我々は任意の機械学習アルゴリズムとハイパーパラメータの組み合わせ$(A^{(m)},\*{\lambda}^{(m)})$に対して$\mathrm{Acc}(A^{(m)}, \*{\lambda}^{(i)}, \mathcal{D}_{\mathrm{train}},\mathcal{D}_{\mathrm{valid}})$を得るために必要な時間$t$を$t=10$秒と固定した．

% 次に，wall-clock timeを指標として実験を行った．
We here consider a wall-clock time based evaluation.
%
% ただし，今回は任意の機械学習アルゴリズムとハイパーパラメータの組み合わせに対して，必要な時間$t$は同一だと仮定する．
To make comparison fair, we assume that all ML algorithms and HPs have the same computational cost $t$ for obtaining its $\mr{Acc}$ value.
%
% 実践的には組み合わせに依存して$t$は変わるが，今回比較する探索手法はいずれも計測コストを考慮しない手法であるため，たまたまコストの大きい設定を選んだ手法が不利になってしまう．
In practice, $t$ can change depending on ML algorithms and HPs.
However, in this study, all the compared methods do not explicitly consider time difference of candidates.
Therefore, if the actual time is used for the evaluation, the optimization method that happens to select low cost ML algorithms can be unfairly advantageous.
%
% そのため，公平に（獲得関数計算のコストを含めた）探索効率の比較を行うために$t$共通の設定を採用した（探索にかかるコスト自体を推定して獲得関数に組み込む方法を使うことも考えられるが，獲得関数の性能以外の要素が実験結果に影響を及ぼしてしまうため本稿では扱わない）．
This hinders fair comparison 
(One possible approach is to estimate the cost itself from observations, and incorporate it into the acquisition function. 
However, since this introduces factors beyond the performance of the acquisition function into the experimental results, we do not consider it in this paper). 
%
% 実験では$t=60$秒と固定した．% これは予備的な実験によりランダムサンプリングしたML algorithmとHPの組み合わせに対する平均時間から定めた (about 59 sec for randomly selected 5 HPs of 12 ML algorithms on 40 datasets)．
Instead, we set $t = 60 (\text{sec})$ for all ML algorithms and HPs, which is the average time obtained by the random sampling of HPs for all ML algorithms (it was about $61$ sec for randomly selected $5$ HPs of $12$ ML algorithms on the $40$ target datasets).

% 実験結果は図\ref{fig:wall-clock_time}のとおりである．
The results are shown in Fig.~\ref{fig:wall-clock_time}.
The horizontal axis is the elapsed time (hours) that contains both the acquisition function computations and the ML algorithm computations (i.e., $t$).
%
% Proposed AL-ML, AHS-MLは開始直後のmeta-featureの計算時間も含んでいる．
Note that the proposed method, AL-ML and AHS-ML contain computations of meta-features and the ranking model prediction.
%
% AL-MLとAHS-ML以外の手法は初期点24点(12MLalg各2点)の観測時間も含まれる（AL-MLとAHS-MLはranking listから観測るため初期点を使わない）．
Except for AL-ML and AHS-ML, the elapsed time for initial $24$ points ($2$ points for each of $12$ ML algorithms) are included (AL-ML and AHS-ML observes from the ranking list without initial points). 
%
% また，事前学習は含まない．
Computational times of pre-training is not included because it is performed beforehand.
%
% この結果から，提案手法は他の手法と比べてwall-clock-timeで評価しても優れた性能を示したことがわかる．
We see that the proposed method still shows the superior performance except for the beginning of iterations, in which AHS-ML shows good performance because of the same reason described in Section~\ref{ssec:experimental_result}. 
%
% 提案法が開始直後に比較的に値が大きいのは，メタ特徴の計算や事前学習データの選択にかかる時間も含めているためである．
The proposed method was not high performance at the beginning.
This is because, among the methods who require the observation cost for initial $24$ points (proposed method, pre-train BO, SMAC and RB-SMAC), only the proposed method requires cost of the meta-feature and the ranking model calculations.
%
% 初期観測データの計測が概ね終わる30minあたりで他手法をうわまるのがわかる．
However, we can see at the around 30 min (i.e., after finishing the initial points observation), the proposed method start outperforming the other methods.

% 実験結果は図\ref{fig:wall-clock_time}のとおりである．
% この結果から，提案手法は他の手法と比べて優れた性能を示したことがわかる．
% しかし，図\ref{fig:iteration}と図\ref{fig:wall-clock_time}を比較すると，提案手法とRB-SMACとの差（例えば，図\ref{fig:iteration}の200iteration時の結果と，図\ref{fig:wall-clock_time}の60分時の結果の差）が小さくなったことがわかる．
% また，表\ref{table:wall-clock_time_num-of-observed}から，提案手法は他の手法と比べて60分の間に観測できるデータ数が少ないことがわかる．
% このことから，提案手法の学習や予測には他の手法と比べて多くの時間を必要としており，最適化に影響を与えているということがわかる．
% しかし，限られた時間で提案手法は最も良い性能を示したことから，提案手法は学習や予測に必要な時間を考慮しても，CASH problemの最適化に最も適した手法であると考えられる．

% --------------------------------------------------
% \section{まとめ}
\section{Conclusions}
\label{sec:conclusion}

% 本稿では複数のクラス分類問題において，問題毎に適切な機械学習アルゴリズムとハイパーパラメータの選択を考えた．
% 我々は各機械学習アルゴリズムのハイパーパラメータ空間を共通の潜在空間上へと写像し，潜在空間上でベイズ最適化を行うことで効率的な探索を行う手法を提案した．
We proposed a BO for the CASH problem, in which the surrogate model is constructed on the latent space shared by HP spaces of different ML algorithms. 
Since the quality of the latent space embedding is an important factor for the performance, we further proposed a pre-training of the embedding models.
Finally, we developed a ranking model that recommend an effective pre-trained embedding for a given target dataset.
%
% 計算機実験では提案手法が既存の手法と比べて有効な探索を行うことを確認した．
In the experiments, we demonstrated that the proposed method shows efficient performance compared with existing studies for actual CASH problems.

\makeatletter
\ifdefined\KDD \if@ACM@review
  % KDD review時は空欄
\else
\section*{Acknowledgement}
% % 本研究の一部は，科学研究費 (20H00601,23K21696,22H00300)，JST CREST (JPMJCR21D3)，JST ムーンショット型研究開発事業(JPMJMS2033-05)，JST AIP 加速研究(JPMJCR21U2)，NEDO (JPNP18002, JPNP20006)，理化学研究所革新知能統合研究センターの補助を受けて行われた．
This work was supported by MEXT KAKENHI (20H00601, 23K21696, 22H00300), JST CREST (JPMJCR21D3), JST Moonshot R\&D (JPMJMS2033-05), JST AIP Acceleration Research (JPMJCR21U2), NEDO (JPNP18002, JPNP20006), and RIKEN Center for Advanced Intelligence Project. 
\fi
\makeatother

\bibliographystyle{ACM-Reference-Format}
\bibliography{refs}

\appendix

% \section*{Appendix}
% \label{sec:Appendix}
% \renewcommand{\thesection}{\Alph{section}}
% \setcounter{section}{0} 

% --------------------------------------------------
% \section{獲得関数最適化について}
\section{Acquisition Function Maximization}
\label{app:acquisition_function}

% 我々は次に観測を行う機械学習アルゴリズムとハイパーパラメータの組み合わせを(\ref{eq:acquisition_function})で決定するとき，獲得関数$a$に以下のExpected Improvement関数\cite{brochu2010tutorial}を利用する．
% \begin{align}
%   a(f_A(\*{\*{\lambda}})) = \mathbb{E}[g(f_{A}(\*{\lambda})) - y^*],
%   \label{eq:ei}
% \end{align}
% ただし，$g$はブラックボックス関数を推定する関数(ガウス過程回帰)であり，$y^*$は現在までに得られた最高予測精度を表している．
% 我々は(\ref{eq:ei})を最大化するが，候補となる機械学習アルゴリズム毎に獲得関数の最大化を行う必要があるため，離散変数・連続変数が混在するこの問題において獲得関数計算は多くの時間を必要としてしまう．
%
We need to maximize $a(A^{(m)}, \*\lambda^{(m)})$ defined in \eq{eq:EI} to determine the next point.
We maximize with respect to $\*\lambda^{(m)}$ for each $m$ in which both discrete and continuous variables can be involved.
%
% 我々は\cite{hutter2011sequential}で提案された獲得関数最大化を行うことで，獲得関数最適化に必要な時間の削減を目指した．
We employ a heuristic optimization algorithm similar to those used in SMAC \cite{hutter2011sequential}.

We first randomly sample an initial HP $\*\lambda^{(m)}$.
Let $\lambda^{(m)}_i$ be the $i$-th dimension of $\*\lambda^{(m)}$.
For each candidate $A^{(m)}$, the acquisition function is maximized by the following procedure:
\begin{enumerate}
 \item If $\lambda^{(m)}_i$ is a continuous variable, $10$ points are sampled from $\cN(\lambda^{(m)}_i, 0.1)$.
Note that each continuous variable is assumed to be scaled in $[0,1]$.
If the sample value is the out of the domain, we reject that value and re-sample from the same distribution.
On the other hand, as shown in Table~\ref{table:algorithm_and_HP}, all our discrete variables are count variables (note that KNeighborsClassifier has a categorical variable. However, since we employ the exhaustive search only for KNeighborsClassifier, we do not consider it here). 
For a count variable $\lambda^{(m)}_i$, $10$ points are sampled from the uniform distribution on 
\begin{align*}
 \{ c \in \NN
 \mid 
 \lambda^{(m)}_i - Z
 \leq c \leq
 \lambda^{(m)}_i + Z
 \},
\end{align*}
where 
$\NN$
is the natural number, and
$Z = \lfloor (\lambda_i^{(m) \max} - \lambda_i^{(m) \min} + 1)/10 \rfloor$
in which 
$\lambda_i^{(m) \max}$ 
and 
$\lambda_i^{(m) \min}$
are the maximum and minimum values of 
$\lambda_i^{(m)}$, 
respectively.
Note that since all our count variables in Table~\ref{table:algorithm_and_HP} are
$\lambda_i^{(m) \max} - \lambda_i^{(m) \min} + 1 \geq 10$, 
we have $Z \geq 1$.

 \item The acquisition function values are calculated for the all $10$ points, and employ the maximum among them as the next $\*\lambda^{(m)}$

 \item We repeat the procedure 1 and 2 until the selected $\*\lambda^{(m)}$ does not have a larger acquisition function value than the previous iteration.
\end{enumerate}
We generate $10$ initial HP vectors and apply the same procedure 1-3 to all of them.

\section{Ranking Model Estimation}
\label{app:create_rankingdata}
% \subsection{ランキングモデルの学習データ生成方法}

The procedure of our ranking model estimation is shown in Algorithm~\ref{alg:ranking_model}.
For each pseudo target $\cD^\prm_\tau$, the score $\mr{Score}_\tau(s)$ is calculated for all other source datasets $s \neq \tau$. 
The dataset for {\tt LGBMRanker} is defined by the meta feature vectors $\*x^{\mr{meta}}_{\tau}$ and $\*x^{\mr{meta}}_s$, the score $\mr{Score}_\tau(s)$, and the index of the pseudo target $\tau$. 
$f_{\mr{rank}}$ can use any feature created from 
$\*x^{\mr{meta}}_{\tau}$ and $\*x^{\mr{meta}}_s$. 
In this study, we use 
$|\*x_\tau^{\rm meta} - \*x_s^{\rm meta}|$. 

\begin{algorithm}[t]
  \caption{Ranking Model Estimation}
  \label{alg:ranking_model}

\SetKwInOut{Require}{Require}

\Require{
  $S$ PTEMs trained by $\cD^\prm_1, \ldots, \cD^\prm_S$, 
  meta-feature 
  $\*x_1^{\mr{meta}}, \ldots, \*x_S^{\mr{meta}}$
}
   
$\cD_{\rm rank} \leftarrow \emptyset$ 

\For(\CommentHere{Create training data for ranking model}){$\tau \in [S]$}{ 
   \For{$s \in [S] \setminus \tau$}{
      Set $\{ \phi^{(m)} \}_{m \in [M]}$ as the PTEMs trained by $\cD^\prm_s$

      Apply BO to the pseudo target $\cD^\prm_\tau$ 

      Calculate $\mr{Score}_\tau(s)$
      
      $\cD_{\rm rank} \leftarrow \cD_{\rm rank} \cup \{(\*x^{\mr{meta}}_{\tau}, \*x^{\mr{meta}}_s, \mr{Score}_\tau(s), \tau)\}$
  }
}

Optimize $f_{\mr{rank}}$ by $\cD_{\rm rank}$ through {\tt LGBMRanker} %$\mathcal{R}$を用いてランキングモデル$f_{\mathrm{rank}}$を学習

\Return $f_{\mr{rank}}$
\end{algorithm}

\section{Detail of Experiments}
\label{app:experiment_setting}

% --------------------------------------------------
% \subsection{正則化係数の決定方法}
\subsection{Regularization Coefficients and Dimension of Latent Space}
\label{app:decide_alpha_and_beta}

% ここでは(\ref{eq:loss_func})，(\ref{eq:pre-train_loss})の正則化係数$\alpha, \beta$および潜在空間の次元数の決定方法について紹介する．
% 我々は用意した161種類の事前学習データセットのうち，150種類のデータセットに対してベイズ最適化を実行する．
We optimize $\alpha$, $\beta$ and the latent space dimension by using the source datasets.
%
% この時，$\alpha=10^{-1},10^{-2},10^{-3},10^{-4}$，$\beta=10^{-1},10^{-2},10^{-3},10^{-4}$の4種類ずつ，次元数$=2,3,4$の3種類，計48種類の組み合わせでベイズ最適化を実行した．
The candidates were 
$\alpha = 10^{-1}, 10^{-2}, 10^{-3}, 10^{-4}$,
$\beta = 10^{-1}, 10^{-2}, 10^{-3}, 10^{-4}$, and 
the latent space dimension $2, 3, 4$.
%
% 150種類のデータセットに対して候補となる$\alpha,\beta$, 次元数の組み合わせでベイズ最適化を実行したところ，$\alpha=10^{-3},\beta=10^{-4}$, 次元数$=3$が平均的に最も良い性能を得ることができたため，我々は$\alpha=10^{-3},\beta=10^{-4}$,次元数$=3$を用いて提案手法の性能評価を行なった．
We applied the proposed method to $150$ source datasets and observed $\alpha=10^{-3}, \beta=10^{-4}$ and the latent dimension $3$ was the best average performance, which was employed in our experiments 
(Note that $150$ source datasets are from the original $161$ source datasets. 
We omitted datasets that require long computational time and that achieve max $\mr{Acc}$, i.e., 1, only by initial points for more than $6$ times out of $10$ trials).

% --------------------------------------------------
% \subsection{比較手法について}
\subsection{Meta-learning based Baselines}
\label{sapp:compared_method}

%  ここでは，実験に利用した比較手法AS-MLとAHS-MLについて紹介する．
%  AS-MLとAHS-MLはともにメタ学習手法である．
In our experiments, we used two meta-learning based baselines, called AS-ML and AHS-ML. 
We build these methods by ourselves based on existing meta-learning studies to make comparison fair (e.g., candidate ML algorithms and HPs). 
Both AS-ML and AHS-ML employ LightGBM as a base ranking model.
%  これは我々が実装した手法が事前学習による過去の知見を活用して最適化を行うため，既存の過去の知識を活用した手法と比較するために用意した．
%  しかし，公開されている既存のメタ学習手法は我々の手法と異なるメタ特徴の使用や，最適化対象の機械学習アルゴリズムやハイパーパラメータの違い，
%  提案手法の詳細な設定の未記載など，提案された既存手法と同等のアルゴリズムを作成することが難しい．
%  そこで，我々はそれらの手法をベースに簡単なメタ学習手法を実装し，実験に活用した．
%  ただし，メタ学習モデルは提案手法と同様にLightGBMを用いたランキングモデルとした．

% {(Memo: why existing method is not used)}

\subsubsection{Algorithm Selection by Meta-Learning (AS-ML)}
\label{ssapp:AS-ML}

% AS-MLはメタ学習により機械学習アルゴリズムを選択し，選択された機械学習アルゴリズムのハイパーパラメータに対してベイズ最適化を行うことでCASH problemを解く手法である．
AS-ML selects the ML algorithm by the ranking model, and then, BO is applied to the selected top ML algorithm. 
%
% この手法は図\ref{fig:AS-ML}のようにメタ学習モデルの学習データを生成する．
Figure~\ref{fig:AS-ML} is an illustration of the training data of AS-ML.
Let $\cD_\tau^\prm$ be a pseudo target dataset from our source datasets $\{ \cD_s^\prm \}_{s \in [S]}$
% 
% まず，メタ学習データセット$\mathcal{D}_i^\prm$からメタ特徴$\*{x}_i$を抽出し，候補となる機械学習アルゴリズム集合$\mathcal{A}$のone-hot表現を作成する．
For the ranking model optimization, the input feature in the training dataset consists of the meta-feature 
$\*x_\tau^{\rm meta}$
of the pseudo target
$\cD_\tau^\prm$
and the one-hot representation of the ML algorithm 
$A^{(m)}$.
%
% この抽出したメタ特徴と機械学習アルゴリズムのone-hot表現をメタ学習モデルの入力とし，メタ学習データセット$\mathcal{D}_i^\prm$に対してどの機械学習アルゴリズムが最も高い予測精度を得られるのかを学習する．
The ranking model is optimized so that the ranking (defined by the validation accuracy) of ML algorithms for each pseudo target
$\cD_\tau^\prm$
can be accurately predicted.

% なぜAutoCASHそのままに実装しないのか本当は説明したいが...

% --------------------------------------------------
% Fig: AS-ML
% --------------------------------------------------
\begin{figure}[t]
  \centering
  \includegraphics[width=1\linewidth]{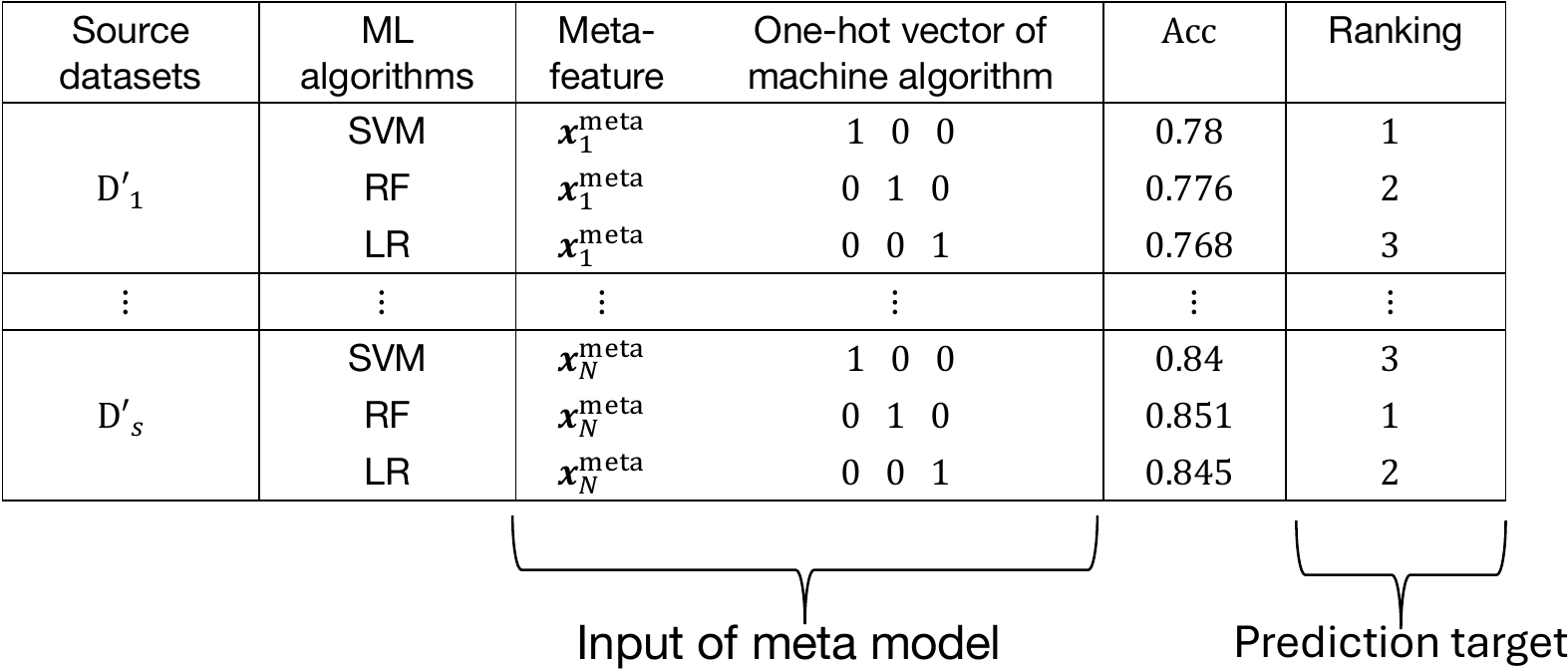}
  \caption{Training dataset of AS-ML.}
  \label{fig:AS-ML}
\end{figure}

% 以下は無しでOK

% \red{
% ターゲットデータセット$\mathcal{D}$が与えられた時，$\mathcal{D}$のメタ特徴を抽出し，抽出したメタ特徴と機械学習アルゴリズムのone-hot表現を学習済みのメタ学習モデルの入力とし，$\mathcal{D}$に対して最も高い予測精度を得られる機械学習アルゴリズム$A^{\mathrm{meta}}$を推薦する．
% %
% その後，推薦された$A^{\mathrm{meta}}$に対して，ベイズ最適化によるハイパーパラメータ最適化を行うことで，$\mathcal{D}$に対して最良と考えられる$(A_*,\*{\lambda}_*)$の探索を行う．
% }

% \red{
%   この手法は単一の機械学習アルゴリズムに対してハイパーパラメータ最適化を実行するため，小規模な探索空間上で最適化が可能となる．
%   つまり，他の既存手法と比べて探索効率が高く，少量のデータ観測で最良の解の発見が期待できる．
%   しかし，メタ学習によって推薦された機械学習アルゴリズムが$\mathcal{D}$に対して最適でない機械学習アルゴリズムの場合，CASH problemの最適解を得ることはできない点が問題点として挙げられる．
% }

\subsubsection{Algorithm and Hyperparameter Selection by Meta-Learning (AHS-ML)}
\label{ssapp:AHS-ML}

% AHS-MLはメタ学習により機械学習アルゴリズムとハイパーパラメータの組み合わせを決定してCASH problemを解く手法である．
AHS-ML creates a ranking list of a pair of an ML algorithm and its HPs.
%
% この手法は図\ref{fig:AHS-ML}のようにメタ学習モデルの学習データを生成する．
The training dataset for AHS-ML is shown in Fig.~\ref{fig:AHS-ML}.
%
% まず，メタ学習データセット$\mathcal{D}_i^\prm$からメタ特徴$\*{x}_i$を抽出し，候補となる機械学習アルゴリズム集合$\mathcal{A}$のone-hot表現を作成する．
%
% このメタ学習データセット$\mathcal{D}_i^\prm$は事前に大量の機械学習アルゴリズムとハイパーパラメータの組み合わせで観測結果を得ているものとする．
%
% この抽出したメタ特徴と機械学習アルゴリズムのone-hot表現，その機械学習アルゴリズムの持つハイパーパラメータの組み合わせをメタ学習モデルの入力とし，メタ学習データセット$\mathcal{D}_i^\prm$に対してどの機械学習アルゴリズムとハイパーパラメータの組み合わせが最も高い予測精度を得られるのかを学習する．
%
% ただし，図\ref{fig:AHS-ML}のように，機械学習アルゴリズム$A_i$が持たないハイパーパラメータの値には$0$を与える．
For the ranking model optimization, the input feature in the training dataset consists of the meta-feature 
$\*x_\tau^{\rm meta}$
of the pseudo target
$\cD_\tau^\prm$, 
the one-hot representation of the ML algorithm 
$A^{(m)}$, 
and the HP vector.
As shown in Fig.~\ref{fig:AHS-ML}, an element of the HP vector is set $0.5$ if $A^{(i)}$ does not have that HP (Note that HPs are scaled in $[0,1]$).
This is inspired by the default value imputation strategy in the conditional HP optimization \cite{levesque2017bayesian}.
%
% テストデータセット$\mathcal{D}$が与えられた時，$\mathcal{D}$のメタ特徴を抽出し，抽出したメタ特徴と機械学習アルゴリズムのone-hot表現，機械学習アルゴリズムの持つハイパーパラメータの組み合わせをメタ学習モデルの入力とし，$\mathcal{D}$に対して最も高い予測精度を得られる機械学習アルゴリズムとハイパーパラメータの組み合わせ$(A^{\mathrm{meta}},\*{\lambda}^{\mathrm{meta}})$を推薦する．
%
% LightGBM libraryの{\tt LGBMRanker}の制約により，ranking listの最大長が$< 10,000$である必要がある
% 全てのML algとHPの組みの候補を含めるとこの上限を超えるため，ヒューリスティックに候補を削減した．
% 各ソースデータセットに対して，上位150のML Algorithm, HPの組みを列挙．
% 全てのデータセットでこれを行い，列挙された組み合わせから重複を除去し，残ったML algorithm, HPの組み合わせを候補とした
%
% AHS-MLでLightGBMの制約で候補を減らした話ですが，各ソースデータセットに対して，上位150のML Algorithm, HPの組みを列挙したのか，各ソースデータとML Alg.の組みに対して上位150をとってきたのかどっちでしたっけ？確か前者だったと思ってるのですが念のため\
In the {\tt LGBMRanker} function of the LightGBM library, the size of the ranking list in the training dataset should be less than $10,000$.
We heuristically reduced the size of candidates to satisfy this condition.
We first collected the top $150$ pairs of an ML algorithm and an HP vector for each source dataset. 
From $150 \times S$ pairs, we removed duplicated settings, and remaining $9,389$ settings were used as candidates from which training ranking list was created.
%
% この機械学習アルゴリズムとハイパーパラメータの推薦を一定回数(\ref{ssec:experimental_result}節のようなiteration比較であれば，指定した回数，\ref{ssec:wall-clock-time}節のようなwall-clock-time比較であれば，指定した時間まで)行い，最適な機械学習アルゴリズムとハイパーパラメータの組み合わせ$(A_*,\*{\lambda}_*)$の探索を行う．
For the final target dataset $\cD$, we created the ranking by the learned ranking model (in this phase, {\tt LGBMRanker} can incorporate all candidates without the $10,000$ constraint), and then, we observe the performance of each pair of the ML algorithm and the HP vector sequentially from the top of the ranking list.

% \red{(Memo: restriction about ranking list size $< 10,000$)}

% --------------------------------------------------
% Fig: AHS-ML
% --------------------------------------------------
\begin{figure}[t]
 \centering
 \includegraphics[width=1\linewidth]{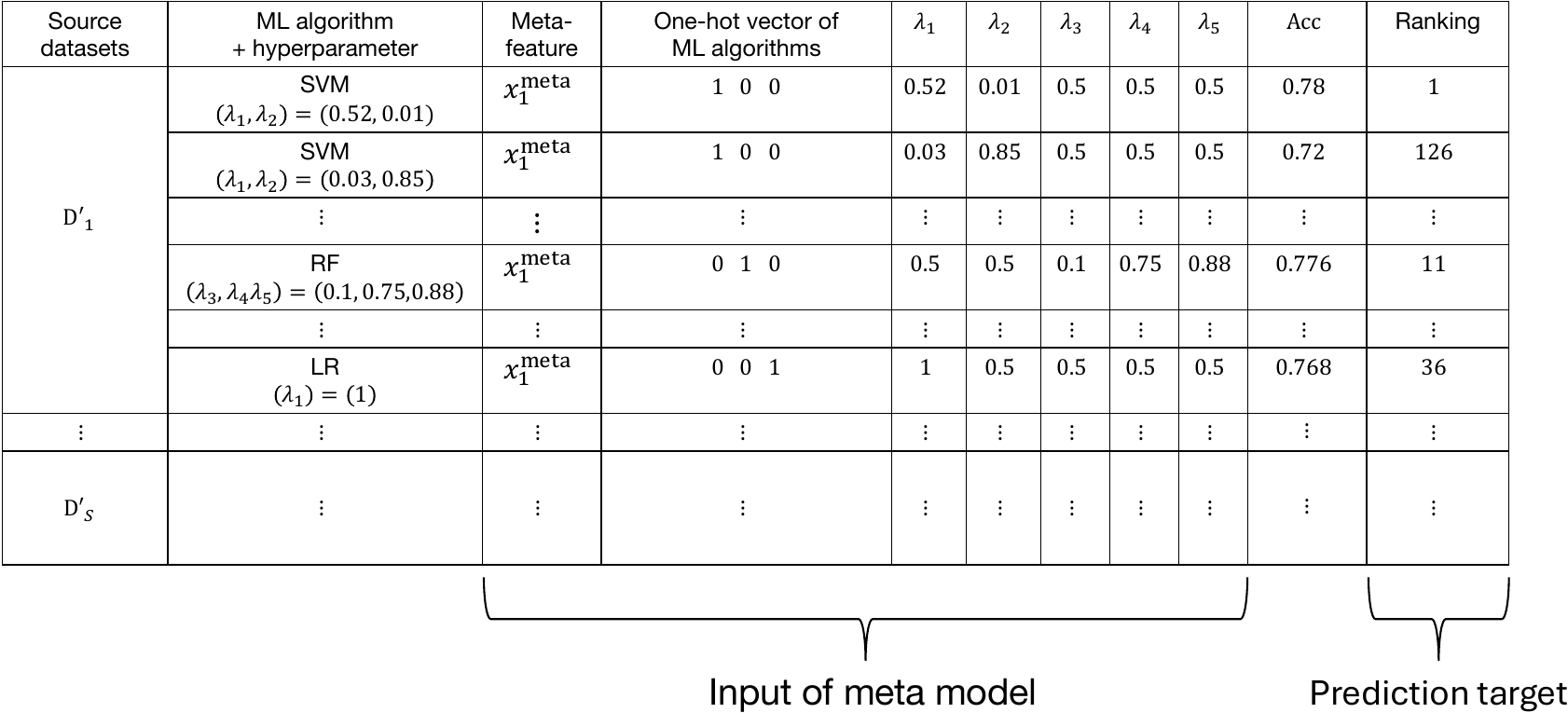}
 \caption{
 %AHS-MLのメタ学習モデルの学習データ生成方法．
 Training dataset of AHS-ML.
 }
  \label{fig:AHS-ML}
\end{figure}

% \red{
%   この手法は過去の実験から高い予測精度が期待される機械学習アルゴリズムとハイパーパラメータの組み合わせを探索初期から推薦する．
%   したがって，探索初期の観測から予測精度の高い機械学習アルゴリズムとハイパーパラメータの組み合わせの選択が期待できる．
%   しかし，この手法は過去の実験の観測結果をもとに推薦を行うため，機械学習アルゴリズムとハイパーパラメータの組み合わせは過去に実験した組み合わせのみ推薦される，
%   つまり，事前に決められた機械学習アルゴリズムとハイパーパラメータの有限個の組み合わせのみ探索は行われる．
%   したがって，与えられたタスクに対して最も高い予測精度を記録する機械学習アルゴリズムとハイパーパラメータの組み合わせが候補に含まれず，最適解を必ずしも得られるわけではない点が問題点として挙げられる．
% }

% --------------------------------------------------
% \subsection{実験に利用した機械学習アルゴリズムとハイパーパラメータ}
\subsection{List of ML algorithms and hyper-parameters}
\label{app:MLAlg_and_HP}

%6章の実験において，我々は表\ref{table:algorithm_and_HP}の機械学習アルゴリズムとハイパーパラメータを使用した．
Table~\ref{table:algorithm_and_HP} shows ML algorithms and hyper-parameters used in Section~\ref{sec:experiments}.
%
% ただし，各機械学習アルゴリズム，ハイパーパラメータの名称はそれぞれscikit-learn\cite{scikit-learn}で定義されている関数名，変数名で記載されている．
The table is described by function and variable names in scikit-learn \cite{scikit-learn}.
%
% --------------------------------------------------
% Table: 機械学習アルゴリズムとハイパーパラメータ
% --------------------------------------------------
\begin{table*}
  \centering
 \caption{
 List of ML algorithms and HPs.
 All discrete variables are count variables.
 % 実験に使用した機械学習アルゴリズムとハイパーパラメータの種類．4列目は各ハイパーパラメータの変数の種類を示している．離散変数の場合，値は1刻みとなっている．
 %
 % また，カテゴリ変数(KNeighborsClassifierのweights)はラベルエンコーディングをしている．
 }
    \label{table:algorithm_and_HP}

    \begin{tabular}{|c|c|c|c|}\hline

     % 機械学習アルゴリズム & ハイパーパラメータ & 探索範囲 & ハイパーパラメータの種類\\ \hline
     ML algorithm & Hyper-parameters & Range & Type \\ \hline \hline
      
      AdaBoostClassifier & \begin{tabular}{c}
          learning\_rage\\
          n\_estimators
      \end{tabular}
      & \begin{tabular}{c}
        $10^{-3} \sim 1$\\
        $1 \sim 50$
      \end{tabular}
      & \begin{tabular}{c}
          Continuous\\
          Discrete
      \end{tabular}\\
      \hline

      BaggingClassifier & \begin{tabular}{c}
        max\_features\\
        n\_estimators\\
        max\_samples
      \end{tabular}
      &
      \begin{tabular}{c}
        $10^{-3} \sim 1$\\
        $1 \sim 50$\\
        $10^{-3} \sim 1$
      \end{tabular}
      &
      \begin{tabular}{c}
        Continuous\\
        Discrete\\
        Continuous
      \end{tabular}\\
      \hline

      DecisionTreeClassifier & \begin{tabular}{c}
        max\_features\\
        max\_depth\\
        min\_samples\_leaf\\
        min\_samples\_split
      \end{tabular}
      &
      \begin{tabular}{c}
        $10^{-3} \sim 1$\\
        $1 \sim 50$\\
        $2 \sim 512$\\
        $2 \sim 512$
      \end{tabular}
      &
      \begin{tabular}{c}
        Continuous\\
        Discrete\\
        Discrete\\
        Discrete
      \end{tabular}\\
      \hline
      
      ExtraTreeClassifier & \begin{tabular}{c}
        max\_features\\
        n\_estimators\\
        max\_depth\\
        min\_samples\_leaf\\
        min\_samples\_split
      \end{tabular}
      &
      \begin{tabular}{c}
        $10^{-3} \sim 1$\\
        $1 \sim 50$\\
        $1 \sim 50$\\
        $2 \sim 512$\\
        $2 \sim 512$
      \end{tabular}
      &
      \begin{tabular}{c}
        Continuous\\
        Discrete\\
        Discrete\\
        Discrete\\
        Discrete
      \end{tabular}\\
      \hline
      
      GaussianNB & \begin{tabular}{c}
          var\_smoothing
      \end{tabular}
      & \begin{tabular}{c}
        $10^{-15}\sim 1$
      \end{tabular}
      & \begin{tabular}{c}
          Continuous
      \end{tabular}\\
      \hline

      GradientBoostingClassifier & \begin{tabular}{c}
        learning\_rate\\
        n\_estimators\\
        max\_features\\
        max\_depth\\
        min\_samples\_split
      \end{tabular}
      &
      \begin{tabular}{c}
        $10^{-3} \sim 1$\\
        $1 \sim 50$\\
        $10^{-3} \sim 1$\\
        $1 \sim 50$\\
        $2 \sim 512$
      \end{tabular}
      &
      \begin{tabular}{c}
        Continuous\\
        Discrete\\
        Continuous\\
        Discrete\\
        Discrete
      \end{tabular}\\
      \hline

      KNeighborsClassifier & \begin{tabular}{c}
        n\_neighbors\\
        p\\
        weights
      \end{tabular}
      &
      \begin{tabular}{c}
        $1 \sim 50$\\
        $1 \sim 2$\\
        uniform, distance
      \end{tabular}
      &
      \begin{tabular}{c}
        Discrete\\
        Discrete\\
        Categorical
      \end{tabular}\\
      \hline

      LogisticRegression & \begin{tabular}{c}
        C
      \end{tabular}
      & \begin{tabular}{c}
        $10^{-3}\sim 10^{3}$
      \end{tabular}
      & \begin{tabular}{c}
          Continuous
      \end{tabular}\\
      \hline

      MLPClassifier & \begin{tabular}{c}
        learning\_rate\_init\\
        alpha\\
        first hidden layer's unit size\\
        second hidden layer's unit size\\
        third hidden layer's unit size
      \end{tabular}
      &
      \begin{tabular}{c}
        $10^{-5} \sim 10^{-1}$\\
        $10^{-2} \sim 10^{2}$\\
        $16 \sim 128$\\
        $32 \sim 256$\\
        $8 \sim 64$
      \end{tabular}
      &
      \begin{tabular}{c}
        Continuous\\
        Continuous\\
        Discrete\\
        Discrete\\
        Discrete
      \end{tabular}\\
      \hline

      QuadraticDiscriminantAnalysis & \begin{tabular}{c}
        reg\_param
      \end{tabular}
      & \begin{tabular}{c}
        $10^{-5}\sim 1$
      \end{tabular}
      & \begin{tabular}{c}
          Continuous
      \end{tabular}\\
      \hline

      RandomForestClassifier & \begin{tabular}{c}
        max\_features\\
        n\_estimators\\
        max\_depth\\
        min\_samples\_leaf\\
        min\_samples\_split
      \end{tabular}
      &
      \begin{tabular}{c}
        $10^{-3} \sim 1$\\
        $1 \sim 50$\\
        $1 \sim 50$\\
        $2 \sim 512$\\
        $2 \sim 512$
      \end{tabular}
      &
      \begin{tabular}{c}
        Continuous\\
        Discrete\\
        Discrete\\
        Discrete\\
        Discrete
      \end{tabular}\\
      \hline

      SVC & \begin{tabular}{c}
        gamma\\
        C
      \end{tabular}
      & \begin{tabular}{c}
        $10^{-8}\sim 10^{4}$\\
        $10^{-3}\sim 10^{3}$
      \end{tabular}
      & \begin{tabular}{c}
          Continuous\\
          Discrete
      \end{tabular}\\
      \hline

    \end{tabular}
\end{table*}

% --------------------------------------------------
% \subsection{データセット毎の予測精度}
\subsection{Comparison based on Validation Accuracy}
\label{app:result_details}

% 6章の実験では32種類のデータセットに対して予測を行なっている．
In Section~\ref{sec:experiments}, the ranking based evaluation was shown.
%
% データセット毎の予測精度は表\ref{table:result}の通りである．
Here, we show the final
$\mr{Acc}(\*\lambda, \cD_{\mr{train}}, \cD_{\mr{valid}})$
for each dataset in Table~\ref{table:result}.

% --------------------------------------------------
% Tab: Maximum Acc
% --------------------------------------------------
\begin{table*}[!ht]
 \centering
 \caption{
 % データセット毎の予測精度．
 Maximum 
 $\mr{Acc}(\*\lambda, \cD_{\mr{train}}, \cD_{\mr{valid}})$
 obtained by each method (average of $10$ runs with different initialization).
 The left most column `data id' is the dataset id of OpenML.
 % data idはOpenMLに記載されているデータセットidを表している．
 %
 The red, blue and green fonts are the best, the second best and the third best methods, respectively.
 % また，赤字は比較手法内で最も良い予測精度，青字は比較手法内で2番目に良い予測精度，赤線は比較手法内で3番目に良い予測精度を表している．
 }
  \label{table:result}
  \begin{tabular}{|wc{1.5cm}|wc{1.5cm}|wc{1.5cm}|wc{1.5cm}|wc{1.5cm}|wc{1.5cm}|wc{1.5cm}|wc{1.5cm}|wc{1.5cm}|}
  \hline
    data id & Proposed & SMAC & BanditBO & Pre-train BO & RB-SMAC & Random & AS-ML & AHS-ML  \\ \hline \hline
    % data id & Proposed & SMAC & BanditBO & Pre-train BO & RB-SMAC & Random & ML-BO & ML-RM  \\ \hline \hline
    4134  & \red{0.7948} & 0.7892 & 0.7843 & 0.7839 & 0.7915 & 0.7798 & \blue{0.7938} & \green{0.7922} \\ \hline
    1084  & \green{0.9621} & 0.953 & 0.9606 & 0.9545 & 0.9561 & 0.953 & \blue{0.9682} & \red{0.9697} \\ \hline
    41972  & \green{0.9427} & 0.9269 & 0.9392 & 0.923 & \red{0.9576} & \blue{0.9437} & 0.925 & 0.9264 \\ \hline
    4153  & \red{0.9964} & 0.9745 & \green{0.9891} & 0.9745 & \blue{0.9909} & 0.96 & 0.9564 & 0.9636 \\ \hline
    41082  & \blue{0.9751} & 0.9729 & \green{0.9741} & 0.9737 & \red{0.9766} & 0.974 & 0.9594 & 0.9699 \\ \hline
    1557  & \blue{0.6771} & \green{0.6744} & 0.6728 & 0.6649 & 0.6727 & 0.665 & 0.6629 & \red{0.6794} \\ \hline
    45068  & 0.8752 & 0.874 & 0.8713 & 0.8699 & \red{0.8759} & 0.8705 & \blue{0.8755} & \green{0.8753} \\ \hline
    1457  & \red{0.718} & \green{0.7176} & 0.7144 & 0.7144 & \blue{0.7178} & 0.7098 & 0.6549 & 0.6622 \\ \hline
    1458  & \blue{0.8833} & 0.8483 & 0.865 & 0.84 & \green{0.8667} & 0.8467 & \red{0.8867} & 0.85 \\ \hline
    9  & \red{0.9113} & \blue{0.9048} & 0.8984 & 0.8016 & 0.8919 & 0.8242 & \green{0.9} & 0.8871 \\ \hline
    463  & \green{0.8745} & 0.8727 & 0.8727 & \red{0.8964} & \green{0.8745} & 0.8727 & \blue{0.8782} & 0.8727 \\ \hline
    1460  & \blue{0.896} & 0.8944 & 0.8945 & \red{0.8963} & \green{0.8954} & 0.8942 & 0.8929 & 0.8938 \\ \hline
    40663  & \green{0.7075} & \red{0.7158} & 0.7017 & 0.6775 & 0.6883 & 0.6683 & \blue{0.7092} & 0.6917 \\ \hline
    40711  & \green{0.6352} & 0.6165 & 0.622 & \blue{0.6363} & 0.6198 & 0.6198 & 0.6341 & \red{0.6374} \\ \hline
    180  & 0.8792 & 0.8751 & 0.8786 & 0.8756 & \blue{0.8813} & 0.873 & \green{0.8807} & \red{0.8819} \\ \hline
    846  & \red{0.901} & 0.8934 & 0.8963 & 0.888 & 0.8969 & 0.8844 & \blue{0.9005} & \green{0.897} \\ \hline
    1044  & \red{0.7697} & 0.7632 & 0.7642 & 0.7279 & 0.745 & 0.6955 & \blue{0.7696} & \blue{0.7696} \\ \hline
    1475  & \red{0.6176} & 0.6107 & 0.6112 & 0.6127 & \green{0.6168} & 0.6042 & \blue{0.6172} & 0.61 \\ \hline
    23512  & \red{0.722} & \green{0.7205} & 0.7186 & 0.7147 & 0.7187 & 0.7146 & \blue{0.7215} & 0.718 \\ \hline
    1479  & \green{0.9662} & 0.9654 & \red{0.9684} & \red{0.9684} & 0.9659 & 0.9635 & 0.5409 & 0.5769 \\ \hline
    300  & \blue{0.9722} & 0.9697 & 0.9674 & 0.9674 & \green{0.9719} & 0.9701 & \red{0.9725} & 0.9658 \\ \hline
    41168  & \green{0.7208} & \red{0.7219} & 0.7155 & 0.7042 & 0.7146 & 0.713 & \blue{0.7211} & 0.72 \\ \hline
    184  & \green{0.8541} & 0.7647 & 0.8419 & 0.7582 & 0.8314 & 0.77 & \blue{0.8725} & \red{0.8766} \\ \hline
    396  & \red{0.9211} & 0.9173 & \blue{0.9204} & 0.9163 & \green{0.9186} & 0.9117 & 0.9108 & 0.9148 \\ \hline
    1482  & \blue{0.7282} & 0.6777 & \red{0.732} & 0.6757 & 0.6913 & 0.6777 & 0.6291 & \green{0.699} \\ \hline
    40677  & \blue{0.7367} & \green{0.7362} & 0.7348 & 0.7342 & \red{0.7377} & 0.736 & 0.7357 & 0.7312 \\ \hline
    6  & \green{0.9706} & 0.961 & 0.9651 & 0.9622 & \blue{0.9744} & 0.9662 & \red{0.9752} & 0.9628 \\ \hline
    10  & \green{0.7956} & 0.7644 & 0.7867 & 0.7778 & \red{0.8} & 0.7711 & \blue{0.7978} & 0.7778 \\ \hline
    45067  & 0.7491 & 0.7449 & 0.7476 & 0.7472 & \red{0.7514} & 0.7423 & \blue{0.7509} & \green{0.7495} \\ \hline
    1491  & \green{0.82} & 0.8183 & 0.8121 & \blue{0.8206} & \red{0.8271} & 0.8194 & 0.791 & 0.6937 \\ \hline
    871  & \red{0.5319} & \green{0.5281} & \blue{0.5306} & 0.5266 & 0.5259 & 0.5186 & 0.5127 & 0.5082 \\ \hline
    44161  & \red{0.8045} & \blue{0.8028} & 0.7987 & 0.7978 & \green{0.8024} & 0.7856 & 0.8019 & 0.8017 \\ \hline
    934  & \blue{0.964} & 0.955 & \green{0.9556} & 0.9542 & 0.9513 & 0.9392 & \red{0.9651} & 0.9539 \\ \hline
    841  & \red{0.9723} & \blue{0.9712} & 0.9695 & \green{0.9705} & 0.9681 & 0.9646 & 0.9684 & 0.9684 \\ \hline
    4329  & \blue{0.8369} & 0.8348 & 0.8333 & \red{0.8397} & \blue{0.8369} & 0.8362 & 0.8319 & \blue{0.8369} \\ \hline
    1508  & \red{0.9421} & 0.938 & \green{0.9388} & 0.9322 & \red{0.9421} & 0.9339 & 0.9347 & 0.9339 \\ \hline
    1523  & 0.9323 & \red{0.9355} & 0.9215 & 0.929 & \red{0.9355} & \green{0.9333} & 0.8946 & 0.871 \\ \hline
    41166  & \blue{0.6965} & 0.6903 & 0.6921 & 0.67 & \red{0.7144} & 0.6803 & 0.6913 & \green{0.6931} \\ \hline
    56  & \green{0.9763} & 0.9733 & 0.9718 & 0.9687 & 0.9756 & 0.9733 & \red{0.9771} & \red{0.9771} \\ \hline
    60  & \green{0.8657} & 0.8653 & 0.8637 & \blue{0.8678} & \red{0.8679} & 0.864 & 0.861 & 0.86 \\ \hline
  \end{tabular}
\end{table*}

% --------------------------------------------------
% \subsection{利用したデータセット}
\subsection{Datasets}
\label{app:pre-train_daataset}

% The target and source datasets are shown in Table~\ref{table:target_dataset} and \ref{table:pre-train_dataset}, respectively
The source and target datasets are shown in Table~\ref{table:pre-train_dataset} and \ref{table:target_dataset}, respectively
%
% ここでは，提案手法およびpre-train BOの事前学習，AS-ML，AHS-MLのメタ学習に活用したデータセット，
% 実験で使用したターゲットデータセットを紹介する．
% 提案手法およびpre-train BOの事前学習，AS-ML，AHS-MLのメタ学習に活用したデータセットは表\ref{table:pre-train_dataset}，
% 実験で使用したターゲットデータセットは表\ref{table:target_dataset}のとおりである．
Missing values were imputed using the mode of each feature.
%
% カテゴリ変数のエンコーディングにはLabel encodingを行った．
The categorical variable in {\tt KNeighborsClassifier} is simply encoded as $1$ (uniform) and $2$ (distance).
%
% 表\ref{table:pre-train_dataset,table:target_dataset}から，今回利用したデータセットにはカテゴリ変数および欠損値が含まれているため，
% これらのデータセットをそのまま機械学習モデルの学習に活用することはできない．
% そこで，我々はカテゴリ変数のエンコーディングや欠損値補完を行った．
% 我々はカテゴリ変数のエンコーディングにはLabel encodingを，欠損値補完には各特徴量の最頻値による補完を行った．
The source datasets were used in pre-training of the proposed method and pre-train BO and meta-learning of AS-ML and AHS-ML, for which we observed many $\mr{Acc}(\*\lambda, \cD_{\mr{train}}, \cD_{\mr{valid}})$ beforehand as shown in Table~\ref{table:observed_value}.

% 我々は表\ref{table:pre-train_dataset}の各事前学習データセットに対して，大量の機械学習アルゴリズムとハイパーパラメータの組み合わせで予測精度の観測を行なっている．
%
% 観測を行なった実際のハイパーパラメータの値は表\ref{table:observed_value}のとおりである．

% --------------------------------------------------
% Table: Source datsets 
% --------------------------------------------------
\begin{table*}[!ht]
  \centering
  \caption{
 % 事前学習に利用したデータセット一覧．Nameはデータセット名を，Samplesはそのデータセットのサンプルサイズを，Featuresはそのデータセットの特徴量数を，Numericalは数値型特徴量数を，Categoricalはカテゴリ型特徴量数を，Classesはそのデータセットのクラス数を，Missing Valuesはそのデータセットの欠損値の有無 (\checkmark は欠損値あり)を，idはOpenMLにおけるそのデータセットのidを示している．
 List of source datasets.
 `Missing Values' indicates the existence of missing values (\checkmark means missing values exist). 
 The dataset id of Open ML is shown in `id'.
 }
 \label{table:pre-train_dataset}
 \scalebox{.95}{
  \begin{tabular}{|wc{6.5cm}|wc{1.2cm}|wc{1.2cm}|c|c|wc{1.2cm}|c|wc{1cm}|}
  \hline
    Name & Samples  & Features  & Numerical  & Categorical  & Classes  & Missing Values  & id  \\ \hline \hline
    ailerons & $13750$ & $40$ & $40$ & $0$ & $2$& - & $734$ \\ \hline
    Amazon\_employee\_access & $32769$ & $9$ & $0$ & $9$ & $2$& - & $4135$ \\ \hline
    analcatdata\_authorship & $841$ & $70$ & $70$ & $0$ & $4$& - & $458$ \\ \hline
    analcatdata\_boxing2 & $132$ & $3$ & $0$ & $3$ & $2$& - & $444$ \\ \hline
    analcatdata\_creditscore & $100$ & $6$ & $3$ & $3$ & $2$& - & $461$ \\ \hline
    analcatdata\_dmft & $797$ & $4$ & $0$ & $4$ & $6$& - & $469$ \\ \hline
    analcatdata\_lawsuit & $264$ & $4$ & $3$ & $1$ & $2$& - & $450$ \\ \hline
    analcatdata\_wildcat & $163$ & $5$ & $3$ & $2$ & $2$& - & $748$ \\ \hline
    anneal & $898$ & $38$ & $6$ & $32$ & $5$& - & $42716$ \\ \hline
    AP\_Breast\_Colon & $630$ & $10935$ & $10935$ & $0$ & $2$& - & $1145$ \\ \hline
    AP\_Breast\_Kidney & $604$ & $10935$ & $10935$ & $0$ & $2$& - & $1158$ \\ \hline
    AP\_Breast\_Ovary & $542$ & $10935$ & $10935$ & $0$ & $2$& - & $1165$ \\ \hline
    arrhythmia & $452$ & $279$ & $206$ & $73$ & $13$ & \checkmark & $5$ \\ \hline
    artificial-characters & $10218$ & $7$ & $7$ & $0$ & $10$& - & $1459$ \\ \hline
    Australian & $690$ & $14$ & $6$ & $8$ & $2$& - & $40981$ \\ \hline
    autoUniv-au6-1000 & $1000$ & $40$ & $37$ & $3$ & $8$& - & $1555$ \\ \hline
    balance-scale & $625$ & $4$ & $4$ & $0$ & $3$& - & $11$ \\ \hline
    bank-marketing & $45211$ & $16$ & $7$ & $9$ & $2$& - & $1461$ \\ \hline
    bank32nh & $8192$ & $32$ & $32$ & $0$ & $2$& - & $833$ \\ \hline
    banknote-authentication & $1372$ & $4$ & $4$ & $0$ & $2$& - & $1462$ \\ \hline
    baseball & $1340$ & $16$ & $15$ & $1$ & $3$ & \checkmark & $185$ \\ \hline
    biomed & $209$ & $8$ & $7$ & $1$ & $2$ & \checkmark & $481$ \\ \hline
    Birds & $500$ & $2$ & $0$ & $2$ & $20$& - & $43325$ \\ \hline
    blastchar & $7043$ & $19$ & $3$ & $16$ & $2$ & \checkmark & $46280$ \\ \hline
    blood-transfusion-service-center & $748$ & $4$ & $4$ & $0$ & $2$& - & $1464$ \\ \hline
    breast-tissue & $106$ & $9$ & $9$ & $0$ & $6$& - & $1465$ \\ \hline
    breast-w & $699$ & $9$ & $9$ & $0$ & $2$ & \checkmark & $15$ \\ \hline
    bridges & $105$ & $11$ & $3$ & $8$ & $6$ & \checkmark & $327$ \\ \hline
    car & $1728$ & $6$ & $0$ & $6$ & $4$& - & $40975$ \\ \hline
    cardiotocography & $2126$ & $35$ & $35$ & $0$ & $10$& - & $1466$ \\ \hline
    chatfield\_4 & $235$ & $12$ & $12$ & $0$ & $2$& - & $820$ \\ \hline
    christine & $5418$ & $1636$ & $1599$ & $37$ & $2$& - & $41142$ \\ \hline
    churn & $5000$ & $20$ & $16$ & $4$ & $2$& - & $40701$ \\ \hline
    cjs & $2796$ & $33$ & $31$ & $2$ & $6$ & \checkmark & $23380$ \\ \hline
    Click\_prediction\_small & $39948$ & $9$ & $9$ & $0$ & $2$& - & $1220$ \\ \hline
    climate-model-simulation-crashes & $540$ & $20$ & $20$ & $0$ & $2$& - & $1467$ \\ \hline
    cmc & $1473$ & $9$ & $2$ & $7$ & $3$& - & $23$ \\ \hline
    cnae-9 & $1080$ & $856$ & $856$ & $0$ & $9$& - & $1468$ \\ \hline
    colic & $368$ & $22$ & $7$ & $15$ & $2$ & \checkmark & $27$ \\ \hline
    colleges\_usnews & $1302$ & $33$ & $32$ & $1$ & $2$ & \checkmark & $930$ \\ \hline
    confidence & $72$ & $3$ & $3$ & $0$ & $6$& - & $468$ \\ \hline
    connect-4 & $67557$ & $42$ & $0$ & $42$ & $3$& - & $40668$ \\ \hline
    CPMP-2015-runtime-classification & $527$ & $22$ & $22$ & $0$ & $4$& - & $41919$ \\ \hline
    credit-g & $1000$ & $20$ & $7$ & $13$ & $2$& - & $31$ \\ \hline
    cylinder-bands & $540$ & $37$ & $18$ & $19$ & $2$ & \checkmark & $6332$ \\ \hline
    datatrieve & $130$ & $8$ & $8$ & $0$ & $2$& - & $1075$ \\ \hline
    dermatology & $366$ & $34$ & $1$ & $33$ & $6$ & \checkmark & $35$ \\ \hline
    dgf\_test & $3415$ & $4$ & $2$ & $2$ & $2$ & \checkmark & $42883$ \\ \hline
    diabetes & $768$ & $8$ & $8$ & $0$ & $2$& - & $37$ \\ \hline
    dna & $3186$ & $180$ & $0$ & $180$ & $3$& - & $40670$ \\ \hline
  \end{tabular}
 }
 Continued on next page
\end{table*}

\begin{table*}[!ht]
 \centering
 Continued from previous page
 \scalebox{.95}{
  \begin{tabular}{|wc{6.5cm}|wc{1.2cm}|wc{1.2cm}|c|c|wc{1.2cm}|c|wc{1cm}|}
  \hline
    Name & Samples  & Features  & Numerical  & Categorical  & Classes  & Missing Values  & id  \\ \hline \hline
    dresses-sales & $500$ & $12$ & $1$ & $11$ & $2$ & \checkmark & $23381$ \\ \hline
    eating & $945$ & $6373$ & $6373$ & $0$ & $7$& - & $1233$ \\ \hline
    ecoli & $336$ & $7$ & $7$ & $0$ & $8$& - & $39$ \\ \hline
    eeg-eye-state & $14980$ & $14$ & $14$ & $0$ & $2$& - & $1471$ \\ \hline
    electricity & $45312$ & $8$ & $7$ & $1$ & $2$& - & $151$ \\ \hline
    eucalyptus & $736$ & $19$ & $14$ & $5$ & $5$ & \checkmark & $188$ \\ \hline
    fabert & $8237$ & $800$ & $800$ & $0$ & $7$& - & $41164$ \\ \hline
    flags & $194$ & $28$ & $2$ & $26$ & $8$& - & $285$ \\ \hline
    Flare & $1066$ & $11$ & $0$ & $11$ & $6$& - & $46174$ \\ \hline
    fri\_c3\_1000\_25 & $1000$ & $25$ & $25$ & $0$ & $2$& - & $715$ \\ \hline
    fri\_c3\_1000\_5 & $1000$ & $5$ & $5$ & $0$ & $2$& - & $813$ \\ \hline
    GCM & $190$ & $16063$ & $16063$ & $0$ & $14$& - & $1106$ \\ \hline
    GesturePhaseSegmentationProcessed & $9873$ & $32$ & $32$ & $0$ & $5$& - & $4538$ \\ \hline
    gina\_prior2 & $3468$ & $784$ & $784$ & $0$ & $10$& - & $1041$ \\ \hline
    glass & $214$ & $9$ & $9$ & $0$ & $6$& - & $41$ \\ \hline
    grub-damage & $155$ & $8$ & $2$ & $6$ & $4$& - & $338$ \\ \hline
    haberman & $306$ & $3$ & $2$ & $1$ & $2$& - & $43$ \\ \hline
    hayes-roth & $160$ & $4$ & $4$ & $0$ & $3$& - & $329$ \\ \hline
    heart-long-beach & $200$ & $13$ & $13$ & $0$ & $5$& - & $1512$ \\ \hline
    heart-statlog & $270$ & $13$ & $13$ & $0$ & $2$& - & $53$ \\ \hline
    hepatitis & $155$ & $19$ & $6$ & $13$ & $2$ & \checkmark & $55$ \\ \hline
    ilpd & $583$ & $10$ & $9$ & $1$ & $2$& - & $1480$ \\ \hline
    IndoorScenes & $15620$ & $2$ & $0$ & $2$ & $67$& - & $45936$ \\ \hline
    Internet-Advertisements & $3279$ & $1558$ & $3$ & $1555$ & $2$& - & $40978$ \\ \hline
    ionosphere & $351$ & $34$ & $34$ & $0$ & $2$& - & $59$ \\ \hline
    iris & $150$ & $4$ & $4$ & $0$ & $3$& - & $61$ \\ \hline
    irish & $500$ & $5$ & $2$ & $3$ & $2$ & \checkmark & $451$ \\ \hline
    JapaneseVowels & $9961$ & $14$ & $14$ & $0$ & $9$& - & $375$ \\ \hline
    jasmine & $2984$ & $144$ & $8$ & $136$ & $2$& - & $41143$ \\ \hline
    jungle\_chess\_2pcs\_raw\_endgame\_complete & $44819$ & $6$ & $6$ & $0$ & $3$& - & $41027$ \\ \hline
    kr-vs-kp & $3196$ & $36$ & $0$ & $36$ & $2$& - & $3$ \\ \hline
    ldpa & $164860$ & $7$ & $5$ & $2$ & $11$& - & $1483$ \\ \hline
    LED-display-domain-7digit & $500$ & $7$ & $7$ & $0$ & $10$& - & $40496$ \\ \hline
    madelon & $2600$ & $500$ & $500$ & $0$ & $2$& - & $1485$ \\ \hline
    MagicTelescope & $19020$ & $10$ & $10$ & $0$ & $2$& - & $1120$ \\ \hline
    Mammographic-Mass-Data-Set & $961$ & $4$ & $2$ & $2$ & $2$ & \checkmark & $45557$ \\ \hline
    meta\_instanceincremental.arff & $74$ & $62$ & $62$ & $0$ & $4$& - & $278$ \\ \hline
    mfeat-factors & $2000$ & $216$ & $216$ & $0$ & $10$& - & $12$ \\ \hline
    mfeat-fourier & $2000$ & $76$ & $76$ & $0$ & $10$& - & $14$ \\ \hline
    mfeat-karhunen & $2000$ & $64$ & $64$ & $0$ & $10$& - & $16$ \\ \hline
    mfeat-morphological & $2000$ & $6$ & $6$ & $0$ & $10$& - & $18$ \\ \hline
    mfeat-pixel & $2000$ & $240$ & $0$ & $240$ & $10$& - & $20$ \\ \hline
    mfeat-zernike & $2000$ & $47$ & $47$ & $0$ & $10$& - & $22$ \\ \hline
    MiceProtein & $1080$ & $77$ & $77$ & $0$ & $8$ & \checkmark & $40966$ \\ \hline
    micro-mass & $571$ & $1300$ & $1300$ & $0$ & $20$& - & $1515$ \\ \hline
    microaggregation2 & $20000$ & $20$ & $20$ & $0$ & $5$& - & $41671$ \\ \hline
    Midwest\_survey & $2494$ & $27$ & $0$ & $27$ & $9$ & \checkmark & $42805$ \\ \hline
    monks-problems-2 & $601$ & $6$ & $0$ & $6$ & $2$& - & $334$ \\ \hline
    mushroom & $8124$ & $22$ & $0$ & $22$ & $2$ & \checkmark & $24$ \\ \hline
    musk & $6598$ & $167$ & $166$ & $1$ & $2$& - & $1116$ \\ \hline
  \end{tabular}
 }
 Continued on next page
\end{table*}

\begin{table*}[!ht]
 \centering
 Continued from previous page
 \scalebox{.95}{
  \begin{tabular}{|wc{6.5cm}|wc{1.2cm}|wc{1.2cm}|c|c|wc{1.2cm}|c|wc{1cm}|}
  \hline
    Name & Samples  & Features  & Numerical  & Categorical  & Classes  & Missing Values  & id  \\ \hline \hline
    nomao & $34465$ & $118$ & $89$ & $29$ & $2$& - & $1486$ \\ \hline
    nursery & $12960$ & $8$ & $0$ & $8$ & $5$& - & $26$ \\ \hline
    optdigits & $5620$ & $64$ & $64$ & $0$ & $10$& - & $28$ \\ \hline
    ozone-level-8hr & $2534$ & $72$ & $72$ & $0$ & $2$& - & $1487$ \\ \hline
    page-blocks & $5473$ & $10$ & $10$ & $0$ & $5$& - & $30$ \\ \hline
    pc1 & $1109$ & $21$ & $21$ & $0$ & $2$& - & $1068$ \\ \hline
    pc2 & $5589$ & $36$ & $36$ & $0$ & $2$& - & $1069$ \\ \hline
    pc3 & $1563$ & $37$ & $37$ & $0$ & $2$& - & $1050$ \\ \hline
    pc4 & $1458$ & $37$ & $37$ & $0$ & $2$& - & $1049$ \\ \hline
    pendigits & $10992$ & $16$ & $16$ & $0$ & $10$& - & $32$ \\ \hline
    philippine & $5832$ & $308$ & $308$ & $0$ & $2$& - & $41145$ \\ \hline
    PhishingWebsites & $11055$ & $30$ & $0$ & $30$ & $2$& - & $4534$ \\ \hline
    phoneme & $5404$ & $5$ & $5$ & $0$ & $2$& - & $1489$ \\ \hline
    PizzaCutter1 & $661$ & $37$ & $37$ & $0$ & $2$& - & $1443$ \\ \hline
    PopularKids & $478$ & $10$ & $6$ & $4$ & $3$& - & $1100$ \\ \hline
    prnn\_fglass & $214$ & $9$ & $9$ & $0$ & $6$& - & $952$ \\ \hline
    prnn\_synth & $250$ & $2$ & $2$ & $0$ & $2$& - & $464$ \\ \hline
    profb & $672$ & $9$ & $5$ & $4$ & $2$ & \checkmark & $470$ \\ \hline
    qsar-biodeg & $1055$ & $41$ & $41$ & $0$ & $2$& - & $1494$ \\ \hline
    regime\_alimentaire & $202$ & $19$ & $3$ & $16$ & $2$ & \checkmark & $42172$ \\ \hline
    rmftsa\_sleepdata & $1024$ & $2$ & $2$ & $0$ & $4$& - & $679$ \\ \hline
    rsctc2010\_1 & $105$ & $22283$ & $22283$ & $0$ & $3$& - & $1077$ \\ \hline
    Run\_or\_walk\_information & $88588$ & $6$ & $6$ & $0$ & $2$& - & $40922$ \\ \hline
    Satellite & $5100$ & $36$ & $36$ & $0$ & $2$& - & $40900$ \\ \hline
    satimage & $6430$ & $36$ & $36$ & $0$ & $6$& - & $182$ \\ \hline
    scene & $2407$ & $299$ & $294$ & $5$ & $2$& - & $312$ \\ \hline
    schizo & $340$ & $14$ & $12$ & $2$ & $2$ & \checkmark & $466$ \\ \hline
    seismic-bumps & $210$ & $7$ & $7$ & $0$ & $3$& - & $1500$ \\ \hline
    semeion & $1593$ & $256$ & $256$ & $0$ & $10$& - & $1501$ \\ \hline
    shuttle & $58000$ & $9$ & $9$ & $0$ & $7$& - & $40685$ \\ \hline
    solar-flare & $1066$ & $12$ & $0$ & $12$ & $6$& - & $40687$ \\ \hline
    sonar & $208$ & $60$ & $60$ & $0$ & $2$& - & $40$ \\ \hline
    soybean & $683$ & $35$ & $0$ & $35$ & $19$ & \checkmark & $42$ \\ \hline
    spambase & $4601$ & $57$ & $57$ & $0$ & $2$& - & $44$ \\ \hline
    SPECT & $267$ & $22$ & $0$ & $22$ & $2$& - & $336$ \\ \hline
    SPECTF & $349$ & $44$ & $44$ & $0$ & $2$& - & $337$ \\ \hline
    Speech & $3686$ & $400$ & $400$ & $0$ & $2$& - & $40910$ \\ \hline
    SpeedDating & $8378$ & $120$ & $59$ & $61$ & $2$ & \checkmark & $40536$ \\ \hline
    splice & $3190$ & $60$ & $0$ & $60$ & $3$& - & $46$ \\ \hline
    SRBCT & $83$ & $2308$ & $2308$ & $0$ & $4$& - & $45101$ \\ \hline
    steel-plates-fault & $1941$ & $33$ & $33$ & $0$ & $2$& - & $1504$ \\ \hline
    synthetic\_control & $600$ & $60$ & $60$ & $0$ & $6$& - & $377$ \\ \hline
    tae & $151$ & $5$ & $3$ & $2$ & $3$& - & $48$ \\ \hline
    texture & $5500$ & $40$ & $40$ & $0$ & $11$& - & $40499$ \\ \hline
    thyroid-allrep & $2800$ & $26$ & $6$ & $20$ & $5$& - & $40477$ \\ \hline
    Titanic & $1309$ & $13$ & $6$ & $7$ & $2$ & \checkmark & $40945$ \\ \hline
    tokyo1 & $959$ & $44$ & $42$ & $2$ & $2$& - & $40705$ \\ \hline
    Traffic\_violations & $70340$ & $20$ & $1$ & $19$ & $3$ & \checkmark & $42345$ \\ \hline
    vehicle & $846$ & $18$ & $18$ & $0$ & $4$& - & $54$ \\ \hline
    volcanoes-a1 & $3252$ & $3$ & $3$ & $0$ & $5$& - & $1527$ \\ \hline
  \end{tabular}
 }
 Continued on next page
\end{table*}

\begin{table*}
 \centering
 Continued from previous page
 \scalebox{.95}{
  \begin{tabular}{|wc{6.5cm}|wc{1.2cm}|wc{1.2cm}|c|c|wc{1.2cm}|c|wc{1cm}|}
  \hline
    Name & Samples  & Features  & Numerical  & Categorical  & Classes  & Missing Values  & id  \\ \hline \hline
    volcanoes-a2 & $1623$ & $3$ & $3$ & $0$ & $5$& - & $1528$ \\ \hline
    volcanoes-a3 & $1521$ & $3$ & $3$ & $0$ & $5$& - & $1529$ \\ \hline
    vowel & $990$ & $12$ & $10$ & $2$ & $11$& - & $307$ \\ \hline
    wall-robot-navigation & $5456$ & $24$ & $24$ & $0$ & $4$& - & $1497$ \\ \hline
    wdbc & $569$ & $30$ & $30$ & $0$ & $2$& - & $1510$ \\ \hline
    wilt & $4839$ & $5$ & $5$ & $0$ & $2$& - & $40983$ \\ \hline
    wine-quality-red & $1599$ & $11$ & $11$ & $0$ & $6$& - & $40691$ \\ \hline
    wine-quality-white & $4898$ & $11$ & $11$ & $0$ & $7$& - & $40498$ \\ \hline
    WMO-Hurricane-Survival-Dataset & $5021$ & $22$ & $1$ & $21$ & $2$ & \checkmark & $43607$ \\ \hline
    yeast & $1484$ & $8$ & $8$ & $0$ & $10$& - & $181$ \\ \hline
    zoo & $101$ & $16$ & $1$ & $15$ & $7$& - & $62$ \\ \hline
  \end{tabular}}
\end{table*}

% --------------------------------------------------
% Tab: Target dataset
% --------------------------------------------------
\begin{table*}[!ht]
  \centering
  \caption{
 % 実験に利用したターゲットデータセット一覧．Nameはデータセット名を，Samplesはそのデータセットのサンプルサイズを，Featuresはそのデータセットの特徴量数を，Numericalは数値型特徴量数を，Categoricalはカテゴリ型特徴量数を，Classesはそのデータセットのクラス数を，Missing Valuesはそのデータセットの欠損値の有無 (\checkmark は欠損値あり)を，idはOpenMLにおけるそのデータセットのidを示している．
 List of target datasets. 
 `Missing Values' indicates the existence of missing values (\checkmark means missing values exist). 
 The dataset id of Open ML is shown in `id'.
 }
 \label{table:target_dataset}
 \scalebox{.95}{
  \begin{tabular}{|wc{6.5cm}|wc{1.2cm}|wc{1.2cm}|c|c|wc{1.2cm}|c|wc{1cm}|}
  \hline
    name & Samples  & Features  & Numerical  & Categorical  & Classes  & Missing Values  & id  \\ \hline \hline
    abalone & $4177$ & $8$ & $7$ & $1$ & $3$& - & $1557$ \\ \hline
    adult & $48842$ & $14$ & $6$ & $8$ & $2$& - & $45068$ \\ \hline
    amazon-commerce-reviews & $1500$ & $10000$ & $10000$ & $0$ & $50$& - & $1457$ \\ \hline
    arcene & $200$ & $10000$ & $10000$ & $0$ & $2$& - & $1458$ \\ \hline
    autos & $205$ & $25$ & $15$ & $10$ & $6$ & $\checkmark$ & $9$ \\ \hline
    backache & $180$ & $31$ & $5$ & $26$ & $2$& - & $463$ \\ \hline
    banana & $5300$ & $2$ & $2$ & $0$ & $2$& - & $1460$ \\ \hline
    Bioresponse & $3751$ & $1776$ & $1776$ & $0$ & $2$& - & $4134$ \\ \hline
    BurkittLymphoma & $220$ & $22283$ & $22283$ & $0$ & $3$& - & $1084$ \\ \hline
    calendarDOW & $399$ & $32$ & $12$ & $20$ & $5$& - & $40663$ \\ \hline
    cleveland-nominal & $303$ & $7$ & $0$ & $7$ & $5$& - & $40711$ \\ \hline
    covertype & $110393$ & $54$ & $14$ & $40$ & $7$& - & $180$ \\ \hline
    elevators & $16599$ & $18$ & $18$ & $0$ & $2$& - & $846$ \\ \hline
    eye\_movements & $10936$ & $27$ & $24$ & $3$ & $3$& - & $1044$ \\ \hline
    first-order-theorem-proving & $6118$ & $51$ & $51$ & $0$ & $6$& - & $1475$ \\ \hline
    higgs & $98050$ & $28$ & $28$ & $0$ & $2$ & $\checkmark$ & $23512$ \\ \hline
    hill-valley & $1212$ & $100$ & $100$ & $0$ & $2$& - & $1479$ \\ \hline
    Indian\_pines & $9144$ & $220$ & $220$ & $0$ & $8$& - & $41972$ \\ \hline
    isolet & $7797$ & $617$ & $617$ & $0$ & $26$& - & $300$ \\ \hline
    jannis & $83733$ & $54$ & $54$ & $0$ & $4$& - & $41168$ \\ \hline
    kropt & $28056$ & $6$ & $0$ & $6$ & $18$& - & $184$ \\ \hline
    la1s.wc & $3204$ & $13195$ & $13195$ & $0$ & $6$& - & $396$ \\ \hline
    leaf & $340$ & $15$ & $15$ & $0$ & $30$& - & $1482$ \\ \hline
    led24 & $3200$ & $24$ & $0$ & $24$ & $10$& - & $40677$ \\ \hline
    letter & $20000$ & $16$ & $16$ & $0$ & $26$& - & $6$ \\ \hline
    lymph & $148$ & $18$ & $3$ & $15$ & $4$& - & $10$ \\ \hline
    okcupid\_stem & $26677$ & $13$ & $2$ & $11$ & $3$& - & $45067$ \\ \hline
    one-hundred-plants-margin & $1600$ & $64$ & $64$ & $0$ & $100$& - & $1491$ \\ \hline
    pollen & $3848$ & $5$ & $5$ & $0$ & $2$& - & $871$ \\ \hline
    road-safety & $111762$ & $32$ & $29$ & $3$ & $2$& - & $44161$ \\ \hline
    {\footnotesize Smartphone-Based\_Recognition\_of\_Human\_Activities} & $180$ & $66$ & $66$ & $0$ & $6$& - & $4153$ \\ \hline
    socmob & $1156$ & $5$ & $1$ & $4$ & $2$& - & $934$ \\ \hline
    stock & $950$ & $9$ & $9$ & $0$ & $2$& - & $841$ \\ \hline
    thoracic\_surgery & $470$ & $16$ & $3$ & $13$ & $2$& - & $4329$ \\ \hline
    user-knowledge & $403$ & $5$ & $5$ & $0$ & $5$& - & $1508$ \\ \hline
    USPS & $9298$ & $256$ & $256$ & $0$ & $10$& - & $41082$ \\ \hline
    vertebra-column & $310$ & $6$ & $6$ & $0$ & $3$& - & $1523$ \\ \hline
    volkert & $58310$ & $180$ & $180$ & $0$ & $10$& - & $41166$ \\ \hline
    vote & $435$ & $16$ & $0$ & $16$ & $2$ & $\checkmark$ & $56$ \\ \hline
    waveform-5000 & $5000$ & $40$ & $40$ & $0$ & $3$& - & $60$ \\ \hline
  \end{tabular}}
\end{table*}

\begin{table*}
  \centering
  \caption{
 % 事前学習に用いた観測データ一覧．
 List of HPs used in pre-training. 
 %
 % 候補点は実際に用いた候補点の種類を表している．
 % ただし，$\mathrm{np.linspace}(i,j,k)$はpythonライブラリnumpy\cite{harris2020array}における，$[i,j]$の範囲で$k$個の等間隔の数値を生成するための関数を表している．
 $\mathrm{np.linspace}(i,j,k)$ is from python library numpy \cite{harris2020array} that returns $k$ uniform grid points in $[i,j]$.
 }
  \label{table:observed_value}
  \begin{tabular}{|c|c|c|c|}
   \hline
    ML algorithms & HPs & HP values & \# observations $N^\prm_m$ 
	       \\ \hline \hline
    AdaBoostClassifier & \begin{tabular}{c}
      learning\_rage\\
      n\_estimators
  \end{tabular}
  & \begin{tabular}{c}
    $\{10^i \mid i \in  \mathrm{np.linspace(-3, 0, 10)}\}$\\
    $\{i \mid i \in \mathrm{np.linspace(1,50,10,dtype = int)}\}$
  \end{tabular}
  & 100\\
  \hline

  BaggingClassifier & \begin{tabular}{c}
    max\_features\\
    n\_estimators\\
    max\_samples
  \end{tabular}
  &
  \begin{tabular}{c}
    $\{10^i \mid i \in \mathrm{np.linspace(-3, 0, 5)}\}$\\
    $\{i \mid i \in \mathrm{np.linspace(1,50,10,dtype = int)}\}$\\
    $\{10^i \mid i \in \mathrm{np.linspace(-3, 0, 7)}\}$
  \end{tabular}
  & 350
  \\
  \hline

  DecisionTreeClassifier & \begin{tabular}{c}
    max\_features\\
    max\_depth\\
    min\_samples\_leaf\\
    min\_samples\_split
  \end{tabular}
  &
  \begin{tabular}{c}
    $\{10^i \mid i \in \mathrm{np.linspace(-3, 0, 10)}\}$\\
    $\{i \mid i \in \mathrm{np.linspace(1,50,10,dtype = int)}\}$\\
    $\{2^i \mid i \in \mathrm{np.linspace(1, 9, 9,dtype = int)}\}$\\
    $\{2^i \mid i \in \mathrm{np.linspace(1, 9, 9,dtype = int)}\}$\\
  \end{tabular}
  &
  $8100$\\
  \hline

  ExtraTreeClassifier & \begin{tabular}{c}
    max\_features\\
    n\_estimators\\
    max\_depth\\
    min\_samples\_leaf\\
    min\_samples\_split
  \end{tabular}
  &
  \begin{tabular}{c}
    $\{10^i \mid i \in \mathrm{np.linspace(-3, 0, 5)}\}$\\
    $\{i \mid i \in \mathrm{np.linspace(1,50,10,dtype = int)}\}$\\
    $\{i \mid i \in \mathrm{np.linspace(1,50,5,dtype = int)}\}$\\
    $\{2^i \mid i \in \mathrm{np.linspace(1, 9, 5,dtype = int)}\}$\\
    $\{2^i \mid i \in \mathrm{np.linspace(1, 9, 5,dtype = int)}\}$\\
  \end{tabular}
  &
  $6250$\\
  \hline

  GaussianNB & \begin{tabular}{c}
      var\_smoothing
  \end{tabular}
  & \begin{tabular}{c}
    $\{10^i \mid i \in \mathrm{np.linspace(-15, 0, 50)}\}$\\
  \end{tabular}
  & $50$\\
  \hline

  GradientBoostingClassifier & \begin{tabular}{c}
    learning\_rate\\
    n\_estimators\\
    max\_features\\
    max\_depth\\
    min\_samples\_split
  \end{tabular}
  &
  \begin{tabular}{c}
    $\{10^i \mid i \in \mathrm{np.linspace(-3, 0, 7)}\}$\\
    $\{i \mid i \in \mathrm{np.linspace(1,50,10,dtype = int)}\}$\\
    $\{10^i \mid i \in \mathrm{np.linspace(-3, 0, 5)}\}$\\
    $\{i \mid i \in \mathrm{np.linspace(1,50,5,dtype = int)}\}$\\
    $\{2^i \mid i \in \mathrm{np.linspace(1, 9, 5,dtype = int)}\}$\\
  \end{tabular}
  &
  $8750$\\
  \hline

  KNeighborsClassifier & \begin{tabular}{c}
    n\_neighbors\\
    p\\
    weights
  \end{tabular}
  &
  \begin{tabular}{c}
    $\{i \mid i \in \mathrm{np.linspace(1,50,50,dtype = int)}\}$\\
    $\{1, 2\}$\\
    $\{\mathrm{uniform, distance}\}$
  \end{tabular}
  &
  $200$\\
  \hline

  LogisticRegression & \begin{tabular}{c}
    C
  \end{tabular}
  & \begin{tabular}{c}
    $\{10^i \mid i \in \mathrm{np.linspace(-3, 3, 50)}\}$\\
  \end{tabular}
  & $50$\\
  \hline

  MLPClassifier & \begin{tabular}{c}
    learning\_rate\_init\\
    alpha\\
    first hidden layer's unit size\\
    second hidden layer's unit size\\
    third hidden layer's unit size
  \end{tabular}
  &
  \begin{tabular}{c}
    $\{10^i \mid i \in \mathrm{np.linspace(-5, -1, 9)}\}$\\
    $\{10^i \mid i \in \mathrm{np.linspace(-2, 2, 9)}\}$\\
    $\{2^i \mid i \in \mathrm{np.linspace(4, 7, 4,dtype = int)}\}$\\
    $\{2^i \mid i \in \mathrm{np.linspace(5, 8, 4,dtype = int)}\}$\\
    $\{2^i \mid i \in \mathrm{np.linspace(3, 6, 4,dtype = int)}\}$\\
  \end{tabular}
  &
  $5184$\\
  \hline

  QuadraticDiscriminantAnalysis & \begin{tabular}{c}
    reg\_param
  \end{tabular}
  & \begin{tabular}{c}
    $\{10^i \mid i \in \mathrm{np.linspace(-5, 0, 50)}\}$\\
  \end{tabular}
  & $50$\\
  \hline

  RandomForestClassifier & \begin{tabular}{c}
    max\_features\\
    n\_estimators\\
    max\_depth\\
    min\_samples\_leaf\\
    min\_samples\_split
  \end{tabular}
  &
  \begin{tabular}{c}
    $\{10^i \mid i \in \mathrm{np.linspace(-3, 0, 5)}\}$\\
    $\{i \mid i \in \mathrm{np.linspace(1,50,10,dtype = int)}\}$\\
    $\{i \mid i \in \mathrm{np.linspace(1,50,5,dtype = int)}\}$\\
    $\{2^i \mid i \in \mathrm{np.linspace(1, 9, 5,dtype = int)}\}$\\
    $\{2^i \mid i \in \mathrm{np.linspace(1, 9, 5,dtype = int)}\}$\\
  \end{tabular}
  &
  $6250$\\
  \hline

  SVC & \begin{tabular}{c}
    gamma\\
    C
  \end{tabular}
  & \begin{tabular}{c}
    $\{10^i \mid i \in \mathrm{np.linspace(-8, 4, 30)}\}$\\
    $\{10^i \mid i \in \mathrm{np.linspace(-3, 3, 20)}\}$\\
  \end{tabular}
  & $600$\\
  \hline
 \end{tabular}
\end{table*}

\end{document}